# Caroline: An Autonomously Driving Vehicle for Urban Environments

Fred W. Rauskolb[1], Kai Berger[2], Christian Lipski[2], Marcus Magnor[2], Karsten Cornelsen[3], Jan Effertz[3], Thomas Form[3], Fabian Graefe[3], Sebastian Ohl[3], Walter Schumacher[3], Jörn-Marten Wille[3], Peter Hecker[4], Tobias Nothdurft[4], Michael Doering[5], Kai Homeier[5], Johannes Morgenroth[5], Lars Wolf[5], Christian Basarke[6], Christian Berger[6], Tim Gülke[6], Felix Klose[6], and Bernhard Rumpe[6]

[1] Herzfeld & Rubin, P.C.
40 Wall Street
New York, NY 10005

[2] Institute of Computer Graphics
Mühlenpfordtstraße 23
38106 Braunschweig, Germany

[3] Institute of Control Engineering
Hans-Sommer-Straße 66
38106 Braunschweig, Germany

[4] Institute of Flight Guidance
Hermann-Blenk-Straße 27
38108 Braunschweig, Germany

[5] Institute of Operating Systems and Computer Networks
Mühlenpfordtstraße 23
38106 Braunschweig, Germany

[6] Institute of Software Systems Engineering
Mühlenpfordtstraße 23
38106 Braunschweig, Germany
carolo-uc@tu-bs.de

**Abstract.** The 2007 DARPA Urban Challenge afforded the golden opportunity for the Technische Universität Braunschweig to demonstrate its abilities to develop an autonomously driving vehicle to compete with the world's best competitors. After several stages of qualification, our team CarOLO qualified early for the DARPA Urban Challenge Final Event and was among only eleven teams from initially 89 competitors to compete in the final. We had the ability to work together in a large group of experts, each contributing his expertise in his discipline, and significant organisational, financial and technical support by local sponsors who helped us to become the best non-US team.

In this report, we describe the 2007 DARPA Urban Challenge, our contribution "Caroline", the technology and algorithms along with her performance in the DARPA Urban Challenge Final Event on November 3, 2007.

# 1 Motivation and Introduction

Focused research is often centered around interesting challenges and awards. The airplane industry started off with awards for the first flight over the British Channel as well as the Atlantic Ocean. The Human Genome Project, the RoboCups and the series of DARPA Grand Challenges for autonomous vehicles serve this very same purpose to foster research and development in a particular direction. The 2007 DARPA Urban Challenge is taking place to boost development of unmanned vehicles for urban areas. Although there is an obvious direct benefit for DARPA and the U.S. government, there will also be a large number of spin-offs in technologies, tools and engineering techniques, both for autonomous vehicles, but also for intelligent driver assistance. An intelligent driver assistance function needs to be able to understand the surroundings of the car, evaluate potential risks and help the driver to behave correctly, safely and, in case it is desired, also efficiently. These topics do not only affect ordinary cars, but also buses, trucks, convoys, taxis, special-purpose vehicles in factories, airports and more. It will take a number of years before we will have a mass market for cars that actively and safely protect the passenger and the surroundings, like pedestrians, from accidents in any situation.

Intelligent functions in vehicles are obviously complex systems. Large issues in this project where primarily the methods, techniques and tools for the development of such a highly critical, reliable and complex system. Adapting and combining methods from different engineering disciplines were an important prerequisite for our success. For a stringent deadline-oriented development of such a system it is necessary to rely on a clear structure of the project, a dedicated development process and an efficient engineering that fits the project's needs. Thus, we did not only concentrate on the single software modules of our autonomously driving vehicle named Caroline, but also on the process itself. We furthermore needed an appropriate tool suite that allowed us to run the development and in particular the testing process as efficient as possible. This includes a simulator allowing us to simulate traffic situations and therefore achieve a sufficient coverage of test situations that would have been hardly to conduct in reality. Only a good collaboration between the participating disciplines allowed us to develop Caroline in time to achieve such a good result in the 2007 DARPA Urban Challenge.

In the long term, our goal was not only to participate in a competition but also to establish a sound basis for further research on how to enhance vehicle safety by implementing new technologies to provide vehicle users with reliable and robust driver assistance systems, e.g. by giving special attention on technology for sensor data fusion and robust and reliable system architectures including new methods for simulation and testing. Therefore, the 2007 DARPA Urban Challenge provided a golden opportunity to combine several expertise from several fields of science and engineering. For this purpose, the interdisciplinary team CarOLO had been founded, which drew its members from five different institutes. In addition, the team received support from a consortium of national and international companies.

In this paper, we firstly introduce the 2007 DARPA Urban Challenge and derive the basic requirements for the car from its rules in section 2. Section 3 describes the overall architecture of the system, which is detailed in section 4 describing sensor fusion, vision, artificial intelligence, vehicle control and along with safety concepts. Section 5 describes the overall development process, discusses quality assurance and the simulator used to achieve sufficient testing coverage in detail. Section 6 finally describes the evaluation of Caroline, namely the performance during the National Qualification Event and the DARPA Urban Challenge Final Event in Victorville, California, the results we found and the conclusions to draw from our performance.

## 2 2007 DARPA Urban Challenge

The 2007 DARPA Urban Challenge is the continuation of the well-known Grand Challenge events of 2004 and 2005, which were entitled "Barstow to Primm" and "Desert Classic". To continue the tradition of having names reflect the actual task, DARPA named the 2007 event "Urban Challenge", announcing with it the nature of the mission to be accomplished.

The 2004 course, as shown in Fig. 1, led from the Barstow, California (A) to Primm, Nevada (B) and had a total length of about 142 miles. Prior to the main event, DARPA held a qualification, inspection and demonstration for each robot. Nevertheless, none of the original fifteen vehicles managed to come even close to the goal of successfully completing the course. With 7.4 miles as the farthest distance travelled, the challenge ended very disappointingly and no one won the $1 million cash prize.

Thereafter, the DARPA program managers heightened the barriers for entering the 2005 challenge significantly. They also modified the entire quality inspection process to one involving a step-by-step application process, including a video of the car in action and the holding of so-called Site Visits, which involved the visit of DARPA officials to team-chosen test sites. The rules for these Site Visits were very strict, e.g. determining exactly how the courses had to be equipped and what obstacles had to be available. From initially 195 teams, 118 were selected for site visits and 43 had finally made it into the National Qualification Event at the California Speedway in Ontario, California. The NQE consisted of several tasks to be completed and obstacles to overcome autonomously by the participating vehicles, including tank traps, a tunnel, speed bumps, stationary cars to pass and many more.

On October 5, 2005, DARPA announced the 23 teams that would participate in the final event. The course started in Primm, Nevada, where the 2004 challenge should have ended. With a total distance of 131.6 miles and several natural obstacles, the course was by no means easier than the one from the year before. At the end, five teams completed it and the rest did significantly better as the teams the year before. The Stanford Racing Team was awarded the $2 million first prize.

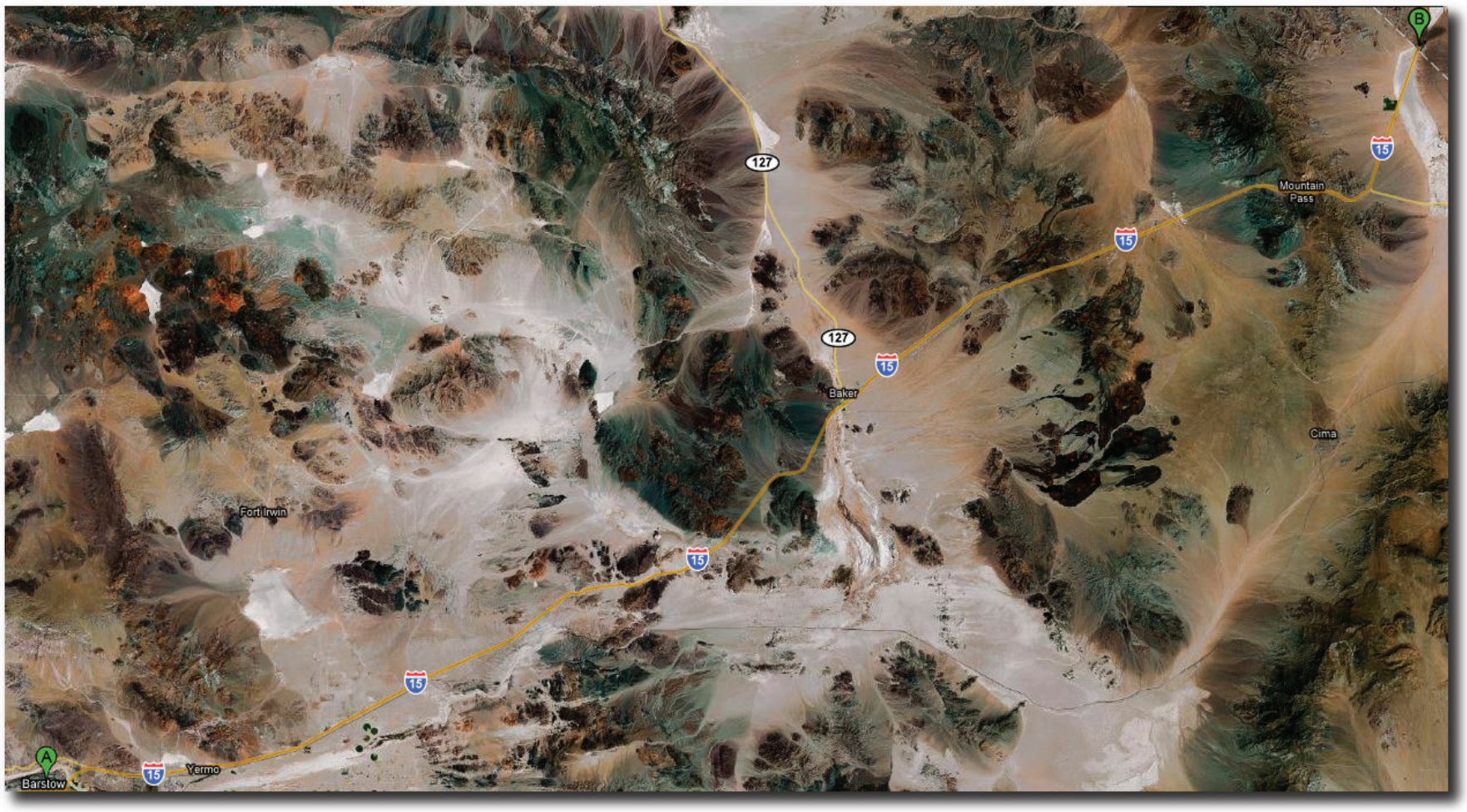

**Fig. 1.** 2004 DARPA Grand Challenge Area between Barstow, CA (A) and Primm, NV (B).

In 2007, DARPA wanted to increase the difficulty of the requirements, in order to meet the goal set by Congress and the Department of Defense that by 2015 a third of the Army's ground combat vehicles would operate unmanned. Having already proved the feasibility of crossing a desert and overcome natural obstacles without human intervention, now a tougher task had to be mastered. As the United States Armed Forces are currently facing serious challenges in urban environments, the choice of such seemed logical. DARPA used the good experience and knowledge gained from the first and second Grand Challenge event to define the tasks for the autonomous vehicles. The 2007 DARPA Urban Challenge took place in Vicorville, CA as depicted in Fig. 2.

The Technische Universität Braunschweig started in June 2006 as a newcomer in the 2007 DARPA Urban Challenge. Significantly supported by industrial partners, five institutes from the faculties of computer science and mechanical and electrical engineering equipped a 2006 Volkswagen Passat station wagon named "Caroline" to participate in the DARPA Urban Challenge as a "Track B" competitor.

Track B competitors did not receive any financial support from the DARPA compared to "Track A" competitors. Track A teams had to submit technical proposals to get technology development funding awards up to $1,000,000 in fall 2006. Track B teams had to provide a 5 minutes video demonstrating the vehicles capabilities in April 2007. Using these videos, DARPA selected 53 teams of the initial 89 teams that advanced to the next stage in the

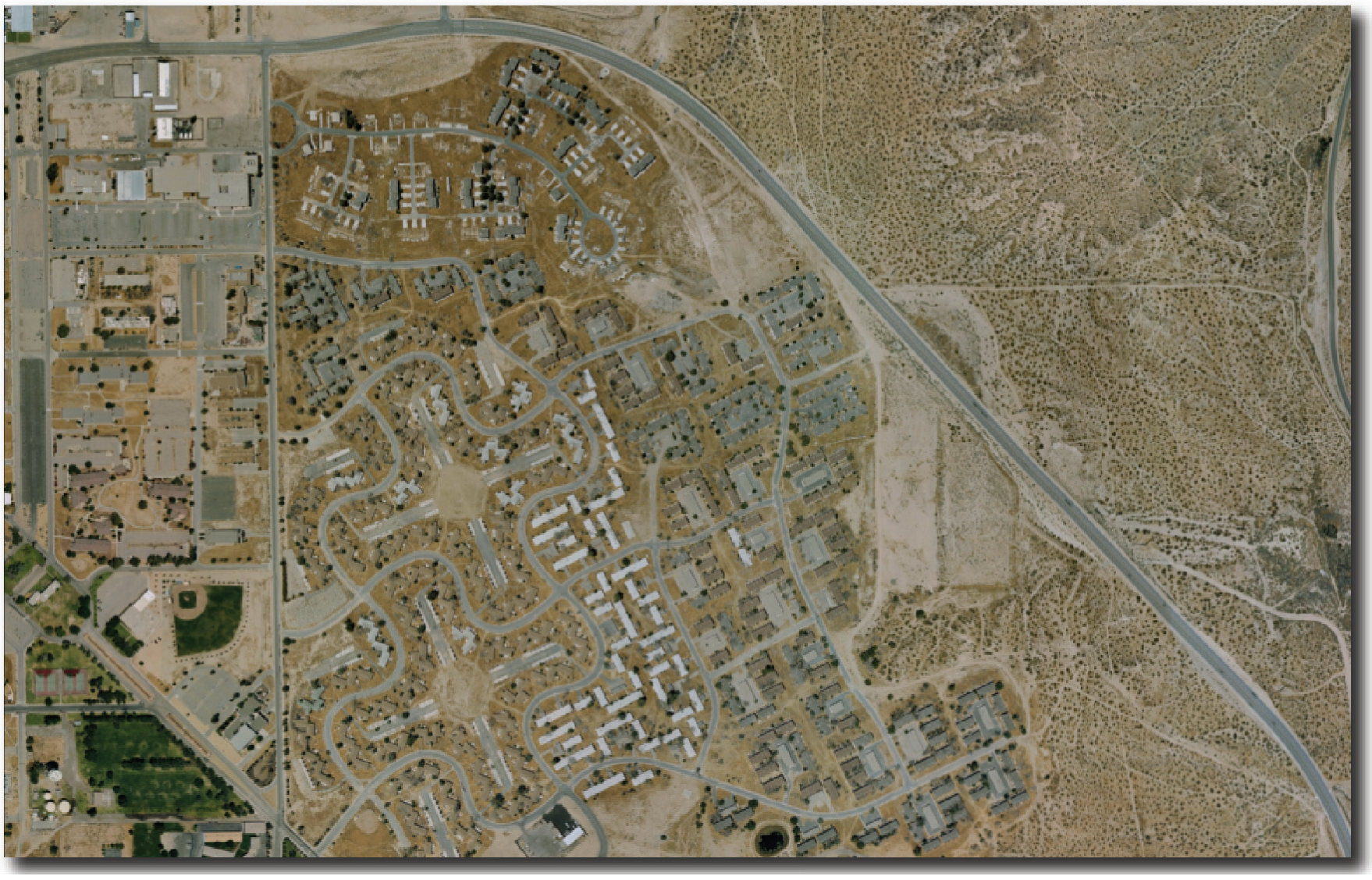

**Fig. 2.** 2007 DARPA Grand Challenge Area in Victorville, CA.

qualification process, the "Site Visit" as already conducted in the 2005 Grand Challenge.

Team CarOLO got an invitation for a Site Visit that had to take place in the United States. Therefore, team CarOLO accepted gratefully an offer from the Southwest Research Insitute in San Antonio, Texas providing a location for the Site Visit. On June 20, Caroline proved that she was ready for the National Qualification Event in fall 2007. Against great odds, she showed her abilities to the DARPA officials when a huge thunderstorm hit San Antonio during the Site Visit. The tasks to complete included the correct handling of intersection precedence, passing of vehicles, lane keeping and general safe behaviour. After the demonstration, the team returned to Germany together with Caroline.

On August 9, the team received the results of the Site Visit event together with an inivitation to the next stage of the qualification process: The National Qualification Event in Victorville, California. Being a semi-finalist team, the team returned at the end of September to the Southwest Research Institute in San Antonio to finalize the development and tests. Three weeks later, Caroline and the team arrived in Victorville, California and participated in the National Qualification Event. To qualify for the Final Event, three courses had to be mastered by the vehicles, each one covering a certain part of the requirements. At the first course, called "Track A", the robots needed to merge into moving traffic, "Track B" required the handling of very long and complex routes with stationary obstacles and "Track C" tested intersections and how the vehicles handle the blockage of roads. Demonstrating repeatedly the performance of Caroline in all tracks of the National Qualification Event, Caroline qualified early for the final stage, the DARPA Urban Challenge Final Event held on November 3. In chapter 6, the overall performance of Caroline in the National Qualification Event and the DARPA Urban Challenge Final Event is illustrated.

## 3 System Architecture

Caroline is a standard 2006 Volkswagen Passat station wagon equipped with a variety of sensors, actuators and computers to function as an autonomous mobile robot. In front, two multi-level laser scanners, one multi-beam lidar sensor and one radar sensor cover a field of view up to 200 meters for approaching traffic or stationary obstacles. In addition, four cameras detect and track lane markings in order to allow precise lane keeping. The stereo vision system behind the windshield and another color camera combined with two laser scanners mounted on the roof were installed to provide information about the drivability of the terrain in front of the vehicle. Very similar to the front of the vehicle, one multi-level laser scanner, one medium range radar, one lidar and two radar-based blind-spot-detectors enable Caroline to detect obstacles at the rear. All these sensors are depicted in Fig. 3.

An array of automotive PCs mounted on a rack shown in Fig. 4 functions as the hardware platform for a distributed software architecture with all internal communication based upon Ethernet. The access to Caroline's by-wire steering, brake and throttle system as well as to other low level actuators is provided through a CANLOG III command interface, which also connects to the vehicle's E-stop system to provide emergency stop functionality even if the complete software system described below should fail. Regardless to those lower level components described above, all computing and control hardware is based on industrial PC technology, thereby reducing hardware variety, simplifying failure management along with component replacement.

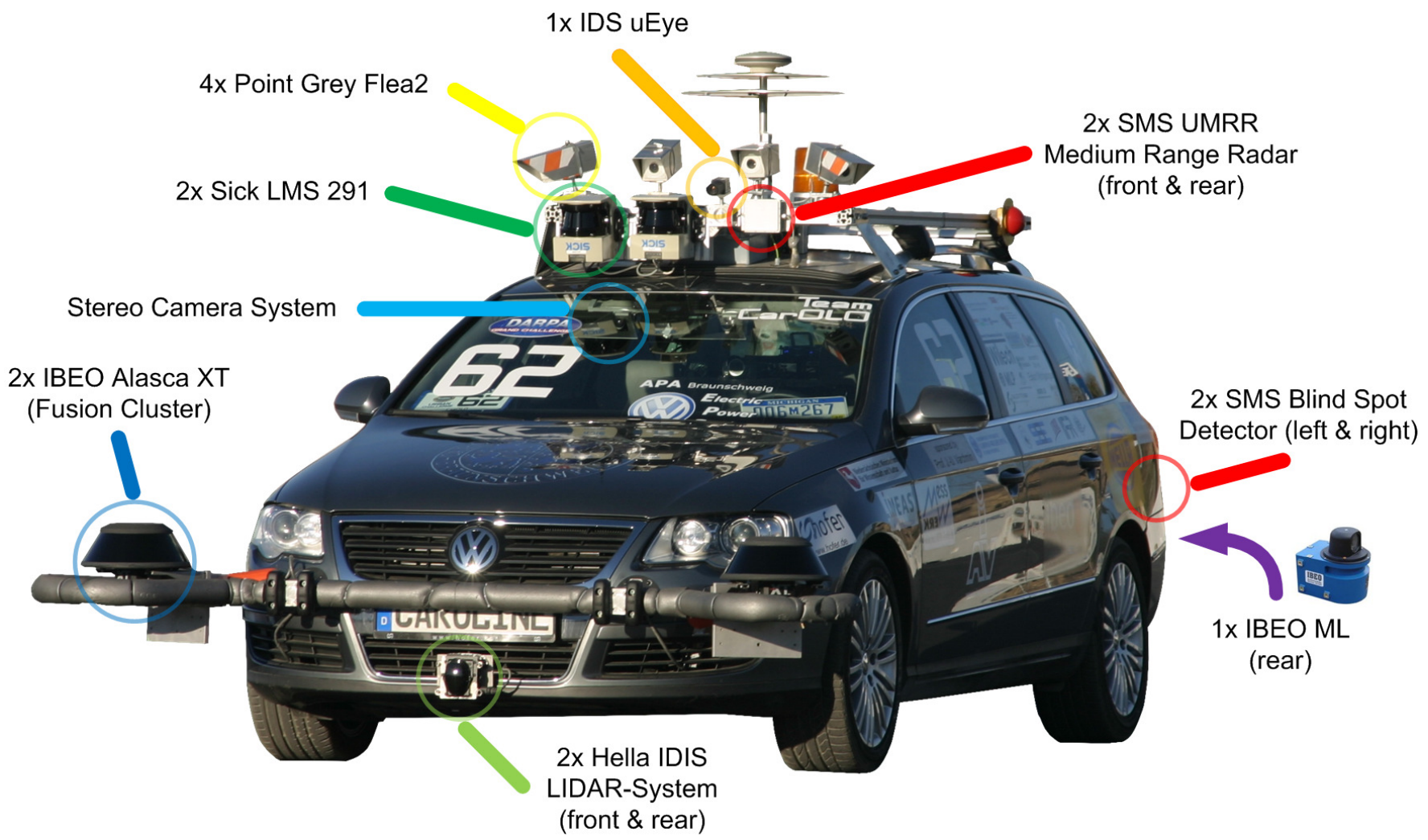

**Fig. 3.** The perception system.

The development of Caroline is divided among a number of institutes and disciplines, including faculties for computer science and mechanical and electrical engineering. Mirroring this internal structure, Caroline's architecture is grouped into eight principal modules, interconnected with predefined interfaces as shown in Fig. 5: Sensor Data Acquisition, Sensor Data Fusion, Image Processing, Digital Map, Artificial Intelligence, Vehicle Path Planning and Low Level Control, Supervisory Watchdog and Online-Diagnosis, Telemetry and Data Storage for Offline Analysis. Due to the intentionally linear signal flow between each function module without major signal loops, we are able to develop different modules independently and with minimum interference.

Starting at the bottom of this linear flow, the data acquisition unit provides necessary hard- and software modules to collect and process incoming data from all active sensors for object recognition. Since all of the sensors used are standard components originating from contemporary automotive driver assistance systems, they are equipped with a Controller Area Network (CAN) communication interface. Taking into account the limitation of this bus standard regarding data throughput and determinism, a private sensor CAN was chosen for each sensor to keep latencies small and to avoid bus conflicts.

The acquisition of GPS and INS data (referred as Ego State in the following) was moved directly into the real-time vehicle control unit in order to avoid large latencies within the closed loop dynamic control. The time of day is obtained from the GPS and distributed via the network time protocol (NTP) to all subsystems.

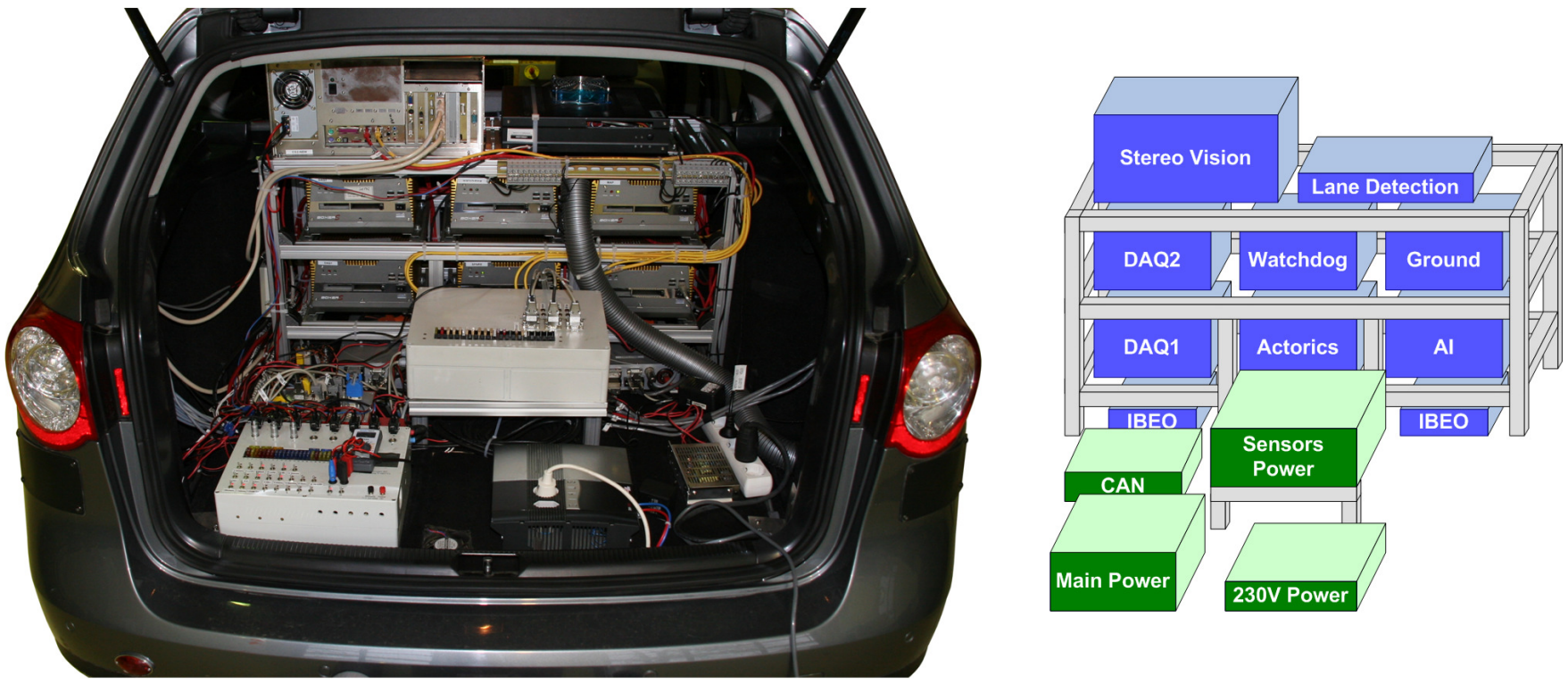

**Fig. 4.** Computer rack and power supply.

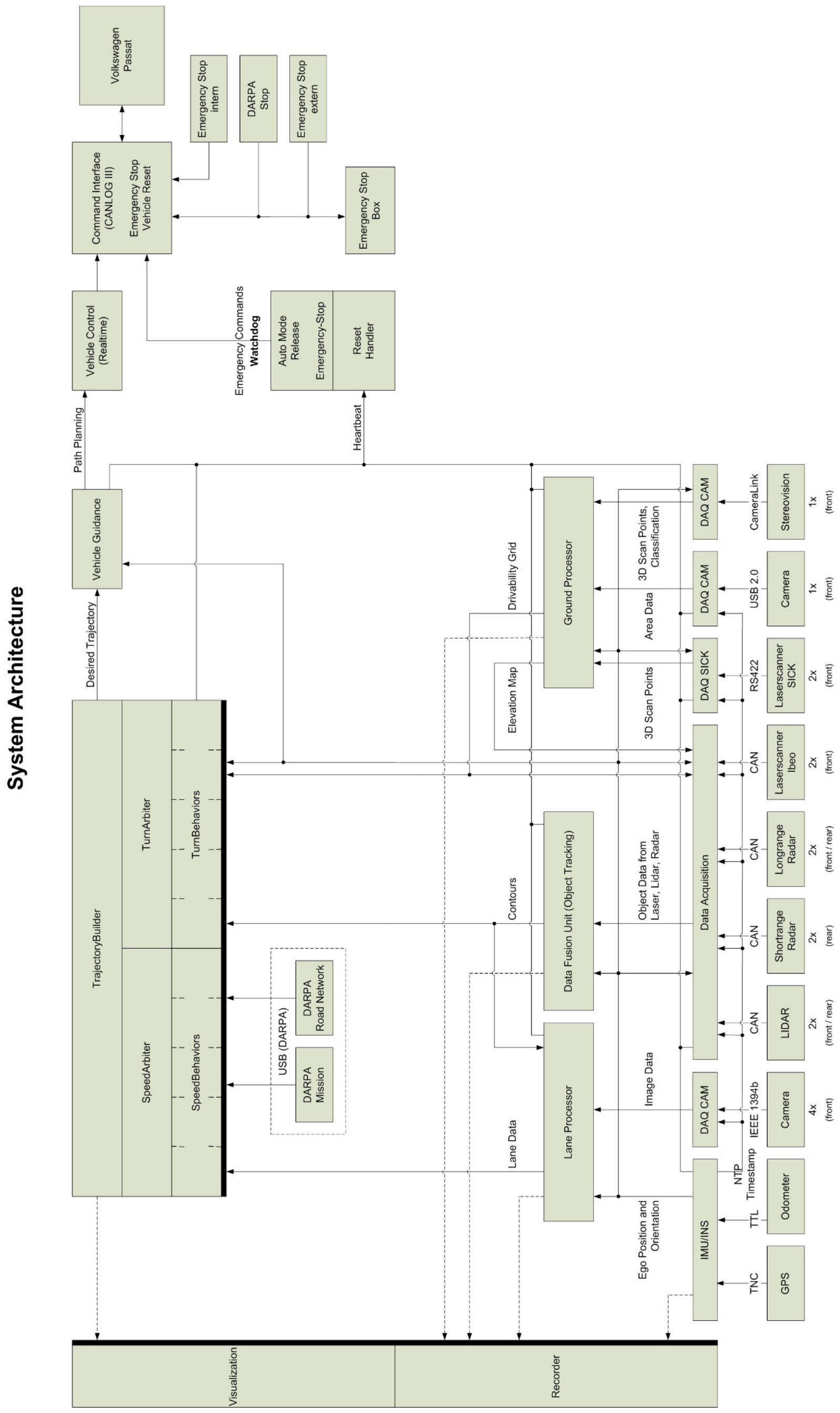

**Fig. 5.** System architecture.

Incoming video data is sampled from the assigned IEEE 1394 interface, preprocessed and interpreted directly on the image acquisition PCs to avoid overload of the vehicle's internal network by image data. Lane detection data is directly passed to the artificial intelligence. The stereo vision system delivers 3D scan points along with area data describing the drivability of the road. This data is fused with further scan points obtained from the laser scanners and area data from the additional color camera observing the ground in front of the vehicle. This fusion results in a drivability grid which is sent to the artificial intelligence module.

Furthermore, following Caroline's signal flow, sensor data of all object-recognition sensors is processed within a central sensor data fusion unit as described in section 4.1, which transmits the object-based vehicle's surroundings containing all static and dynamic targets in Carolines field of view to the digital map. The digital map combines online environmental information with available offline information generated from mission definition files (MDF) and route network definition files (RNDF) provided for the mission. This combined data is the basis for the artificial intelligence to generate driving decisions based on a Distributed Architecture for Mobile Navigation scheme (DAMN) as proposed by [Rosenblatt, 1997] and described in section 4.3.

The driving commands obtained, e.g. "follow a given road" are issued to the soft real time control module, which carries out trajectory generation and optimization based on driving dynamics of the vehicle. The driving trajectories generated are then passed along into hard real time control that addresses the vehicle actuators.

All modules previously described are supervised by a central watchdog process with the possibility to kill and restart one or several processes, computers or sensors independently. Thus, a maximum of self-healing capability is installed in Caroline's systems.

The visualization module is used during development in order to display all exchanged object data. This data consists of e.g. obstacles, lanes, terrain drivability, the planned path and mission data files. A recorder and a player module which logs data for the purpose of offline-analysis, are also integrated in this module.

## 4 System Modules

Caroline's software system consists of five modules. Tasks to be mastered in order to compete in the 2007 DARPA Urban Challenge are environment recognition, road finding, situation assessment and vehicle control supervised by a safety module. These core modules are described below.

## 4.1 Sensor Fusion

Perception is one of Caroline's key systems. The system detects obstacles as well as the drivability of the environment. The sensor fusion system is separated in two parts. The first one is responsible for obstables, such as other cars, walls or pedestrians. The other one takes care of the drivability of the environment. Thus, it is possible to keep the car on the road even in rough evironments. Based on this information, the artificial inteligence is able to find a safe path through traffic. The perception system will be described in greater detail in the following sections. The following section introduces the sensor concept, followed by the object-based data fusion and end with the grid based fusion of the drivability.

### 4.1.1 Sensor Concept

A variety of sensor types originating from the field of driver assistance systems were chosen to provide detection of static and dynamic obstacles in the vehicle's surroundings as depicted in Fig. 3:

- Dark green: A stationary beam LIDAR sensor placed in the front and rear of the vehicle, have a range of approximately 200 meters with an opening angle of 12 degrees. The unit has an internal preprocessing stage and thus delivers its readings in an object oriented fashion, providing target distance, target width and relative target velocity with respect to the car's fixed sensor coordinate frame.
- Red: 24 GHz radar sensors were added to the front, rear, rear left and right side of the vehicle. While the center front and rear sensors provide a detection range of approximately 150 meters with an opening angle of 40 degrees, the rear right and left sensors function as blind-spot detectors with a range of 15 meters and an opening angle of 140 degrees due to their specific antenna structure. The front sensor acts as a stand-alone unit delivering object-oriented target data, such as position and velocity through its assigned external control unit (ECU). The three radar sensors in the rear section operate as a combined sensor cluster using an additional ECU, providing object-oriented target data in the same fashion as the front system. From the perspective of the post processing fusion system, the three rear sensors can therefore be regarded as one unit.
- Blue: Two Ibeo ALASCA XT laser scanners were installed in the vehicle's front section, each providing an opening angle of 240 degrees with a detection range of approximately 60 meters. The raw measurement data of both front laser scanners is preprocessed and fused on their corresponding ECU, delivering complex object-oriented target descriptions consisting of target contour information, target velocity and additional classification information. Additionally, the raw scan data of both laser scanners can be read by the fusion system's grid-based subsection.

- Purple: One Ibeo ML laser scanner was added to the rear side, providing similar detection capabilities as the two front sensors, with a reduced opening angle of 180 degrees due to its mounting position. All Ibeo sensors are based on a four-plane scanning principle with a vertical opening angle of 3.2 degrees between the top and bottom scan plane. This opening angle enables smaller pitch movements of the vehicle to be covered.
- Green: Two SICK LMS-291 laser scanners were mounted on the vehicle's front roof section. These scanners are based on a single-plane technology. They were set to measure the terrain profile at 10 and 20 meters, respectively. The view angle was limited to 120 degrees by software.
- Light blue: A stereo vision system mounted behind the vehicle's front window covers an area of approximately 60 degrees within a range of 50 meters, providing 3-dimensional terrain profile data for all stereo vision points retrieved. A simple classification into the driveway, curb and obstacles classes is also available.
- Orange: A USB-based color mono camera installed on the front roof section, covering an opening angle of approximately 60 degrees.

The sensors view areas are shown in Fig. 6. These view areas overlap for a more reliable view of the environment.

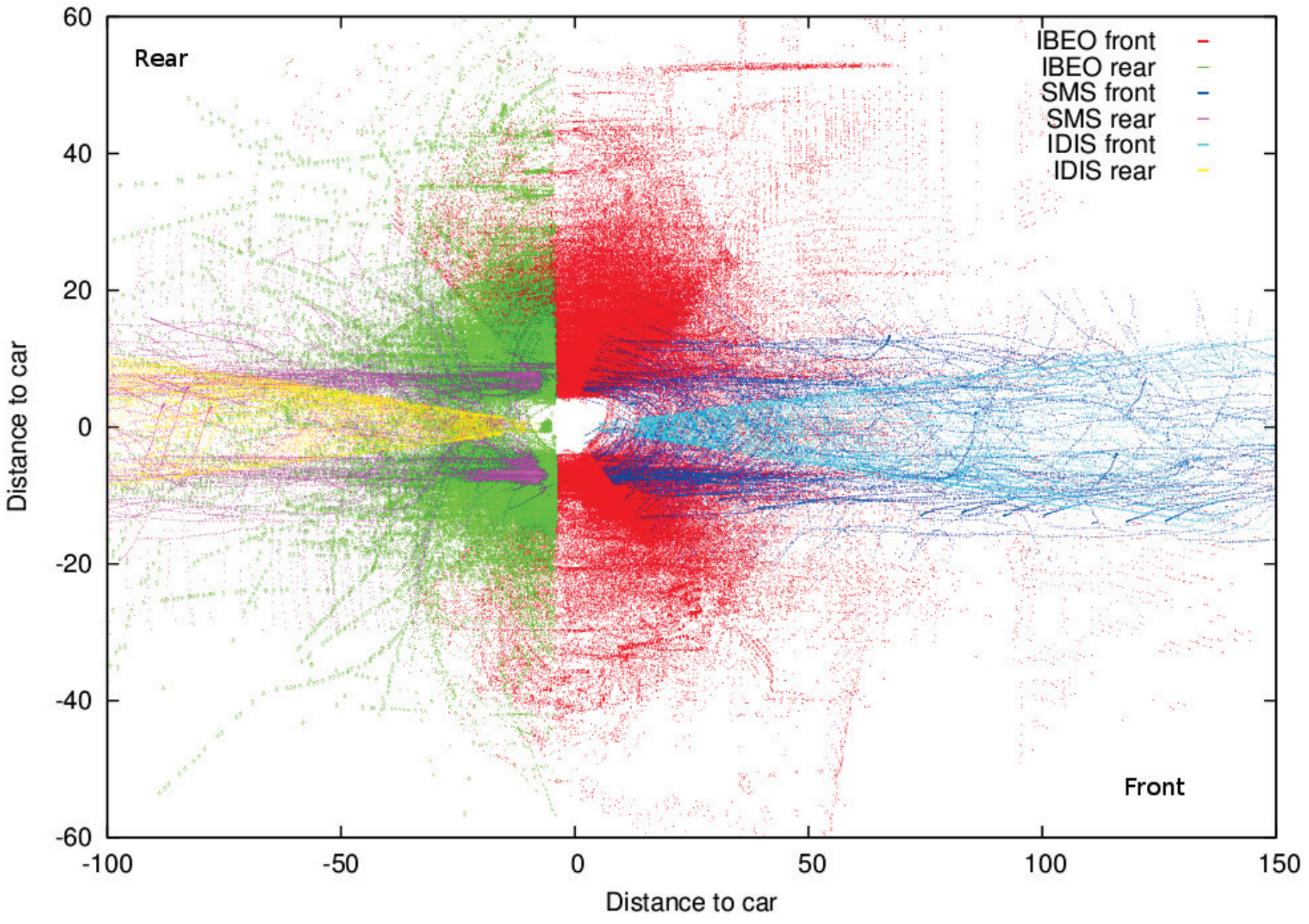

**Fig. 6.** Sensor view areas.

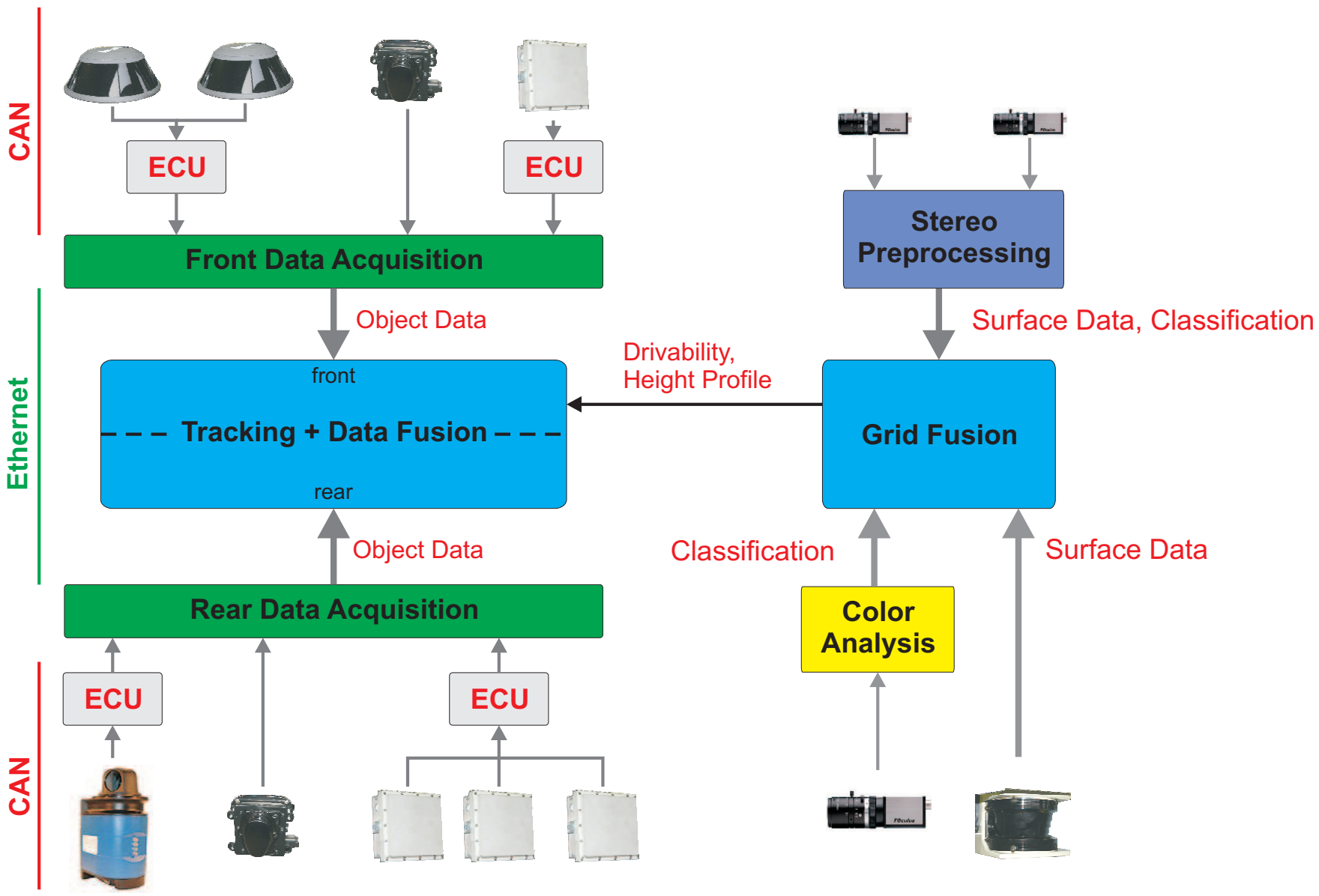

**Fig. 7.** Fusion architecture.

The sensor architecture described reflects the hybrid post-processing scheme applied. While the first four sensors deliver their data in an object-oriented fashion and are therefore treated within the system's object tracking and data fusion stage, the three last sensors described are evaluated based on their raw measurement data in the grid-based subsection. A distributed data fusion system consisting of three interconnected units was set up. In order to equally balance the available computing power, the object tracking system was split into two independent modules, covering the front and rear sections independently. Therefore, two automotive computers carry out data acquisition and data fusion of the front and rear object detecting sensors, while the third PC is used to fuse the raw sensor readings of the SICK scanners, stereo vision system and mono color camera as shown in Fig. 7.

#### 4.1.2 Object Tracking Fusion

The object fusion system is based on a pipes and filters pattern as depicted in Fig. 8. All incoming sensor data is queued and then processed sequentially using a first in - first out strategy. Within the first step, data association is carried out in order to assign incoming sensor objects to their corresponding tracks in the fusion system that are taken from a real-time track database. In case of a positive match between an existing track and incoming sensor object, this pairing is then pushed into the processing queue of the system's Extended Kalman Filter in order to correct the track with new measurement data. If no match can be found, the sensor object is regarded as a potential new target and pushed into the pretracking system. Within pretracking, sensor data is justified against time and all other sensors taking into account sensor redundancy where applicable. Pretracking and data association will be described later in greater detail.

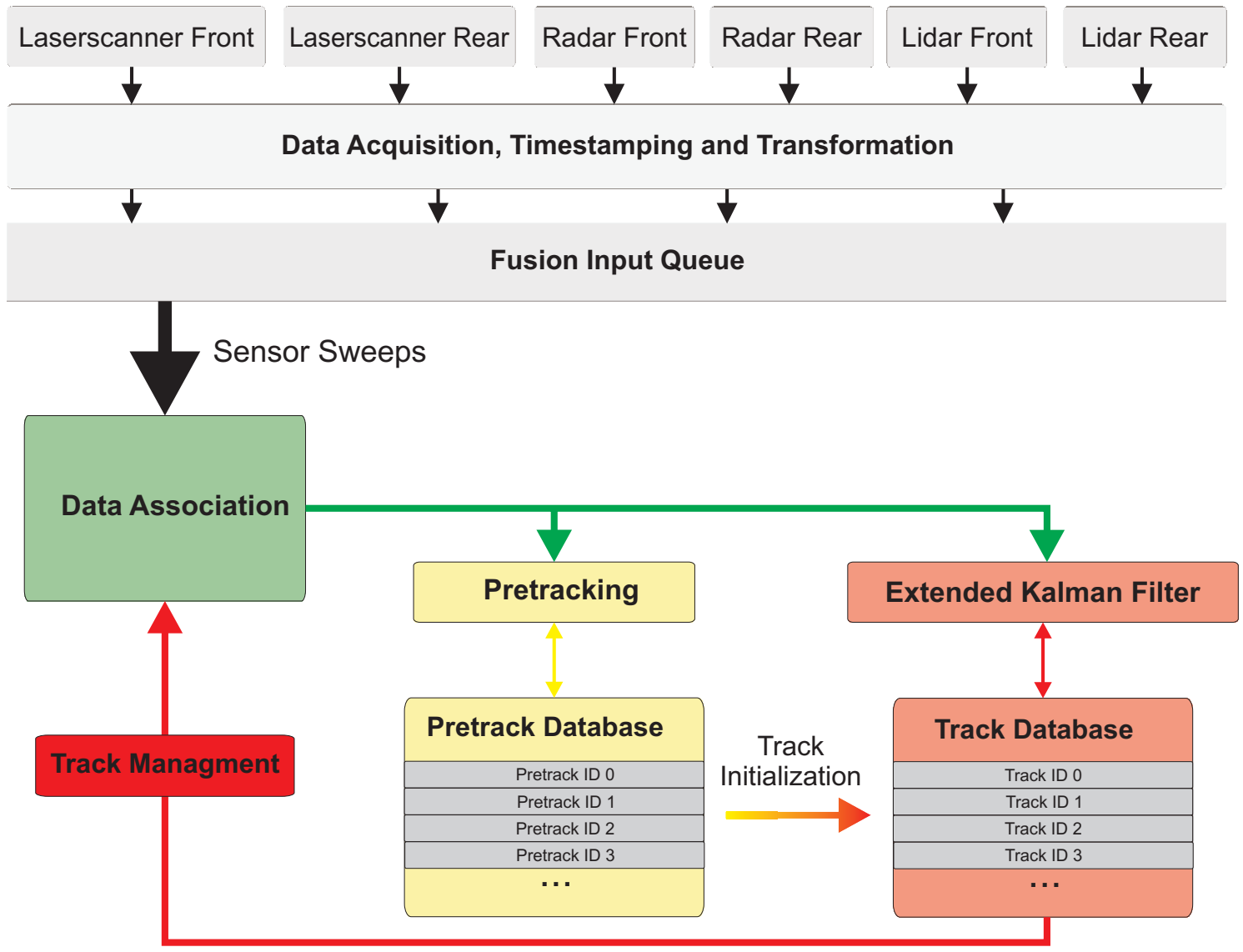

**Fig. 8.** Object fusion system architecture.

If a sensor object has reached a certain level of justification, a new track will be instantiated and pushed into the real-time track database. Parallel to data association, pretracking and final object tracking, a track management unit periodically scans the track database for "dead" tracks - i.e. fusion objects that have not been updated for a certain amount of time. In addition to this garbage collection, all valid tracks are compared to each other for track merging and track splitting, which is necessary to handle situations including a passenger entering or leaving his vehicle or any other situation where two objects in the real world converge or split. Instead of transferring a whole track database image to downstream modules, create, update and delete messages of the track database are issued via the network. Every client is then capable of maintaining it's own track database. Therefore, network load can be significantly reduced without any loss of information.

**Data Association and Pretracking.** Data association and pretracking have a key functionality within Caroline's fusion system. Imperfect data association leads inevitably to incorrect tracks, whereas incorrect track initialization during pretracking leads to imperfect data association, since correct tracks and false alarms will then compete for incoming measurement data.

With this central position, the association and pretracking stage dominates the state estimator in the main tracking stage, since no state estimator can transform falsely associated sensor readings into useful update information for a track. In classical tracking approaches where objects are mostly described through a state vector consisting of a generalized object position, velocity and, if applicable, further derivatives of these quantities, data association can be performed in a point-to-point matching process.

Within Caroline's fusion system, these approaches had to be extended in order to handle spread objects with complex shapes. Three different types of sensor objects have to be processed: complex contours delivered by laser scanners, line-shaped objects delivered by the LIDAR system and classical point-shaped objects received from radar sensors. It is not possible to define a common general object position seen by all sensors, since each sensor will most likely see the target differently. For example the point of reflection delivered from a radar is unknown compared to precise contour measurements gained from a laser scanner. Additionally, as the vehicle moves through the real world, the point of reflection of each sensor type moves on the outline of a real-world object. Therefore a multi-point track model was chosen, describing a detected object by an arbitrary number of contour points and postulating a common movement vector following a rigid body assumption. This way each contour measurement can be matched to the tracked contour point with the best fit. A two-staged data association process was set up, with the first stage serving as a justification as to whether or not track and measurement describe the same real-world object and in the second stage then calculating the optimal contour association between measured and tracked object points. Within stage one, a weighting function counting for the minimum Euclidian distance and similarity of velocities is calculated,

$$w_{i,j} = a \cdot min[|x_k^i - x_l^j|, \forall k, l] + b \cdot |v_i - v_j| \tag{1}$$

with $w_{i,j}$ being a scalar weight for association between track $i$ and measurement $j$ with tracked and measured velocity vectors $v_i$, $v_j$, $x_k^i$, $x_l^j$ being the $k^{th}$ and $l^{th}$ contour point position of track $i$ and measurement $j$ and $a$, $b$ serve as tuning parameters. A threshold for this weight is further defined and an association below that threshold level will be pushed into stage two.

In stage two, an optimal match between all measured and tracked contour points is calculated based on an association matrix ,

$$\Omega = \begin{bmatrix} |x_1^i - x_1^j| \dots |x_1^i - x_l^j| \\ \dots \quad \dots \quad \dots \\ |x_k^i - x_1^j| \dots |x_k^i - x_l^j| \end{bmatrix} \tag{2}$$

Optimization can be carried out with standard algorithms such as the Hungarian/Munkres method, Nearest Neighbor or similar approaches. We used the fast Minimum-algorithm. This two-staged association process avoids unnecessary computational load on the system, since unlikely associations will be filtered out in stage one while the computational challenging minimization is only carried out for positive matches.

During pretracking, incoming sensor data is first associated with preliminary track objects (pretracks) using methods similar to those described above. A pretrack carries along a vector of sensor assignments, storing for each sensor type the last assigned sensor object id. A simple Kalman filter based on a constant velocity motion model is calculated for each pretrack to update its position given by incoming sensor data. In addition to the vector of sensor assignments, an update counter is carried along storing the number of positive association events. Taking into account sensor redundancies read from a configuration file, a threshold for track activation is evaluated based on this update counter, depending on the level of redundancy in the affected observation area of that object. A simple description language was implemented to efficiently model these redundancies and to influence the update count threshold for track activation, e.g.:

```
polygon={0,2;10,2;10,-2;0,-2}
modifyCount=2000
condition=( RADARFront && !( LASERFront || LIDARFront ) ),
```

which means for the fusion system "Activate track in a 2 x 10 meter, box-shaped view area after 2000 positive matches when it is only seen by the front radar system and not by the laser scanners or LIDAR sensors", which, in this case, serves as protection against random, unstable false alarms from the radar sensor directly in front of the vehicle.

**Tracking and Data Fusion.** For the main tracking algorithm, a model-switching Extended Kalman Filter, based on two track motion models was implemented. A six-dimensional motion model describes fast-moving objects using a state vector,

$$x^{6D} = \begin{pmatrix} x_{1...n} \\ y_{1...n} \\ v \\ a \\ \alpha \\ \omega \end{pmatrix} \tag{3}$$

with $x_{1...n}$ and $y_{1...n}$ being the x and y coordinate of the n contour points, the common velocity, acceleration, course angle and course angle velocity with respect to the global earth-fixed reference frame. For slow or static objects, a simpler four-dimensional state vector was chosen,

$$x^{4D} = \begin{pmatrix} x_{1...n} \\ y_{1...n} \\ v \\ a \end{pmatrix} \tag{4}$$

thus taking into account that the majority of detected objects are of a rather static nature and distribution of available sensor information in unnecessary many state variables is suboptimal in that case. As seen in equations (4) and (5), the classical state vector has been enriched by the number of contour points, thus making it necessary to extend the Kalman Filter algorithm (see [Kalman, 1960] for reference) to handle multiple positions within the same state vector. Similarly, we define the sensor measurement vector for a sensor object consisting of m contour points,

$$y = \begin{pmatrix} x_{1 \ldots m} \\ y_{1 \ldots m} \\ v_x \\ v_y \end{pmatrix} \tag{5}$$

with $x_1$, $y_1$, $v_x$, $v_y$ being measured contour point $x$- and $y$-coordinates as well as $x$- and $y$-velocity components with respect to the global earth fixed reference frame. Postulating a common position noise covariance for all contour points within track and measurement, the update algorithm can be extended as follows:

$$\begin{aligned} x_k(v+1|v) &= f(x_k(v)) \\ P(v+1|v) &= F^T \cdot P \cdot F + Q \\ s_{k,l} &= y_l(v+1) - h(x_k(v+1|v)) \\ S(v+1) &= H \cdot P(v+1|v) \cdot H^T + R \\ K(v+1) &= P(v+1|v) \cdot H^T \cdot S(v+1)^{-1} \\ r_{k,l}(v+1) &= K(v+1) \cdot s_{k,l}(v+1) \end{aligned} \tag{6}$$

with $x_k$ being the track state vector regarding contour point $k$, $f(x)$ the nonlinear system transfer function, $P$ the common state covariance matrix, $F$ the system transfer Jacobian, $Q$ the system noise covariance, $s_{k,l}$ the innovation vector of tracked contour point $k$ compared with measured point $l$ of the associated sensor object, $y_l$ the sensor measurement vector regarding measured point $l$, $h(x)$ the nonlinear system output function, $S$ the common innovation covariance matrix, $H$ the system output Jacobian, $R$ the estimated measurement noise, $K$ the Kalman gain in this update cycle and $r_{k,l}$ the correction vector for tracked contour point $k$ getting updated with measured point $l$.

The tracked contour points can then be updated by adding the first two components of the associated vector $r_{k,l}$. In order to calculate updated common velocity, acceleration, course angle and course angle velocity in the six dimensional movement model, the mean value for vector $r_{k,l}$ is calculated over all given contour point associations,

$$r_{mean} = \frac{1}{N} \sum_{k,l=1}^{N} r_{k,l} \tag{7}$$

with $N$ being the total number of acquired contour point matches within the second stage of data association. Corrected common values can then be acquired by adding the last four components of vector $r_{mean}$ to the corresponding elements in the track state vector.

Obviously, by postulating a common system and measurement noise covariance for all contour points, Kalman gain can be computed once per update cycle. While it would theoretically be possible to calculate a separate Kalman gain for each tracked contour point and therefore removing the limitations to system and measurement covariance, this would lead to a N-times bigger computational load, since matrix inversion of the system innovation covariance matrix is the most costly part of the algorithm. In this case, the algorithm would simply calculate a separate Kalman filter for each contour point, which is not practically realizable in a real-time application. In the approach described we have no significantly higher computational effort compared to a standard EKF, while at the same time realizing spread-contour functionality and removing the need for a stable point of reference for tracked objects.

In order to prevent the track from being flooded with contour points, a garbage collection mechanism was installed by carrying along update counters for each contour point, which stores the last update timestamp and the overall number of updates counted to that point in time. In this manner, inactive contour points can be detected easily and removed from the track's point list.

**Object splitting and termination.** Because of the track's polyline object model, it is necessary to implement a track splitting algorithm. If there is no such method, one track can collect points from many objects and grow to a rather huge but meaningless track. For example, a person dropping off a car and moveing away would still be part of the car track because of the data association algorithm depiced in Fig. 9. When the person just dropped off and is still near the car, it will become one track. After moving away from the car, the contour points will still be updated because there is an object at the position of the car and the person is also still there. Between these two objects there is nothing but the polyline from the track still describing an outline of an object.

To detect these false tracks, an algorithm was developed to split such tracks. The basic idea is to examine the objects based on the raw sensor object data and find indepented sets of objects. These independent sets will become the new tracks. Normally, there are no such sets but in the event of an unsplit track, there are two or more partitions. Polygonal objects around the track will be described as a planar undirected two colored graph. The algorithm contains the following steps:

1. Build planar undirected colorable rectangular graph. set the color of every node to black.
2. Set the polylines of every sensor object of the track to white.
3. Search for independend sets in the graph[Cormen et al., 2002].
4. If there is more than one set found, build new tracks from the points describing the outlines.

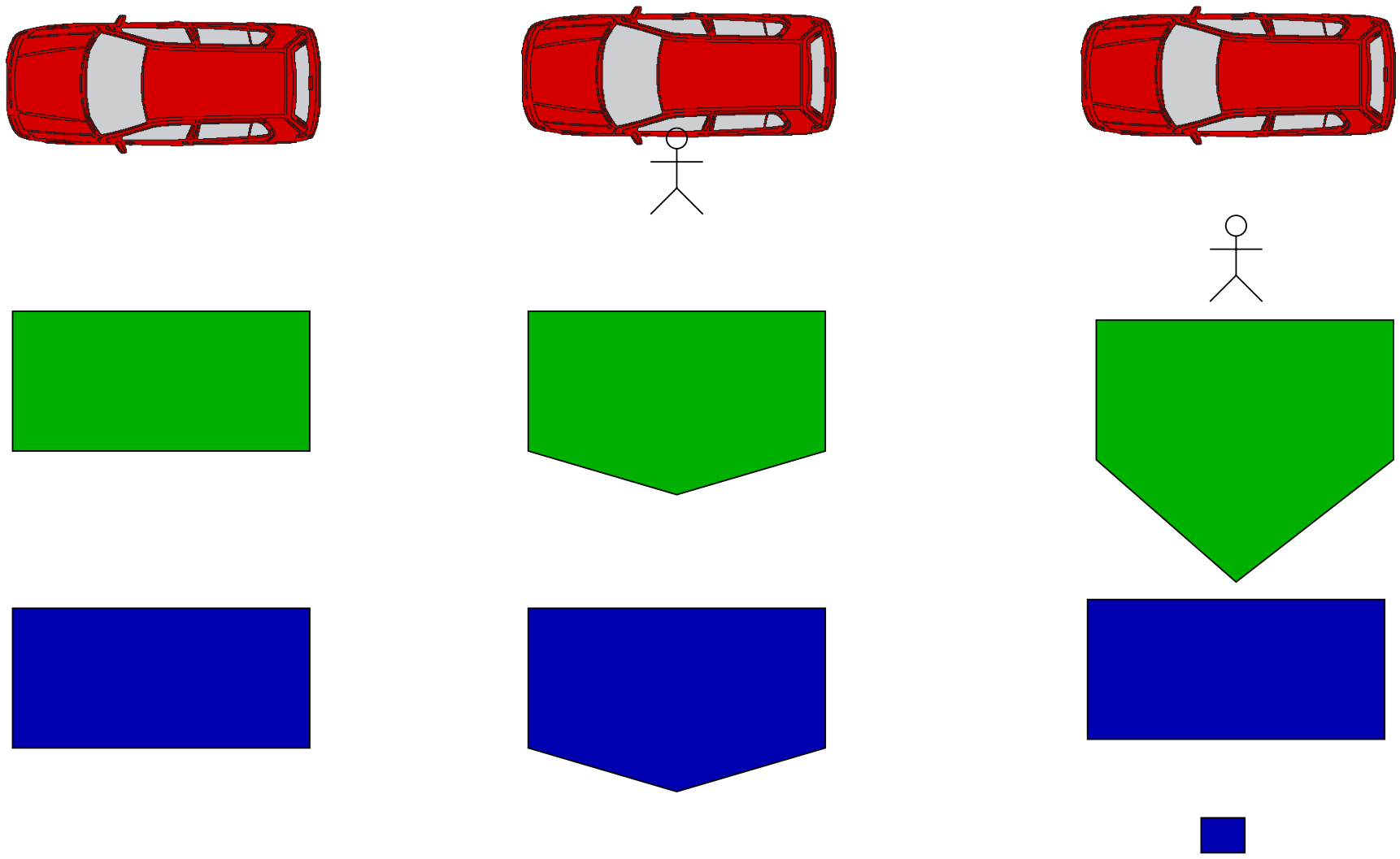

**Fig. 9.** Person who drops off a car. From left to right: Person still in the car, person just dropped off, person moves away. From top to bottom: Reality, track without splitting, track after splitting.

The algorithm runs periodically during track garbage collection. Although complexity depends on the maximal area ($a$) covered by a track ($O(a)$), this algorithm can be implemented efficiently with graphic libraries.

#### 4.1.3 Grid-Based Fusion

In contrast to the object tracking subsystem, the grid fusion system does not describe agents in the vehicle's environment with discrete state vectors, but instead discretizes the whole environment into a rectangular matrix (grid) structure. Each grid cell carries a number of assigned features:

- a height value expressed in the global earth fixed reference frame,
- a gradient value describing the height difference to neighboring cells,
- a set of Dempster-Shafer probability masses counting for the hypotheses undrivable, drivable and unknown,
- a status flag stating whether or not measurement data has been stored within the corresponding cell and
- an update time stamp storing the last time a cell update was carried out.

**Data Structure.** The biggest challenge with grid based models in an automotive environment is the need for real time operation. High maneuvering speeds in automotive applications require update rates greater than 10 Hz, which is almost too low since this equals a travel distance of 1.4 meters at normal urban speeds. The approach of discretizing the environment into grid cells leads to significantly high memory requirements and therefore calls for efficient data structures. As an example, the storage of a view area of only 100 x 100 meters with a resolution of 25 centimeters generates 160,000 grid cells. Assuming a 4-byte floating point value for each feature as described above, this grid extends up to 3 MByte. Together with an update rate of 10 Hz this leads to a constant data throughput of 256 MBit/s, which in any case is more than the standard automotive bus infrastructure would be able to handle. Efficient algorithms and data reduction prior to serialization is therefore the key to a successful application. For addressing these issues we implemented a rolling grid data structure wherein the vehicle's own position is a pointer to the corresponding grid cell. This position will be regarded as virtual origin for all incoming sensor readings, which can then be subsequently accessed by moving through the double linked data structure relative to that virtual origin. The main grid is again subdivided into sub grids whose size match the processor's caching mechanism for optimal usage of the available computing resources. While it would theoretically be possible to make the surface large enough to cover the expected maneuvering area, this would lead to extremely high memory usage and is therefore not feasible. Instead, when the vehicle moves through the world, the reference point shifts along the double linked spherical list. As soon as it crosses the border from one sub grid to the next, the corresponding sub grids at the new horizon of the data structure are cleared and are therefore available for new data storage.

**Treatment of laser and stereo vision point data.** As the first step within grid data fusion, the 3-dimensional point clouds received from the laser scanners and stereo vision system are transformed into the global earth fixed reference frame taking vehicle attitude into account (roll, pitch and yaw) as acquired from the GPS/INS unit and sensor-specific calibration information. The accuracy of these transformations is crucial to subsequent post-processing. Vehicle height as delivered by the GPS is especially important and is therefore subject to further filtering and justification. For each measured point, the corresponding grid cell is retrieved and a ray tracing algorithm (Bresenham) is carried out to update all cells from the sensor coordinate system's origin to the measured target point. Several versions of the Bresenham algorithm are described in the literature, in this case we will introduce the 2D version following Pitteway [Pitteway and M.L.V., 1967] for reasons of simplicity.

The lines are traced similar to the functionality of a plotter, which is basically the origin of that algorithm. On the way through the traced lines, each cell passed is updated according to following rules:

- If the cell lies on the path between sensor origin and measured target point and it's height value exceeds the current Bresenham line height value, reduce the stored value to that of the current Bresenham line.

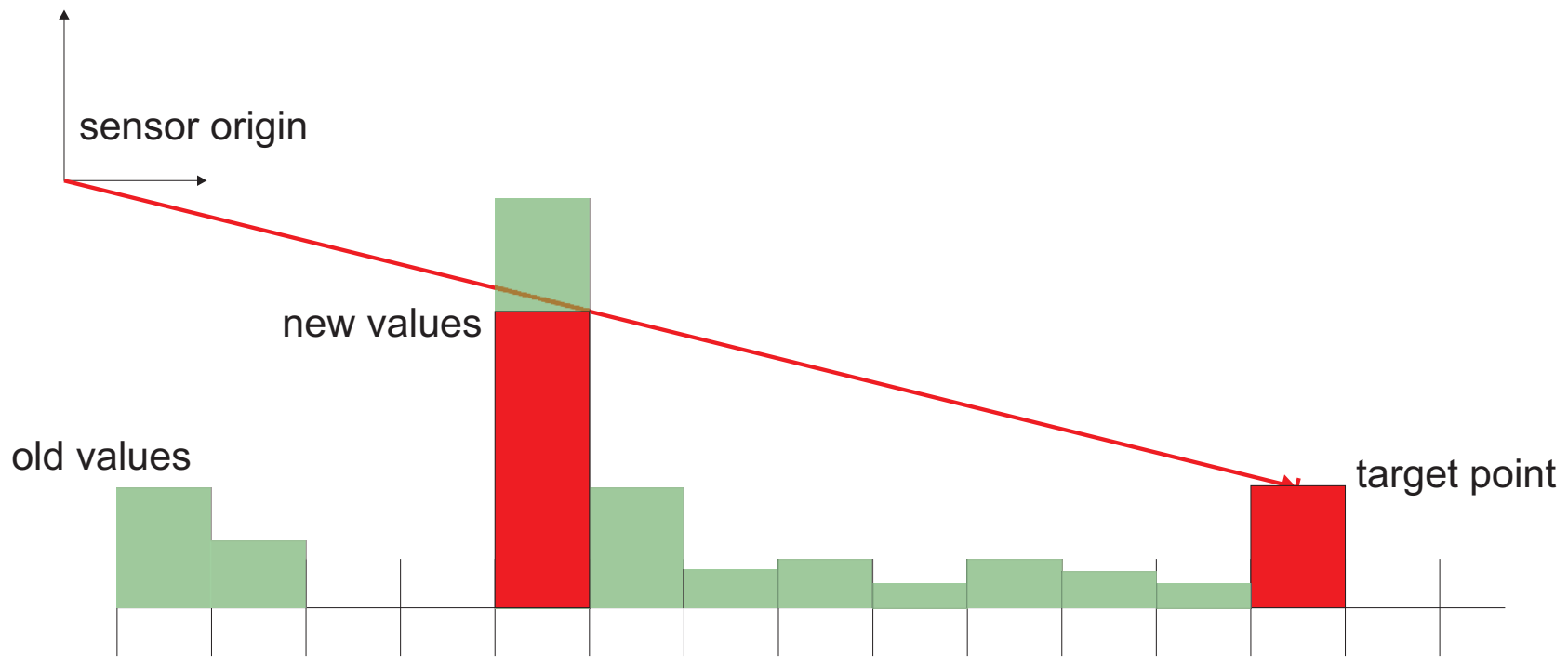

**Fig. 10.** Ray update mechanism.

- If the cell is the end point of the Bresenham line, store the associated height value.
- In both cases, store update time stamp and mark that cell as having been measured.

The grid is updated following the direct optical travel path of any laser ray (or virtual stereo vision ray) starting at the sensor origin and ending at the target point as depicted in Fig. 10. This model follows the assumption that any obstacle would block the passing optical ray and therefore any cells on the traveling path must be lower than the ray itself.

**Data Fusion.** Parallel to entering the 3-dimensional point data acquired from laser scanners and stereo vision, vision-based classification is processed using a Dempster Shafer approach [Shafer, 1976, Shafer, 1990]. A sensor model was created for each data source, mapping the sensor specific classification into the Dempster Shafer probability mass set, which can then be fused into the existing cell probability masses using Dempster's rule of combination,

$$m_c^*(A) = m_c(A) \bigoplus m_m(A) = \frac{1}{1-K} \sum_{B \cap C = A \neq \emptyset} m_c(B) m_m(C), \qquad (8)$$

with $m_c$, $m_m$ being the cell and measurement probability mass set and $m_c^*$ the combined new set of masses for the regarded cell, while the placeholders $A$, $B$ and $C$ can describe any of the three hypotheses drivable, undrivable and unknown. The term $K$ expresses the amount of conflict between existing cell data and incoming measurement, with

$$K = \sum_{B \cap C = \emptyset} m_c(B) m_m(C). \qquad (9)$$

The mass set $m_m$ has to be modeled out of the acquired sensor data.

With respect to the stereo vision system, which is capable of classifying retrieved point clouds into the classes road, curb and obstacle, this mapping is trivial and can be done by assigning an appropriate constant mass set to each classification result. The exact values of these masses can then be subject to further tuning in order to trim the fusion system for maximum performance given real sensor data.

In the case of the mono vision system, Caroline assigns each pixel in the retrieved image a drivability value $P_d$ between 0.0 representing undrivable and 1.0 representing fully drivable; a mapping function is then applied, which creates the three desired mass values $D$: drivable, $U$: undrivable and $N$: unknown, that can be either drivable or undrivable as follows:

$$\begin{aligned} m_m(D) &= D_{max} \cdot P_d, \\ m_m(N) &= (1 - D_{max}), \\ m_m(U) &= 1 - m_m(D) - m_m(N). \end{aligned} \tag{10}$$

The value $D_{max}$ will serve as a tuning parameter, influencing the maximum trust placed into the mono vision system and based on the quality of its incoming data. Both, the classification mechanism of the stereo vision system would be beyond the scope of this paper and will therefore not be explained in detail. Basically, classification within the stereo system is based upon generating a mesh height model out of the point cloud obtained and applying adaptive thresholds to this mesh structure in order to characterize roadway, curb and obstacles. The mono vision system is based on a similar approach to [Thrun et al., 2006].

Prior to mapping the mono vision data into the grid data structure, the image must be transformed into the global world reference frame using the known camera calibration [Heikkil and Silvn, 1996] and height information which can easily be retrieved from the grid itself.

The creation of a sensor model for the 3-dimensional height data is more complex: First, a gradient field is calculated from the stored height profile. In Caroline's grid fusion system, the grid is mapped into image space by converting into a grayscale image data structure, with intensity counting for cell height values. Subsequently, the Sobel operator is applied in both image directions.The results of both convolutions are summed and - after proper normalization - transformed back into the grid structure, storing the gradient values $\frac{\partial h}{\partial x \partial y}$ for each grid cell. Any existing obstacle will usually lead to a bigger peak within the gradient field, which can easily be detected. During the process of forward and reverse transformation, the grid structure in- and out of a grayscale image would initially appear to be redundant, because the gradient operator could easily be applied to the height field itself. Yet, by transforming the information into a commonly used image format, the broad variety of image processing algorithms and operators found in standard image processing toolkits, such as the OpenCV library [OpenCV Website, 2007] can easily be applied, thereby significantly reducing development time.

The acquired gradient values will then subsequently be mapped into a Dempster-Shafer representation, which leads to the desired sensor model combining all acquired height values. Similar to the method with the mono vision system, a mapping function is defined as follows:

$$m_m(D) = \begin{cases} D_{max}, & \left|\frac{\partial h}{\partial x \partial y}\right| \leq G_{D_{max}} \\ 0, & G_{D_{max}} < \left|\frac{\partial h}{\partial x \partial y}\right| \leq G_{U_{min}} \\ 0, & \left|\frac{\partial h}{\partial x \partial y}\right| > G_{U_{min}} \end{cases}$$

$$m_m(U) = \begin{cases} 0, & \left|\frac{\partial h}{\partial x \partial y}\right| \leq G_{D_{max}} \\ \frac{U_{max}}{G_{U_{min}} - G_{D_{max}}} \cdot \left(\left|\frac{\partial h}{\partial x \partial y}\right| - G_{D_{max}}\right), & G_{D_{max}} < \left|\frac{\partial h}{\partial x \partial y}\right| \leq G_{U_{min}} \\ U_{max}, & \left|\frac{\partial h}{\partial x \partial y}\right| > G_{U_{min}} \end{cases}$$

$$m_m(N) = 1 - m_m(D) - m_m(U), \tag{11}$$

with $D_{max}$ and $U_{max}$ serving as parameters for maximum drivability/ undrivability assigned to the gradient field, $G_{D_{max}}$ being the maximum gradient value that is still considered to be fully drivable and $G_{U_{min}}$ the minimum gradient value that is considered to be fully undrivable. By carefully tuning those four parameters, it is possible to suppress unwanted smaller gradients resulting e.g. from unimportant depressions and knolls in the road while supporting higher gradients as originating from curbs or berms in order to correctly fuse this information into the grid cells by using Eq. 8.

## 4.2 Vision

Caroline's computer vision system consists of two separate systems. The first is a monocular color segmentation based system that classifies the ground in front of the car as drivable, undrivable or unknown. It assists in situations where the drivable terrain and the surrounding area (e.g. grass, concrete or shrubs) differ in color. The output of this algorithm contributes to the Grid Based Fusion as described in section 4.1.3. The second vision system is a multi-view lane detection that identifies the different kinds of lanes described by DARPA, such as broken and continuous as well as white and yellow lane markings. Using four high-resolution color cameras and state-of-the-art graphics hardware, it detects its own lane and the two adjacent lanes to the left and right with a field of view of 175 degrees at up to 35 meters. The output of the lane detection algorithm is directly processed by the artificial intelligence.

### 4.2.1 Lane Detection

Detecting lane markings on roads in an urban environment is a difficult but very important task. While concepts exist that depend on additional markings, such as magnetic bands in the street, a more useful method must make intelligent use of what is available on today's roads. Towards this goal, we developed a lane detection system that is capable of analyzing several high-resolution images simultaneously and in real-time. Our lane fitting algorithm uses a very versatile lane model and is robust with respect to outliers and artifacts. It also takes into account lane markings of adjacent lanes. It copes with different road setups, lane markings and lighting situations. The lane detection process is divided into four parts, as shown in Fig. 11. First, the raw images are downloaded from the cameras via the IEEE1394b interface. Second, they are uploaded to graphics hardware, the color information is retrieved from the raw Bayer pattern, and the images are transformed into a single top view perspective, Fig. 12. Third, lane marking features are detected in the image, Fig. 14. In the last step, a lane model is adjusted to match the features detected.

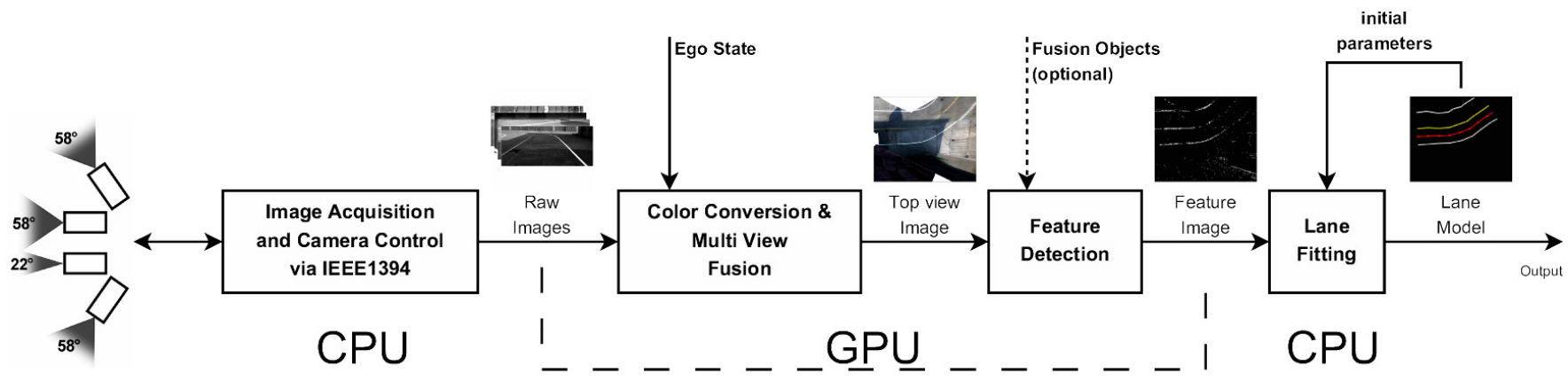

**Fig. 11.** The four stages of the lane detection algorithm.

**Data Acquisition.** Three cameras with field of view of 58 degrees cover the area in front of the car. A 22 degrees telephoto lens camera provides a high-resolution view of the street ahead of the car. The four 1376x600 8-bit raw Bayer images are synchronously acquired via the IEEE1394b interface at 14 frames per second. The images are uploaded to the graphics card and converted to the RGB color space using bilinear interpolation. As the lane fitting algorithm works in a global coordinate system, the position and rotation of the vehicle, also referred to as Ego State, must be available. A transformation function $f_{ego} : p_{car} \mapsto p_{world}$ can be defined if the Ego State is known, where $p_{car}$ is a point in the car's reference system, and $p_{world}$ is a point in a global Cartesian reference system. An Inertial Measurement Unit corrected by a GPS signal was used to generate the Ego State.

**Multi-View Fusion.** Because local changes of the light intensity are an indicator for white lines, and local saturation changes indicate colored lane markings, the RGB images are converted to the HSV color space. This color space encodes saturation and color in separate channels. Knowing the intrinsic and extrinsic parameters of the camera, and including the orientation of the vehicle (pitch and roll), a lookup function that converts top view coordinates to image coordinates can be used to create a single HSV top view image. The lookup operation is applied to each source image. In regions where the projected images overlap, precedence $I_{tele} > I_{middle} > I_{left} > I_{right}$ is maintained as shown in Fig. 12. The region of interest covers the area of up to 30 meters in front of the vehicle and 12 meters to the left and right at a scale of 35 pixels per meter.

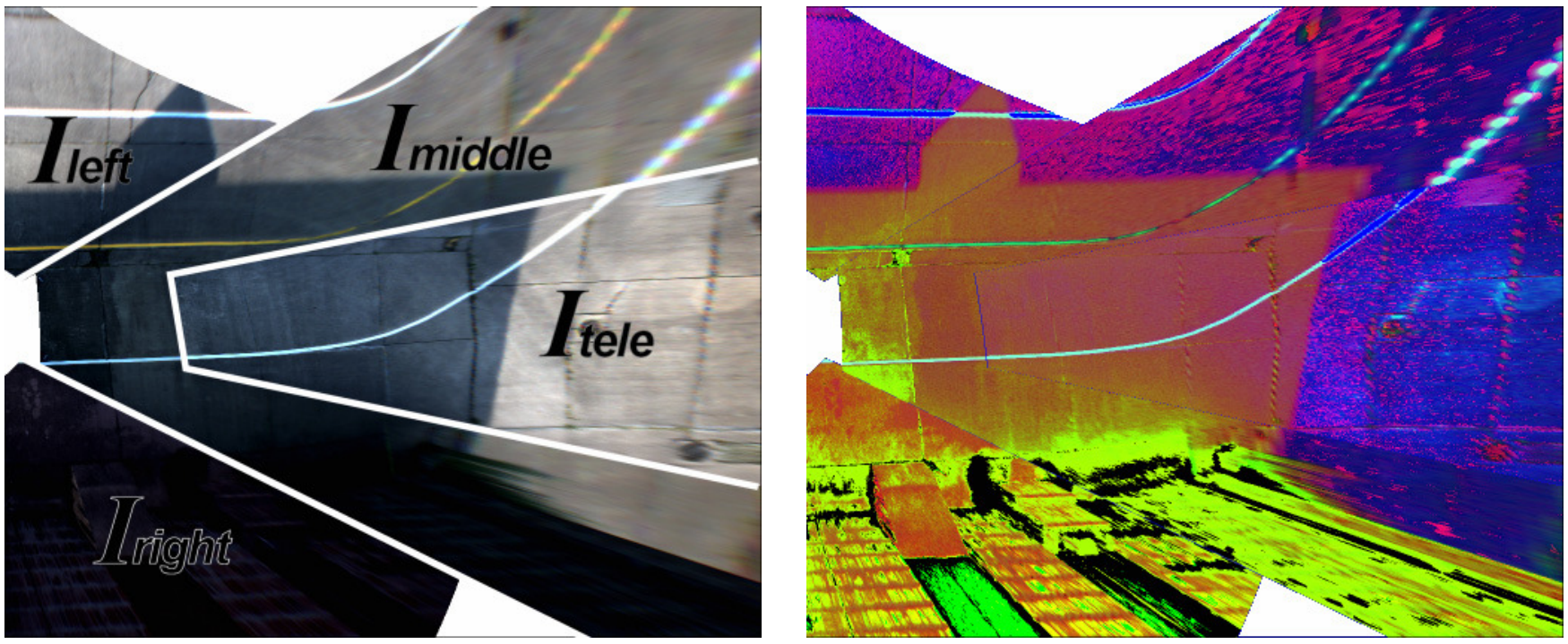

(a) RGB top view

(b) HSV top view

**Fig. 12.** The four different images (a, RGB color space used for visualization) are merged to a single HSV top view image (b).

**Features.** Lane markings can be described as a thin pattern of local differences of the road surface that cover long distances. Therefore, the basic concept underlying feature detection involves identification of these local differences in regions of 8x8-pixels that resemble road patches of approximately 25 by 25 centimeters. Analyzing the HSV top view image, the feature detection's output is a downsampled feature image that encodes the quality, direction and color, i.e., white or yellow, of the lane features in Fig. 14. As lane markings exist in various colors, qualities as well as widths, and appear differently under changing lighting conditions, only few stringent assumptions apply. When analyzing the top view image for features, we check three criteria that must be present:

1. The local contrast $v_{diff}$ must exceed a certain threshold. The local contrast is the difference between the local minimal and maximal value $v_{diff} = v_{max} - v_{min}$.
2. Analyzing a local adaptive histogram, the distance $b_{diff}$ between the two largest bins $b_{high}$ and $b_{low}$ must exceed a certain threshold. This is because it can be assumed that $b_{low}$ contains pixels depicting the street and $b_{high}$ identifies the lane marking.
3. The pixels in $b_{high}$ must have a recognizable shape and orientation. For several discrete orientations, the ratio of the variances of the pixels' x- and y-coordinates is checked.

A detailed description is given in Alg. 1. As this algorithm is prone to discretization errors, supersampling improves the quality of the feature detection.

**Data**: An 8x8 region of a HSV top view image, thresholds $t_{con}, t_{hist}, t_{dir}$ and $t_{col}$

**Result**: A feature quality $q$, direction $a \in \{0, 22.5, ..., 157.5\}$ and color $c \in \{white, yellow, undecided\}$

**for** *the saturation and lightness channel* **do**
    $v_{diff} = v_{max} - v_{min}$; $v_{max}$ and $v_{min}$ are the maximal and minimal values of the current channel
    **if** $v_{diff} < t_{con}$ **then**
        break;
    **end**
    compute adaptive histogram;
    determine two largest bins $b_{high}$ and $b_{low}$, Fig. 13(b) ;
    $b_{diff} = b_{high} - b_{low}$;
    **if** $b_{diff} < t_{hist}$ **then**
        break;
    **end**
    set of pixels $p_{high}$ = pixels in $b_{high}$;
    determine center of mass $R$ of $p_{high}$;
    initialize $r_{max}$ and $a_{max}$ to 0;
    **for** $i = 0; i <= 157.5; i = i + 22.5$ **do**
        rotate $p_{high}$ around $R$ by $i$ degrees. determine ratio of variances $r = \frac{Var(X)}{Var(Y)}$;
    **end**
    **if** $r_{max} < t_{dir}$ **then**
        break;
    **end**
    label this region as a feature;
    **if** *current channel is lightness* **then**
        $q_{white} = b_{diff}$; $a_{white} = a_{max}$
    **else**
        $q_{yellow} = b_{diff}$; $a_{yellow} = a_{max}$
    **end**
**end**
**if** $q_{white} > t_{col}$ *&* $q_{white} > q_{yellow}$ **then**
    $c = white$; $a = a_{white}$
**end**
**if** $q_{yellow} > t_{col}$ *&* $q_{yellow} > q_{white}$ **then**
    $c = yellow$; $a = a_{yellow}$
**end**
$q = max(q_{white}, q_{yellow})$;

**Algorithm 1.** Feature detection algorithm.

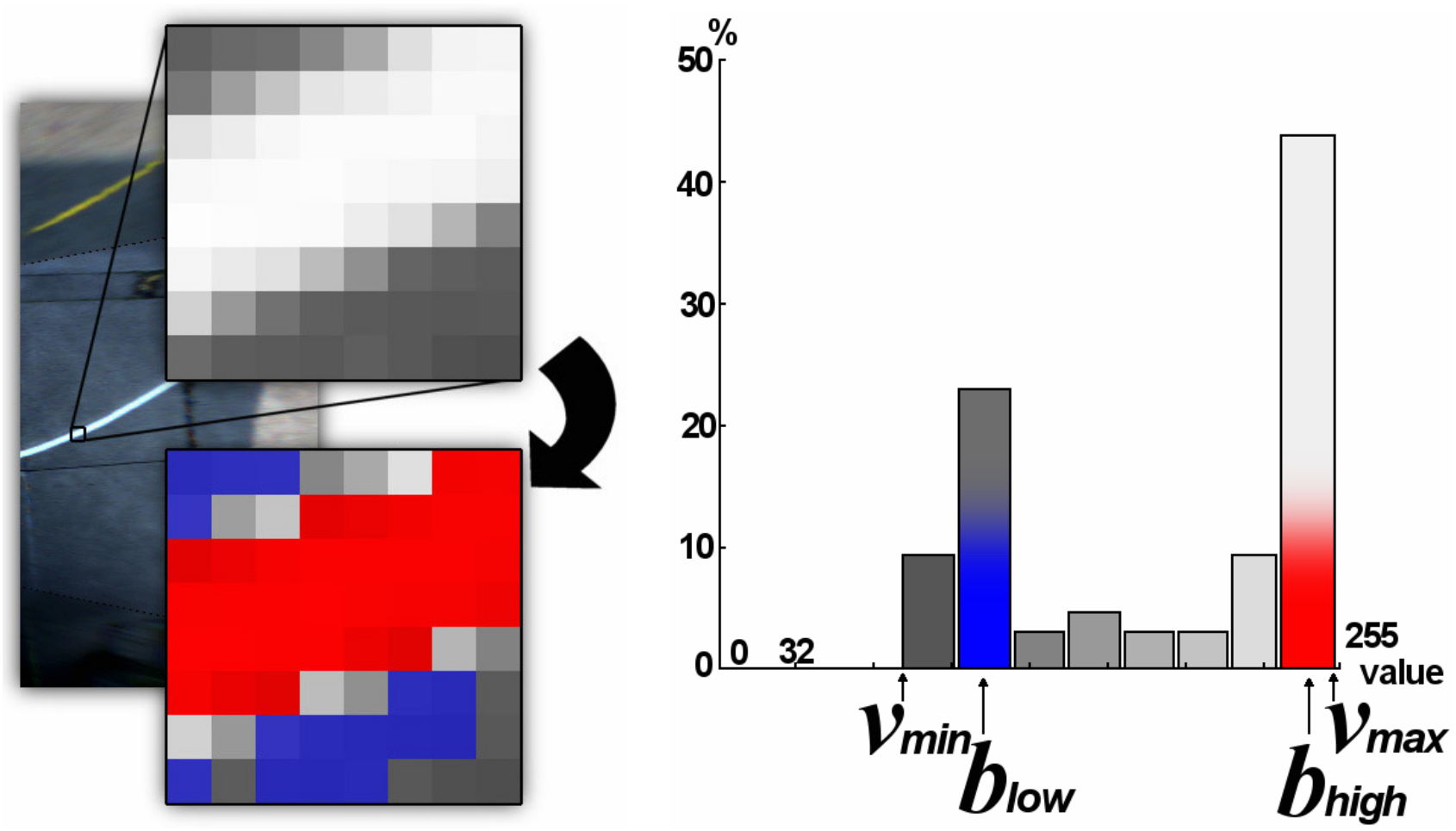

(a) 8x8 regions are analyzed (b) 8 bin histogram of the 8x8 region

**Fig. 13.** 8x8-pixel regions of the top view image (a, up) are tested for possible features. The distance between the two largest bins $b_{low}$ (b, blue) and $b_{high}$ (b, red) of the histogram determines the quality of the feature. The pixels gathered in $b_{high}$ must be arranged in a directed shape (a, red area).

**Lane Model.** The lane model consists of connected lane segments. Each segment $s_i$ is described by a length $l_i$ (given parameter), a width $w_i$ and an angle $d_i = \alpha_i - \alpha_{i-1}$ describing the difference of orientation between this segment and the previous one as shown in Fig. 16. The first segment is initially placed on the current coordinates of the vehicle and facing in the driving direction, assuming that the vehicle is actually located on the street. Knowing the position $c_0$ of the initial segment as well as the lengths $l_i$ and the

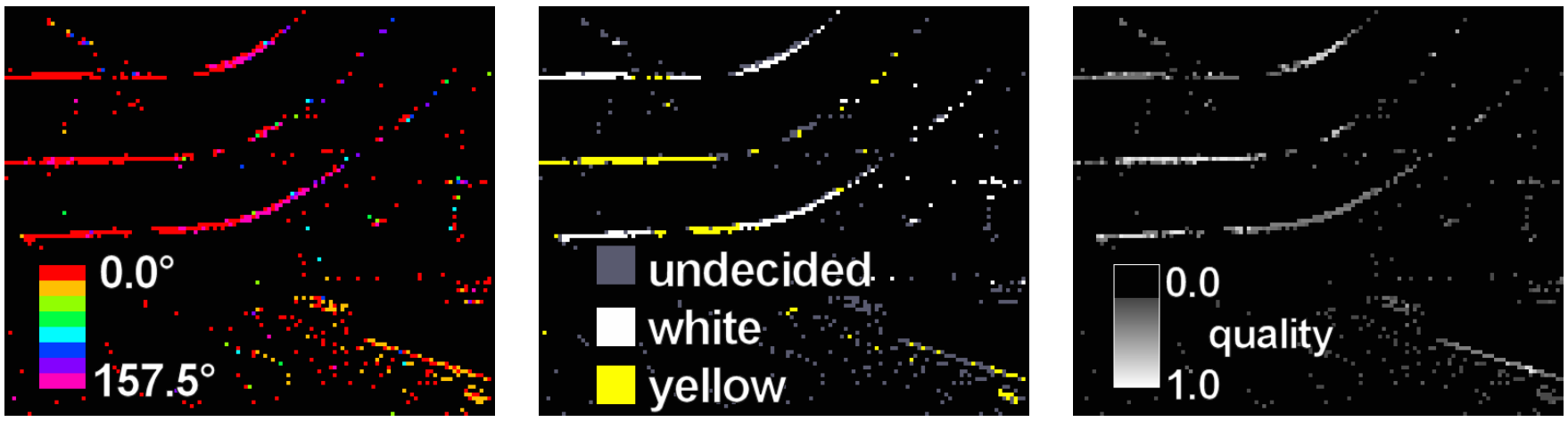

(a) direction information (red channel) (b) color information (green channel) (c) quality information (blue channel)

**Fig. 14.** The direction (a), color (b) and quality (c) of the features are encoded in an RGB image downloaded from the graphics card. For visualization purposes, the channels encoding the direction (a) and color (b) are colorized.

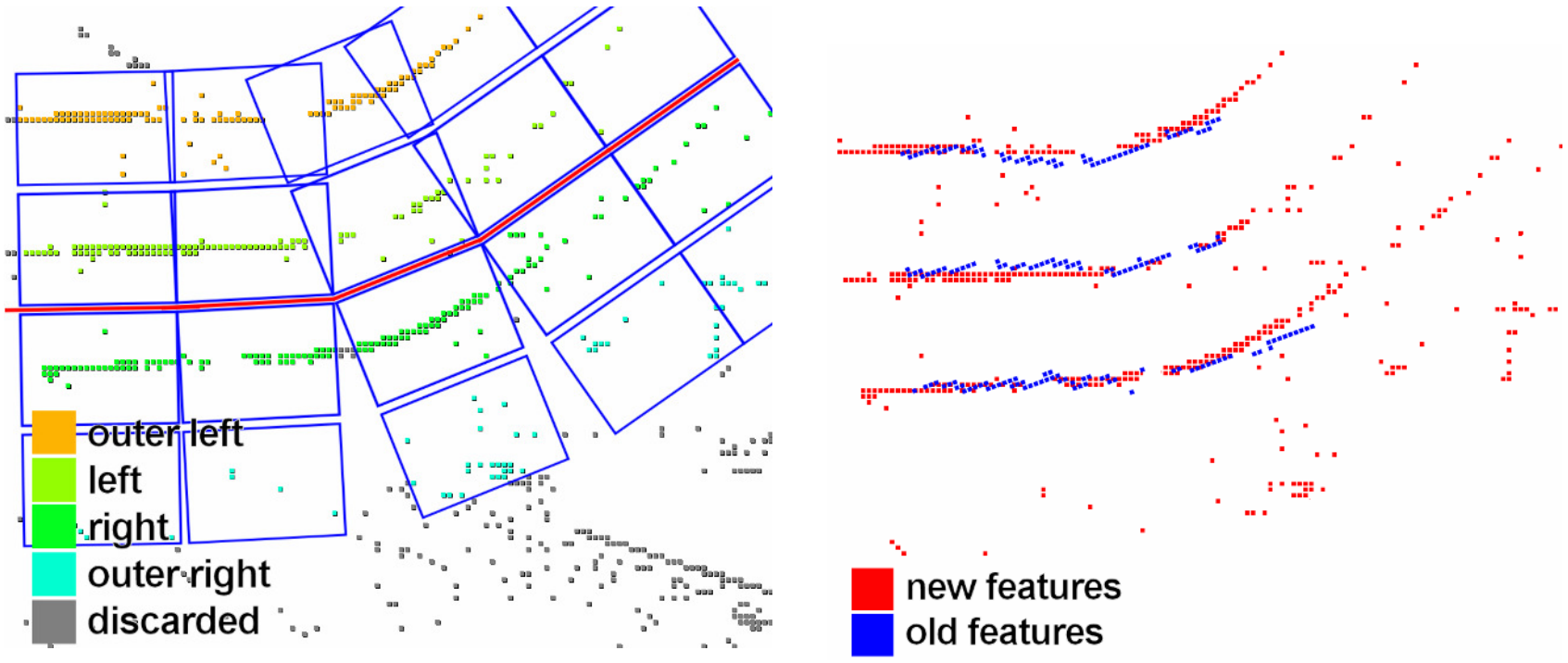

(a) labeling and discarding (b) mixing

**Fig. 15.** Regions of interest (a, blue boxes) determine to which lane marking features are assigned. Afterwards, old and new features are mixed (b).

angular changes $d_i$ of all segments, the position $p_i$ and global orientation $\alpha_i$ of each segment can be computed. Each segment contains information whether the vehicle's lane is confined by lane markings and whether additional lanes to the left and right exist. Straight streets, sharp curves and a mixture of both can all be described by the model.

**Lane Fitting.** The main goal of the lane fitting algorithm is to find a parameter set for a lane model that explains the features found in the current top view image and the previous frames. In order to create a global model of the lane, all feature points are mapped to world space coordinates and inserted into a list $l_p$. This is done using the function $f_{ego} : p_{car} \mapsto p_{world}$ defined by the current Ego State. Old data, i.e. feature points gathered during previous frames, may be kept if the features of a single image are too sparse. For each frame, the existing lane model or an initial guess is used to define four regions of interest as shown in Fig. 15(a). These are the regions expected to contain the own lane's markings and the lane markings of the adjoining lanes. If a feature is inside such a region, it is labeled as *outer left*, *left*, *right* or *outer right*. Otherwise, it is discarded. Afterwards, features from previous frames are mixed with the new data as depicted in 15(b).

As the first currently visible segment $s_f$ of the lane model is determined, older segments are no longer considered. If the list of lane segments is empty, it is initialized with $s_0 \leftarrow s_f$. Starting from $s_f$, each segment $s_i$ is estimated (or reestimated if it has previously been estimated). Therefore, an initial guess as to the orientation $\alpha_i$ of $s_i$ is made as shown in Fig. 16. All local features relevant for estimating $s_i$ are rotated by $\alpha_i$ around the starting point $p_i$ of $s_i$. A RANSAC algorithm is used to estimate the parameter $d_i$ and $w_i$: Iteratively, two feature points $p_x$ and $p_y$ are chosen. Assuming that they are located on the lane markings they were labeled for, the gradient $g_i = m_i/l_i$ as well as the width $w_i$ are derived from their coordinates. All features that are also sufficiently described by $g_i$ and $w_i$ are counted as inliers. This process is repeated $n$ times and the parameter set with most inliers is used to define $s_i$. A quality function $q$ takes into account the ratio of inliers and outliers, the amount of inliers, the quality of the features and states the quality of the segment. The quality is computed for every region of interest ($outer\ left$, $left$, $right$ and $outer\ right$). If the maximum of these qualities exceeds a threshold $t_q$, the segment is considered to be valid and the next segment $s_{i+1}$ is estimated. After all segments are estimated as shown in Fig. 17, a proposal about the lane markings' colors can be made by looking at the inliers' average color.

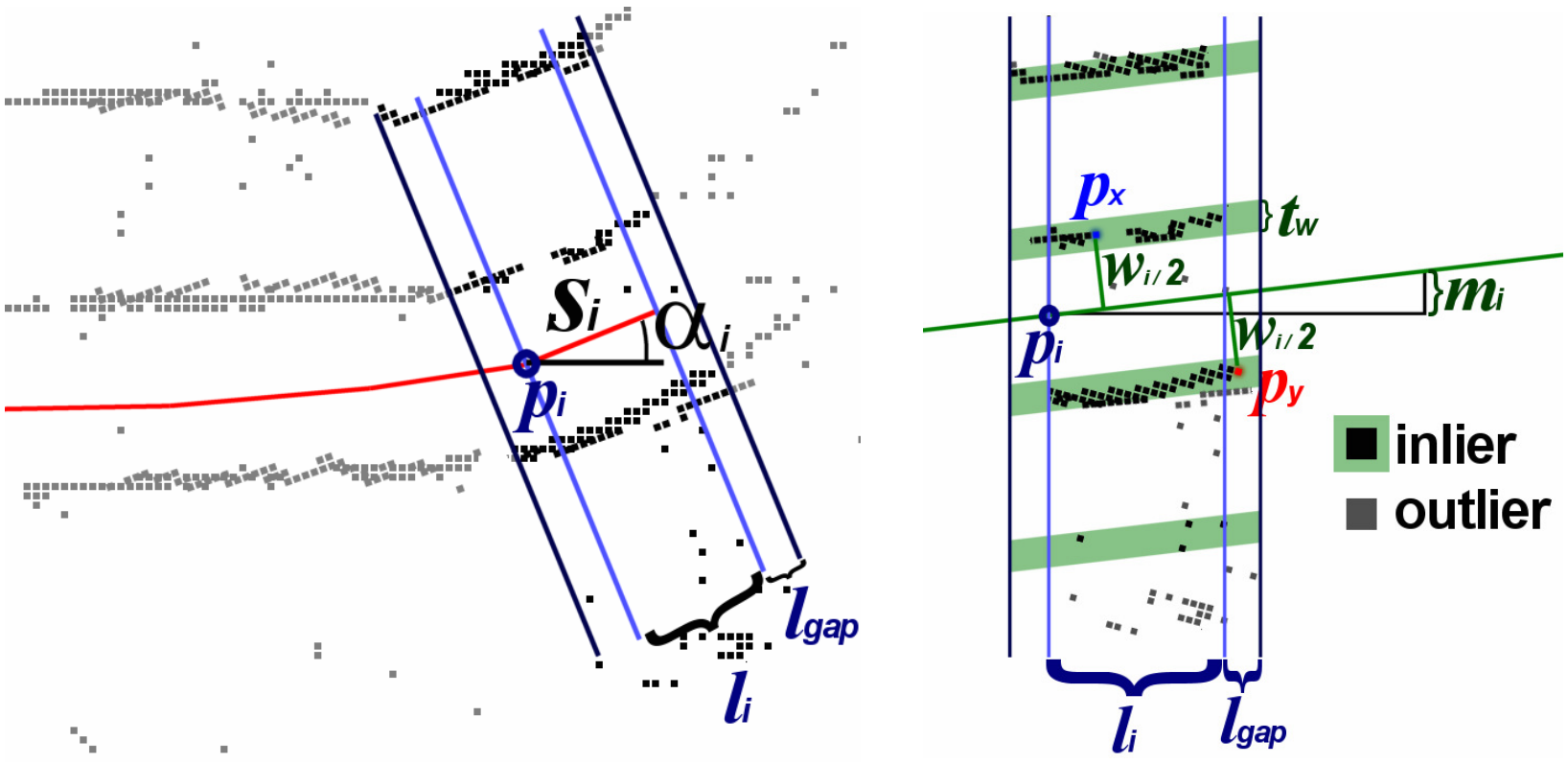

(a) estimating segment $s_i$ (b) RANSAC fitting

**Fig. 16.** $p_i$, $\alpha_i$, $l_i$ and $l_{gap}$ identify the features relevant for $s_i$. After rotating around $\alpha_i$, a RANSAC fitting eliminates outliers among the features.

**Results and Evaluation.** The algorithm was thoroughly tested on several sites in northern Germany and Texas. A frame rate of 10 fps could be maintained using a 2 GHz Intel Core 2 Duo with a GeForce 7600 GTS graphics card. The testing sessions included different weather and lighting conditions. The amount of false positives was reduced significantly by utilizing the vehicle's other sensors. The objects detected by lidar and radar sensors were used to mask out regions in the feature image where other cars, walls, cones and poles caused irritating artifacts in the top view image.

#### 4.2.2 Area Processor

The Area Processor consists of a single IDS color camera whose images are interpreted by a color segmentation algorithm suitable for urban environments. This algorithm separates an image into areas of drivable and non-drivable terrain. Assuming that a part of the image is known to be drivable terrain, other parts of the image are classified by comparing the Euclidean distance of each pixel's color to the mean colors of the drivable area in real-time. Moving the search area depending on each frame's result ensures temporal consistency and coherence. Furthermore, the algorithm classifies artifacts such as white and yellow lane markings and hard shadows as areas of unknown drivability. Although Caroline is able to perform basic driving tasks without this algorithm, it is needed in situations when terrain cannot be distinguished by other sensors, i.e., sections without proper lane markings, streets without high curbs and off-road tracks.

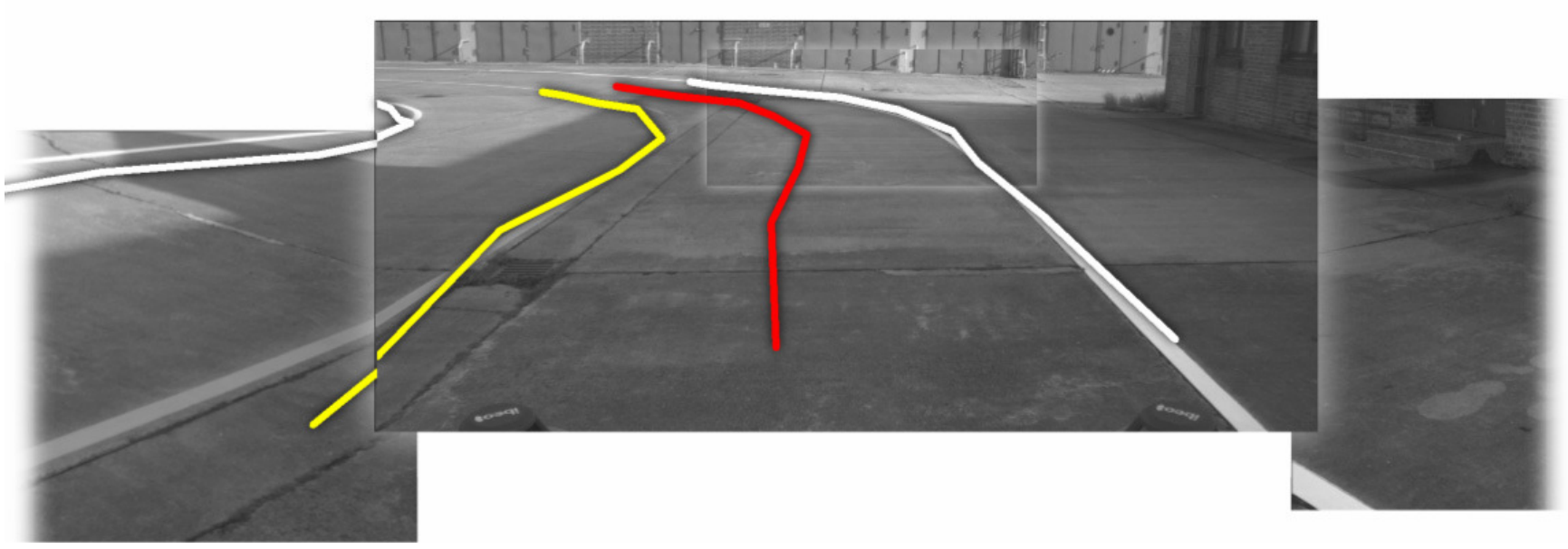

**Fig. 17.** The lane model reprojected onto the original images.

**Related work.** As a foundation for the area detection algorithm we used the real-time approach suggested by Thrun et al. [Thrun et al., 2006] in the 2005 DARPA Grand Challenge. The basic idea is to consider a given region in the actual image as drivable. The predominant mean color values in that area are retrieved and compared to the pixel values in the entire image. Similar pixels are marked as drivable. The algorithm was designed for off-road terrain, therefore it cannot be applied to urban scenarios without fundamental modifications. We will describe the algorithm in the next section. The Expectation Maximization (EM) algorithm used for color clustering in this approach is thoroughly described in [Duda and Hart, 1973] and [Bilmes, 1997]. Instead of the EM algorithm, the KMEANS algorithm that we used during the competition is also suitable for color clustering, as described in [Gary Bradski, 2005]. An algorithm similar to the one mentioned above points out the advantage of other color spaces than RGB [Ulrich and Nourbakhsh, 2000], e.g., the HSI space.

**The Stanford University algorithm for detecting drivable terrain.** The main idea of the algorithm is to use the output of the laser scanner, normally a scan-line, which is integrated over time to a height map in world coordinates. A polygon is defined that covers an area in front of the car identified as level and therefore as a drivable surface. This polygon is transformed into image coordinates from the camera and clipped to the image boundaries. The resulting polygon is considered as the area that is drivable. In this area the pixels' color values are collected and clustered by color, for example bright grey and yellow. These color clusters are compared to the color values of each pixel in the image using distance measurements in the color space. If a resulting distance is smaller than a given threshold, the area comprised by the pixel is marked as drivable. The main benefit of the algorithm is that the range in which drivability can be estimated is enhanced from only a few meters to more than 50 meters.

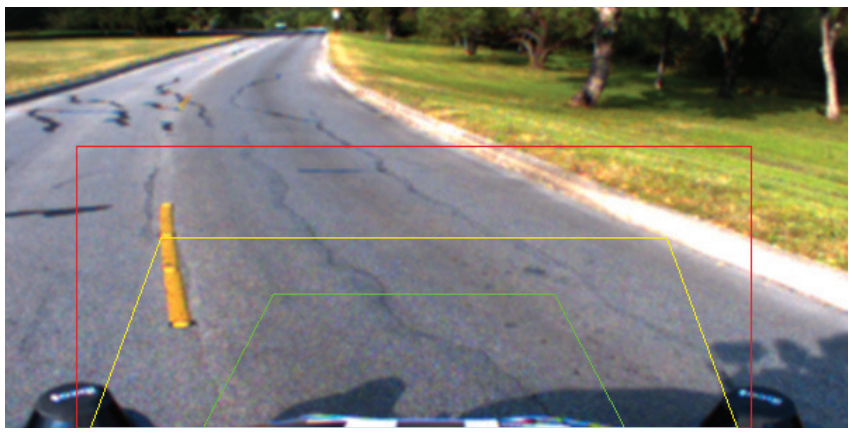

(a) original picture

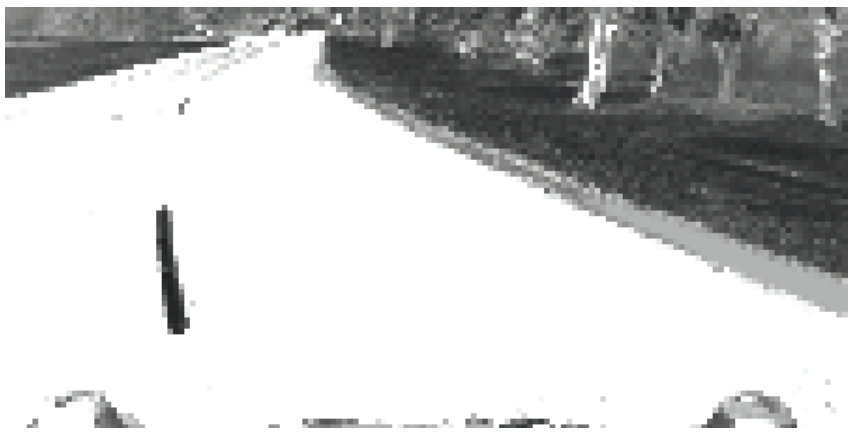

(b) normal drivability grid

**Fig. 18.** The drivability grid (b) depicts the output of algorithm, the results differ from black (undrivable) to white (drivable) . A yellow line (a) is marked as undrivable (b, black) because the color differs by too much from the street color.

**Problems arising in urban and suburban terrain.** Designed for competing in a 60 mile desert course, the basic algorithm succeeds well in explicit off-road areas, which are limited by sand hills or shrubs. When tested in urban areas new problems occur, because there are streets with lane markings in different colors or tall houses casting long shadows. The yellow lane markings are often not inside the area of the polygon $P_{Scanner}$ (output of the laser scanner), so they are not detected as drivable. Especially non-dashed lines prohibit a lane shift as shown in Fig. 18 and stop lines seem to block the road.

Another problem are shadows cast by tall buildings during the afternoon. Small shadows from trees in a fairly diffuse light change the color of the street only slightly and can be adapted easily. But huge and dark shadows appear as a big undrivable area as shown in Fig. 19. Even worse: Once inside a shadowed area, the camera auto exposure adapts to the new light situation, such that the area outside the shadow becomes overexposed and appears again as a big undrivable area as depicted in Fig. 20.

Another problem during the afternoon is the car's own shadow, in this paper referenced as "egoShadow", when the sun is behind the car. Sometimes it is marked as undrivable, sometimes it is completely adapted and marked as drivable, but the rest of the street is marked as undrivable as shown in Fig. 21. A fourth problem occurs when testing on streets without curbs but limited by mowed grassy areas. The laser scanner does not recognize the grass as undrivable, because its level is about the same as the street niveau. This causes the vehicle to move onto the grass, so that colors are adapted by the area processing algorithm, and consequentially keeps the car on the green terrain.

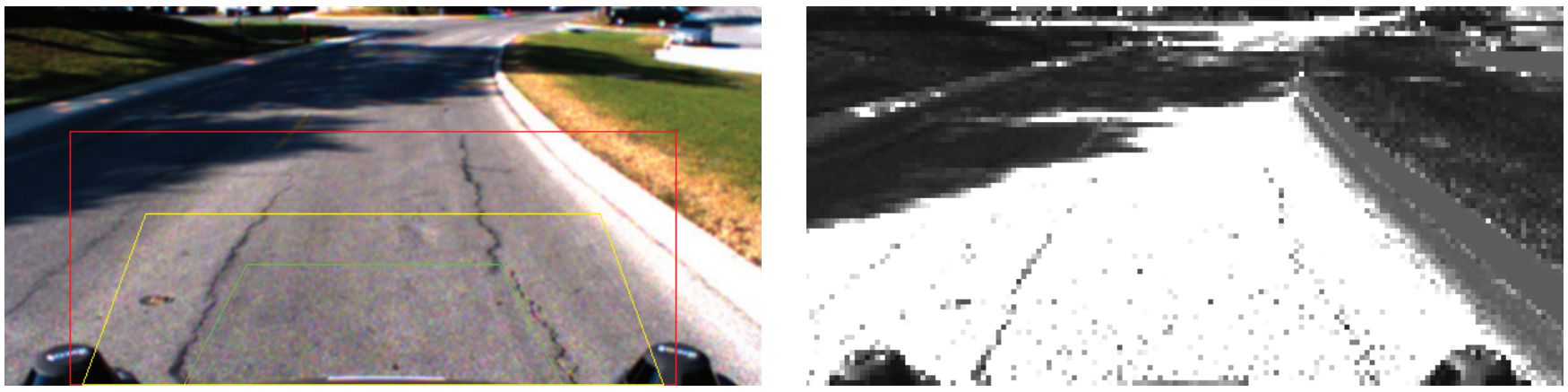

(a) original picture (b) normal drivability grid

**Fig. 19.** Large, dark shadows (a, left) differ too much from the Street Color (b, dark).

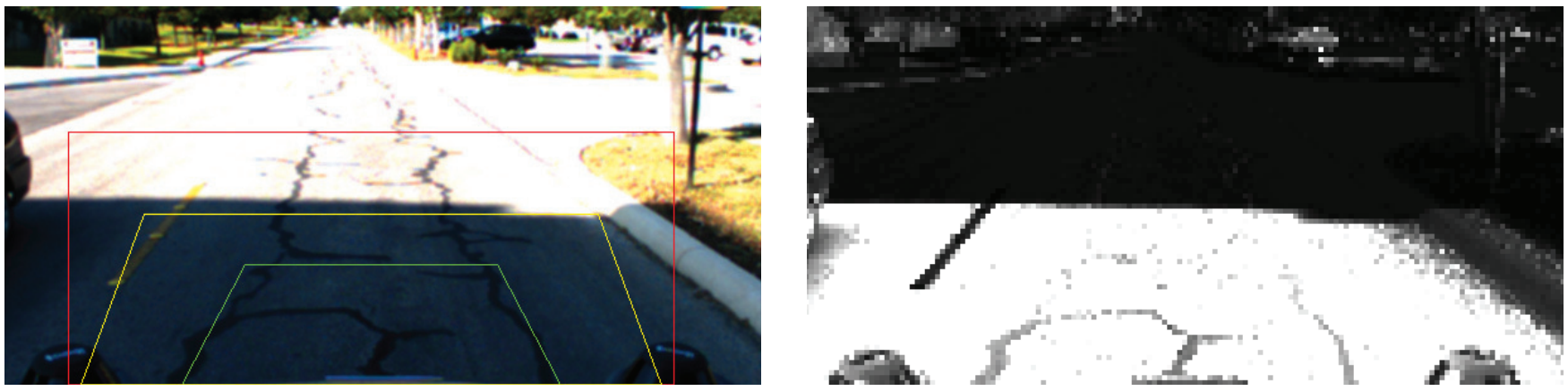

(a) original picture (b) normal drivability grid

**Fig. 20.** Exposure is automatically adapted inside shadows (a). Areas outside the shadow are overexposed and are marked as undrivable (b, dark).

**Alterations to the basic algorithm.** Differing from the original algorithm, our implementation does not classify regions of the image as drivable and undrivable. The result of our distance function is mapped to an integer number ranging from 0 to 127, instead of creating a binary information via a threshold. In addition, a classification into the categories 'known drivability' and 'unknown drivability' is applied to each pixel. These alterations are required because the decision about the drivability of a certain region is not made by the algorithm itself, but by a separate sensor fusion application. For the sake of performance the KMEANS Nearest Neighbors algorithm was chosen instead of the EM-algorithm, because the resulting grids are almost of the same quality but the computation is considerably faster. Tests have shown that better results can be achieved by using a color space that separates the luminance and the chrominance in different channels, e.g. HSV, LAB, YUV. The problem with HLS and HSV is that chrominance information is coded in one hue channel and the color distance is radial. For example, the color at 358 degrees is very similar to that one at 2 degrees, but they are numerically very far away from each other. Thus a color space is chosen where chrominance information is coded in two channels, for example in YUV or LAB, where similarity between two colors can be expressed as Euclidean distance.

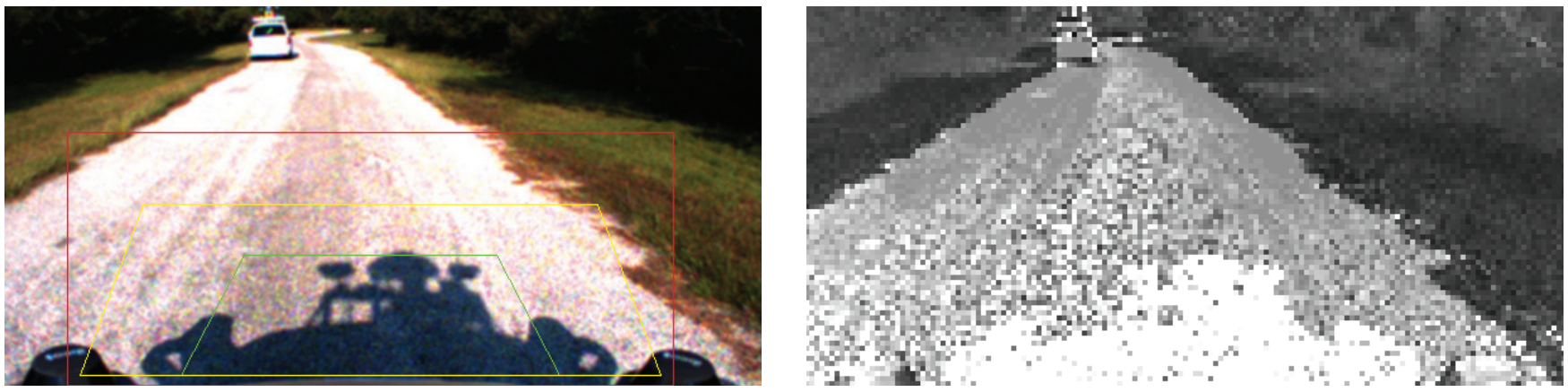

(a) original picture (b) normal drivability grid

**Fig. 21.** The vehicle's own shadow can lead to problems (a), for example if only the shadowed region is used to detect drivable regions (b, white).

**Preprocessing.** To cope with the problems of large shadows and lane markings, a preprocessing system was developed. Before the camera picture is processed, it is handed over to the following preprocessors: White preprocessor (masking out lane markings and overexposed pixels), black preprocessor (masking out large, dark shadows), yellow preprocessor (masking out lane markings), egoShadow preprocessor (masking out the car's shadow in the picture). The output of each preprocessor is a bit mask (1: feature detected, 0: feature not detected), which is used afterwards in the pixel classifying process, to mark the particular pixel as "unknown", which means that the vision-based area processor cannot provide valid information about the area represented by that pixel. In the following, the concept of each preprocessor is described briefly:

**White Preprocessor.** In order to deal with overexposed image areas during shadow traversing, pixels whose brightness value is larger than a given threshold are detected. The preprocessor converts the given image into HSV color space and compares the intensity value for each pixel with a given threshold. If the value is above the threshold, the pixel of the output mask is set to 1.

**Black Preprocessor.** As huge dark shadows differ too much from the street color and would therefore be labeled as impassable terrain, pixels whose brightness value is smaller than a given threshold are masked out. The preprocessor analogously converts the given image into HSV color space and compares the intensity value for each pixel with a given threshold. If the value is below the threshold, the pixel of the output mask is set to 1.

**Yellow Preprocessor.** Small areas of the image which are close to yellow in the RGB color space are detected so that yellow lane markings are not labeled as undrivable but rather as areas of unknown drivability. For each pixel of the given image, the RGB ratios are checked to detect yellow lane markings. If the green value is larger than the blue value and larger or a slightly smaller than the red value, the pixel is not considered yellow. If the red value is larger than the sum of the blue and the green values, the pixel is also not considered yellow. Otherwise, the pixel is set to $\frac{min(R,G)}{B} - 1$. Afterwards, a duplicate of the computed bit mask is smoothed using the mean filter, dilated and subtracted from the bit mask to eliminate huge areas. For different areas of the image, different kernel sizes must be applied. In the end, only the relatively small yellow areas remain. A threshold determines the resulting bit mask of this preprocessor.

**EgoShadow Preprocessor.** When the sun is behind the car, the vehicle's own shadow appears in the picture and is either marked as undrivable, or it is the only area marked as drivable. Therefore, a connected area directly in front of the car is identified whose brightness value is low. At the beginning of the whole computing process a set of base points $p(x, y)$ is specified, which mark the border between the engine hood and the ground in the picture. The region of interesst in each given picture is set to $y_{max}$, the maximum row of the base points, so that the engine hood is cut off. From these base points the preprocessor starts a flood-fill in a copy of each given image, taking advantage of the fact that the car's shadow appears in similar colors. Then the given picture is converted to HSV color space and the flood-filled pixel are checked to determine if their intensity value is small enough. Finally, the sum of the flood-filled pixels is compared to a threshold, which marks the maximum pixel area that constitutes the car's own shadow.

**The dynamic search polygon.** Using the output of the laser scanner to determine the input polygon works quite well if the drivable terrain is limited by tall objects such as sand hills or shrubs. In urban terrain, however, the output of the laser scanner must be sensitized to level distances smaller than curbs (10 to 20 centimeters), which becomes problematic if the street moves along a hill where the distance is much higher. Thus, the laser scanner

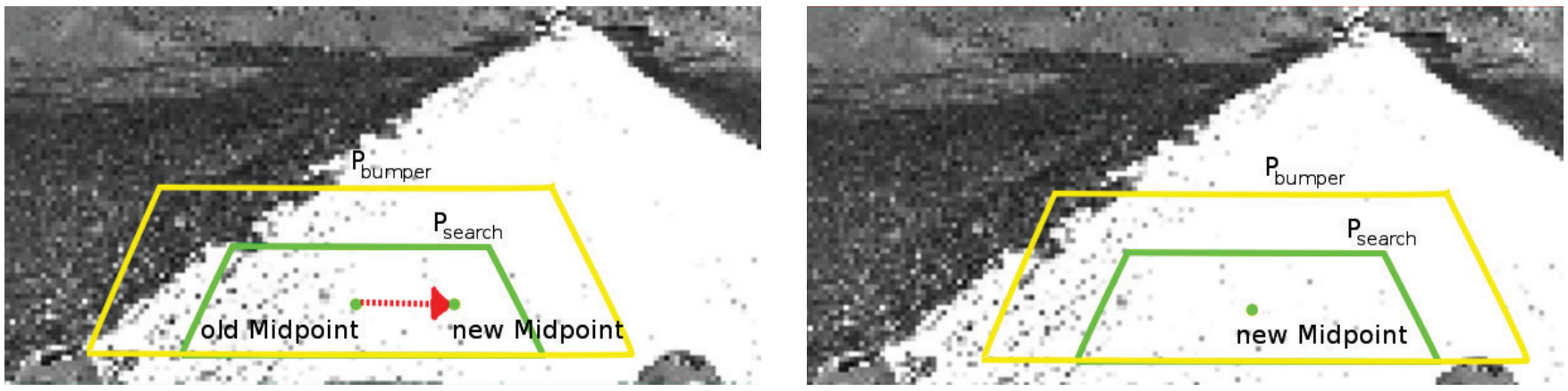

(a) dynamic polygon at time $t$ (b) dynamic polygon at time $t+1$

**Fig. 22.** This Fig. shows how the dynamic search polygon (a, green trapezoid) is transposed to the right (b) because the calculated moment is positive in $x$-direction.

polygon does not remain a reliable source especially because both modules solve different problems: The laser scanner focuses on range-based obstacle-detection [Ulrich and Nourbakhsh, 2000], which is based on analysis of the geometry of the surroundings, whereas the vision-based area processor follows an appearance-based approach. For example, driving through the green grass next to the street is physically possible, and therefore not prohibited by a range-based detection approach, but it must be prevented by the appearance-based system. This led to the concept of implementing a self-dynamic search polygon which has a static shape, but is able to move along both the X- and the Y-axis in a given boundary polygon $P_{boundary}$. The initial direction is zero. Every movement is computed using the output of the last frame's pixel classification. For the computation a bumper polygon $P_{bumper}$ is added, which surrounds the search polygon. The algorithm proceeds in the following steps:

**Implementation and Performance.** The algorithm has been implemented with the Intel OpenCV library [OpenCV Website, 2007]. The framework software is installed on an Intel Core 2 Duo Car PC with a Linux operating system and communicates with an IDS uEye camera via USB. The resolution of a frame is 640*480, but the algorithm applied downsampled images of size 160*120 to attain a manually adjusted average performance of 10 frames per second. The algorithm is confined to a region of interest of 160*75 cutting of the sky and the engine hood.

In Fig. 23 the difference between normal area processing and processing with the black preprocessor is shown. Without the preprocessor, the large shadow of a building to the left of the street is too dark to be similar to the street color and is classified as undrivable. The black preprocessor detects the shadowy pixels, which are classified as unknown (red).

The problem of overexposed areas in the picture is shown in Fig. 24, where the street's color outside the shadow is almost white and therefore classified as undrivable in the normal process. The white preprocessor succeeds in marking the critical area as unknown, so that the vehicle has no problem in leaving the shadowy area.

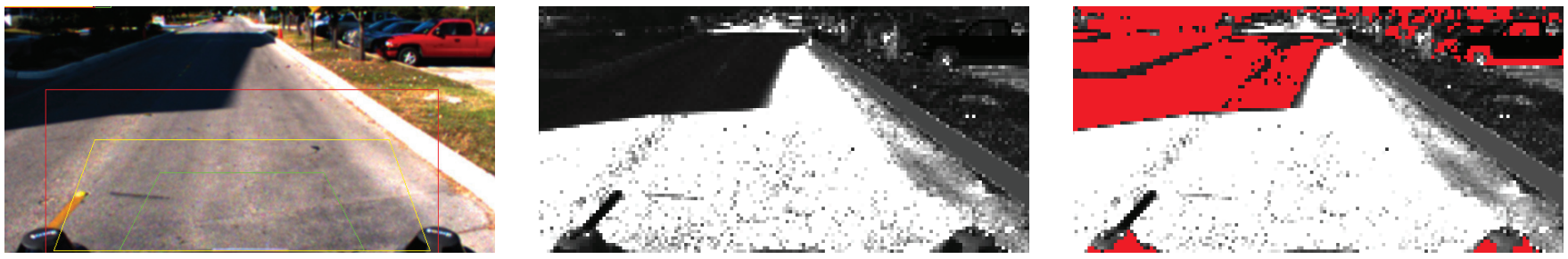

(a) original picture (b) normal drivability grid (c) with black preprocessor

**Fig. 23.** The results with black preprocessor. The picture in the center (b) shows the classification results without the black processor. In picture on the far right (c) the critical region is classified as unknown (red).

**Data**: last frame's grid of classified pixels, actual bumper polygon $P_{bumper}$
**Result**: updated position of the Polygons $P_{bumper}$

**begin**
  Initialize three variables $pixelSum$, $weightedPixelSumX$, $weightedPixelSumY$ to zero
  **foreach** *pixel of the grid which is inside the bumper* **do**
    count the amount $pixelSum$ of visited pixels
    **if** *drivability of the actual pixel is above a given threshold* **then**
      Add the pixel's $x$-Position relative to the midpoint of $P_{bumper}$ to $weightedPixelSumX$
      Add the pixel's $y$-Position relative to the midpoint of $P_{bumper}$ to $weightedPixelSumY$
    **end**
  **end**
  Perform the division $x_{moment} = \frac{weightedPixelSumX}{pixelSum}$ and $y_{moment} = \frac{weightedPixelSumY}{pixelSum}$ and round the results to natural numbers
  /* The value $x_{moment}$ gives the amount and direction of the movement of $P_{bumper}$ in $x$-direction, the value $y_{moment}$ gives the amount and direction of the movement of $P_{bumper}$ in $y$-direction. */
  Add the values $x_{moment}$ and $y_{moment}$ to the values of the actual midpoint of $P_{bumper}$ to retrieve the new midpoint of $P_{bumper}$
  Check the values of the new midpoint of $P_{bumper}$ against the edges of $P_{boundary}$ and adjust the values if necessary
  Add the values $x_{moment}$ and $y_{moment}$ to the values of the actual midpoint of the search polygon to retrieve the new midpoint of the search polygon as shown in Fig. 22
  To prevent that the search polygon gets stuck in a certain corner, it is checked, if $x_{moment} = 0$ or if $y_{moment} = 0$
  /* For example if $x_{moment} = 0$, it is evaluated, if the midpoint of $P_{bumper}$ is located right or left to the midpoint of $P_{boundary}$; $x_{moment}$ is set to 1, if $P_{bumper}$ is located left, otherwise it is set to $-1$. An analogous check can be performed for the $y_{moment}$. */
**end**

**Algorithm 2.** Dynamic search polygon algorithm.

Yellow lane markings differ from pavement in color space so that a human driver can easily detect them even under adverse lighting conditions. This advantage turns out to be a disadvantage for a standard classification system, which also classifies the lane markings as undrivable, as shown in Fig. 25: Lane markings are interpreted as tiny walls on the street. To counteract this problem, we use a preprocessing step, which segments colors similar to yellow. To deal with different light conditions, the color spectrum must be wider so that a brownish or grayish yellow is also detected. This leads to some false positives as shown in Fig. 25, but the disturbing lane markings are clearly classified as unknown. The vehicle is now able to change lanes without further problems.

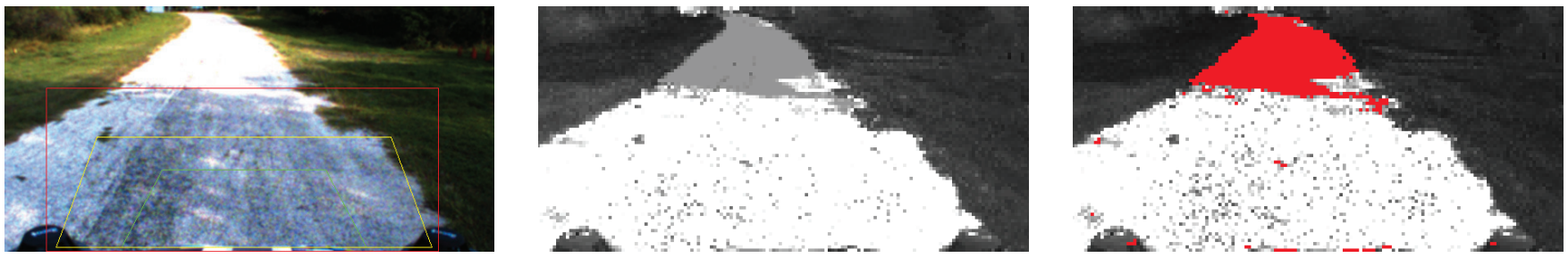

(a) original picture (b) normal drivability grid (c) with white preprocessor

**Fig. 24.** The results with white preprocessor.The picture in the center (b) shows the classification results without white processor. In picture on the far right (c) the critical region is classified as unknown (red).

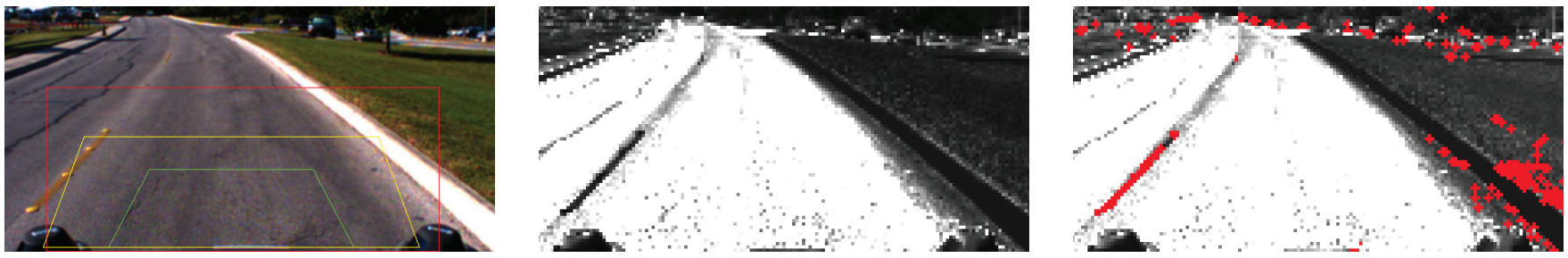

(a) original picture (b) normal drivability grid (c) with yellow preprocessor

**Fig. 25.** The results with yellow preprocessor. The picture in the center (b) shows the classification results without yellow preprocessor. In picture on the far right (c) the lane marks are classified as unknown (red).

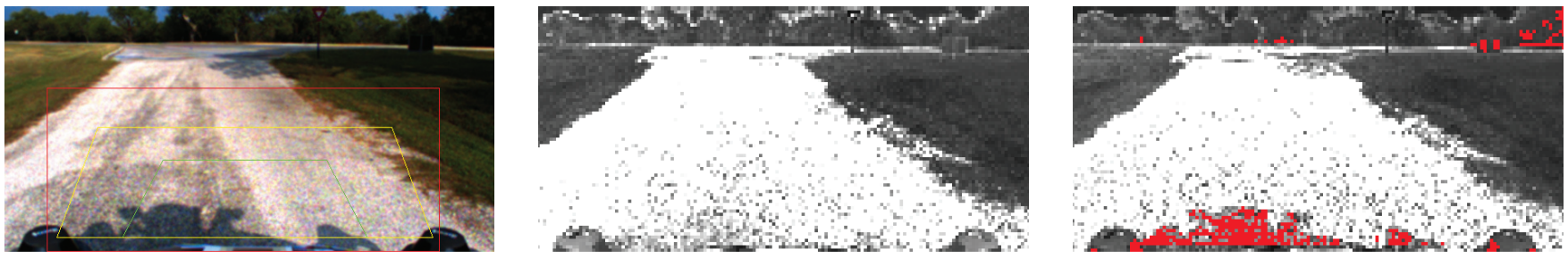

(a) original picture (b) normal drivability grid (c) with egoShadow preprocessor

**Fig. 26.** The results with egoShadow preprocessor. The picture in the center (b) shows the classification results without egoShadow processor. In picture on the far right (c) the car's own shadow is classified as unknown (red).

A problem with the vehicle's own shadow only occurs when the sun is located behind the vehicle, but in these situations the classification can deliver insufficient results. Figure 26 shows the shadowy area in front of the car as unknown.

The benefit of a search polygon that is transposed by the output of the last frame is tested by swerving about so that the car moves very close to the edges of the street. Figure 27 shows the results when moving the car close to the left edge. As the static polygon touches a small green area, a

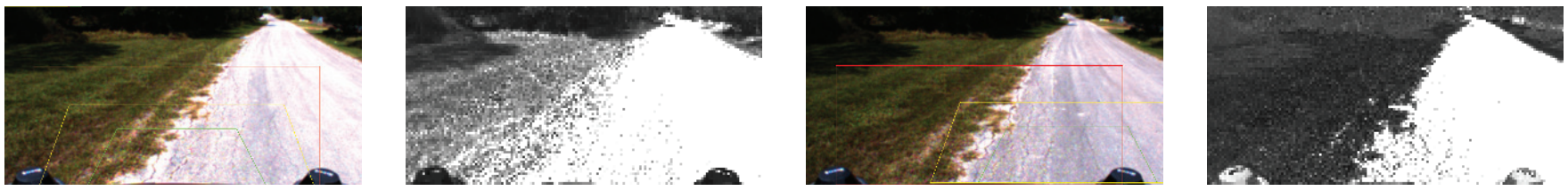

(a) original picture (static Polygon)

(b) drivability grid (static Polygon)

(c) original picture (dynamic Polygon)

(d) normal drivability grid (dynamic Polygon)

**Fig. 27.** The same frame first computed with a static search polygon (a, b), then with the dynamic polygon (c, d). The dynamic movement calculation caused the polygon to move to the right (c).

somewhat green mean value is gathered and so the resulting grid shows a certain amount of drivability in the grassland, whereas the dynamic polygon moves to the right of the picture to avoid touching the green pixels so that the resulting grid does not show drivability on the grassland.

## 4.3 Artificial Intelligence

### 4.3.1 The DAMN-Architecture

To control Caroline's movement, the artificial intelligence computes a speed and a turning wheel angle for every discrete step. Turning the steering wheel results in different circle-radii on which the car will move. Instead of the radii, the approach is based on the inverse, a curvature.

A curvature of 0 represents driving straight ahead, while negative curvatures result in left and positive curvatures in right turns as shown in Fig. 28.

This curvature, as the most important factor to influence, is selected in an arbiter as described in the DAMN-architecture [Rosenblatt, 1997]. This architecture models each input as behavior, which gives a vote for each possible curvature. More behaviors can be added easily to the system, which makes it very modular and extendable. The following behaviors are considered:

- Follow waypoints: Simply move the vehicle from point to point as found in the RNDF.
- Stay in lane: Vote for a curvature that keeps Caroline within the detected lane markings.
- Avoid obstacles: Vote for curvatures that keep the vehicle as far away from obstacles as possible and forbid curvatures leading directly into them.
- Stay on roadway: Avoid curb-like obstacles detected by grid-based fusion with laser scanners and color camera.
- Stay in zone: Keep the vehicle in the zone, defined by perimeter points in the RNDF.

All collected votes are weighted to produce an overall vote. The weights again are not fixed, they depend on factors including distance to an intersection, presence of lanes and more. A trajectory point is calculated by following the best voted curvature for one meter. A trajectory point holds information such as position, orientation and speed. Starting at this trajectory point, all behaviors vote again for curvatures to find the next point until a list of points is computed. This list has to be long enough to come to a complete stop at current speed. The speed is controlled by another arbiter influenced by different behaviors, which each provide a maximum speed. The arbiter simply selects the lowest of these speeds. These behaviors are: RNDFMax, sensor health, zone, reverse, safety zone, obstacle distance and following other obstacles. Based on the trajectory points calculated iteratively we design a drivable corridor for further processing by the next module in the chain, the path planner.

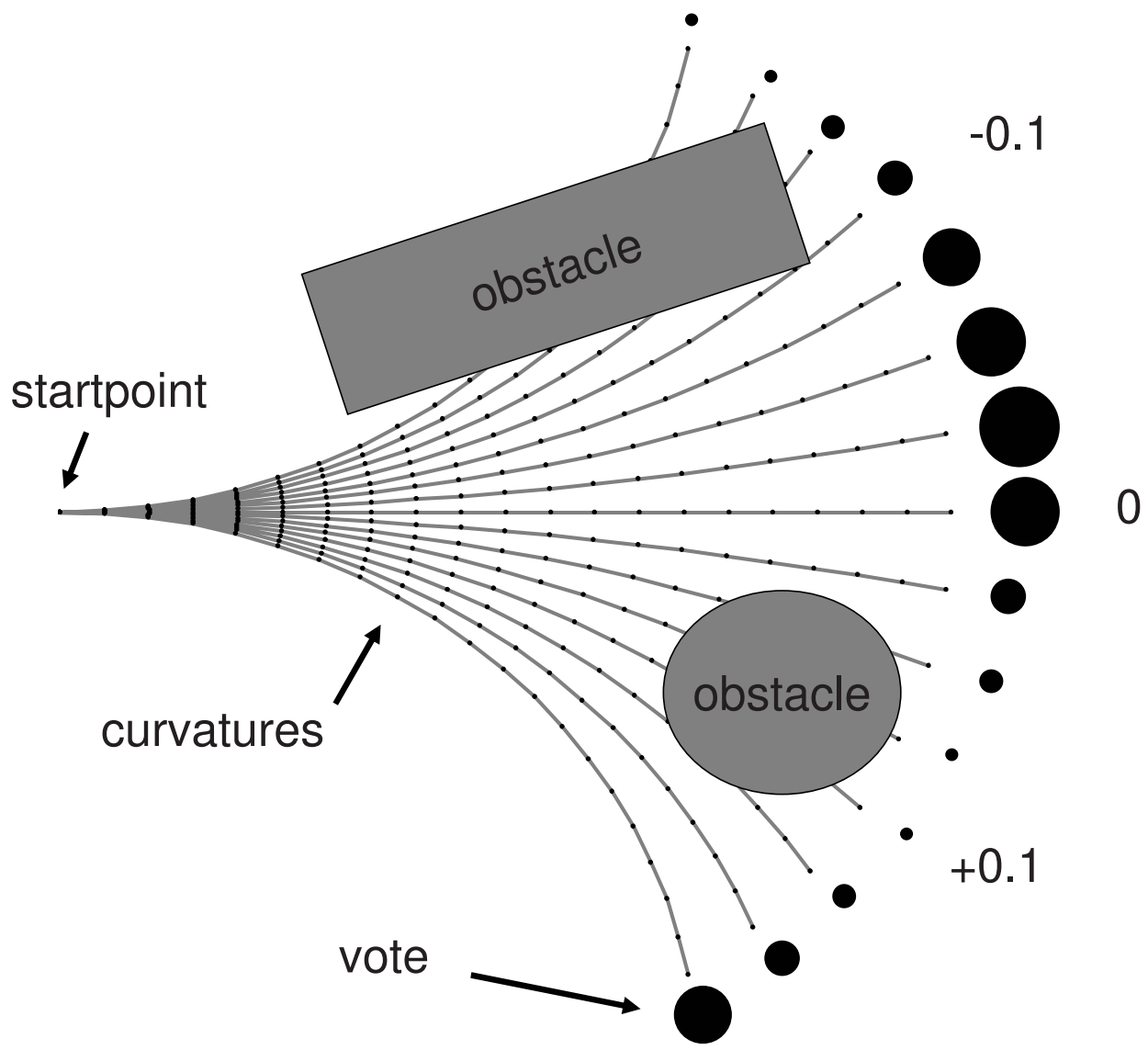

**Fig. 28.** Curvature field: Larger black circles represent preferred votes.

### 4.3.2 Interrupts

Because the AI has to deal with more complex situations, e.g. stopping at a stopline and yielding the right-of-way, than the DAMN-architecture is designed for, we extended DAMN by an interrupt system. At each trajectory point found each interrupt is called upon to decide if it wants to be activated at its location. If so, the speed stored in the trajectory points is reduced to bring the car to a smooth stop. If the point is reached, the interrupt is activated and the arbiters are stopped until the interrupt returns control to the arbiters. Some of our interrupts are:

- Intersection: Activated at a stopline until it is our turn.
- Queue: Wait in a line at an intersection.

- Overtake: Stop the car when the lane is blocked and wait for other lane to clear to start passing maneuver.
- U-turn: Activated at a dead-end street - this interrupt actually performs the U-turn and turns the car around.
- Road blocked: Activated if the entire road is blocked - this interrupt then activates the U-turn interrupt when appropriate.
- Parking: Activated at a good alignment in front of the parkbox - this interrupt returns control after the parking maneuver is finished.
- Pause: Active as long as the car is in pause mode.
- Mission complete: Final checkpoint is reached.

An example can be seen in Fig. 29, where the queueing interrupt has to be activated at some point in the future and the speed must therefore be reduced.

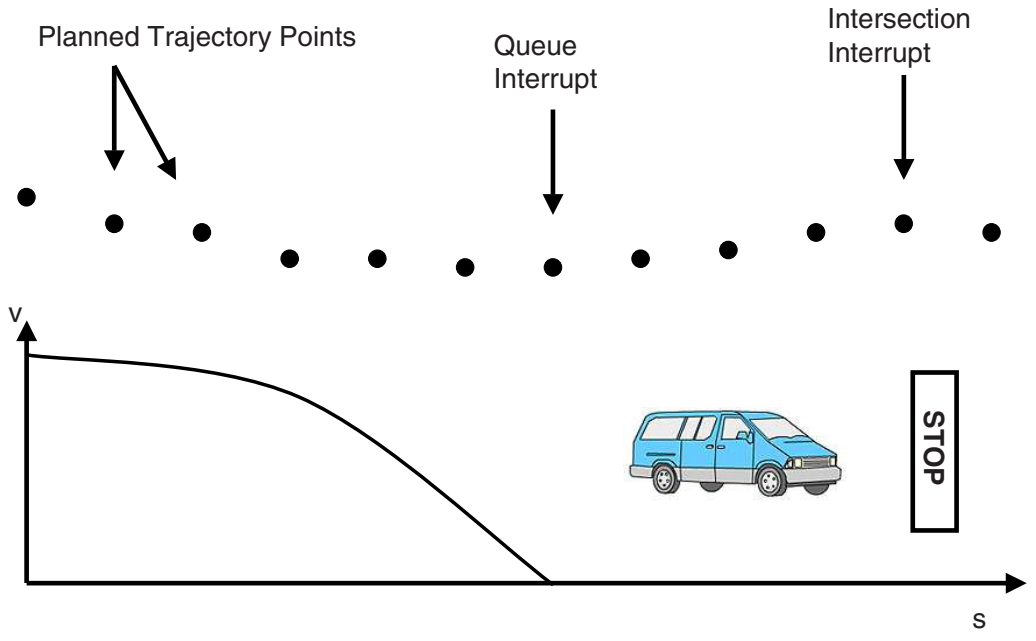

**Fig. 29.** Interrupt example.

### 4.3.3 Example

An example of how different behaviors interact is shown in Fig. 30. In the recorded situation, Caroline just started overtaking another car, blocking its lane. The plots represent the calculation of one trajectory: 20 trajectory points are calculated from the front to the back. For each point votes for 40 curvatures are made, these are displayed from left to right.

The lane behavior (a) demands a sharp left for the first four curvatures, then a right turn which finally transitions to straight driving. This would bring Caroline quickly to the free lane to pass the obstacle vehicle. The obstacle behavior (b) has two obstacles effecting the votes: On the left, a wall forbids going farther to the left, on the right one can see the car that is be passed. Finally the waypoint behavior (c) wants to go to the right all the time, because that is the lane where Caroline should be and where the waypoints are, but is outvoted by the other behaviors in (d).

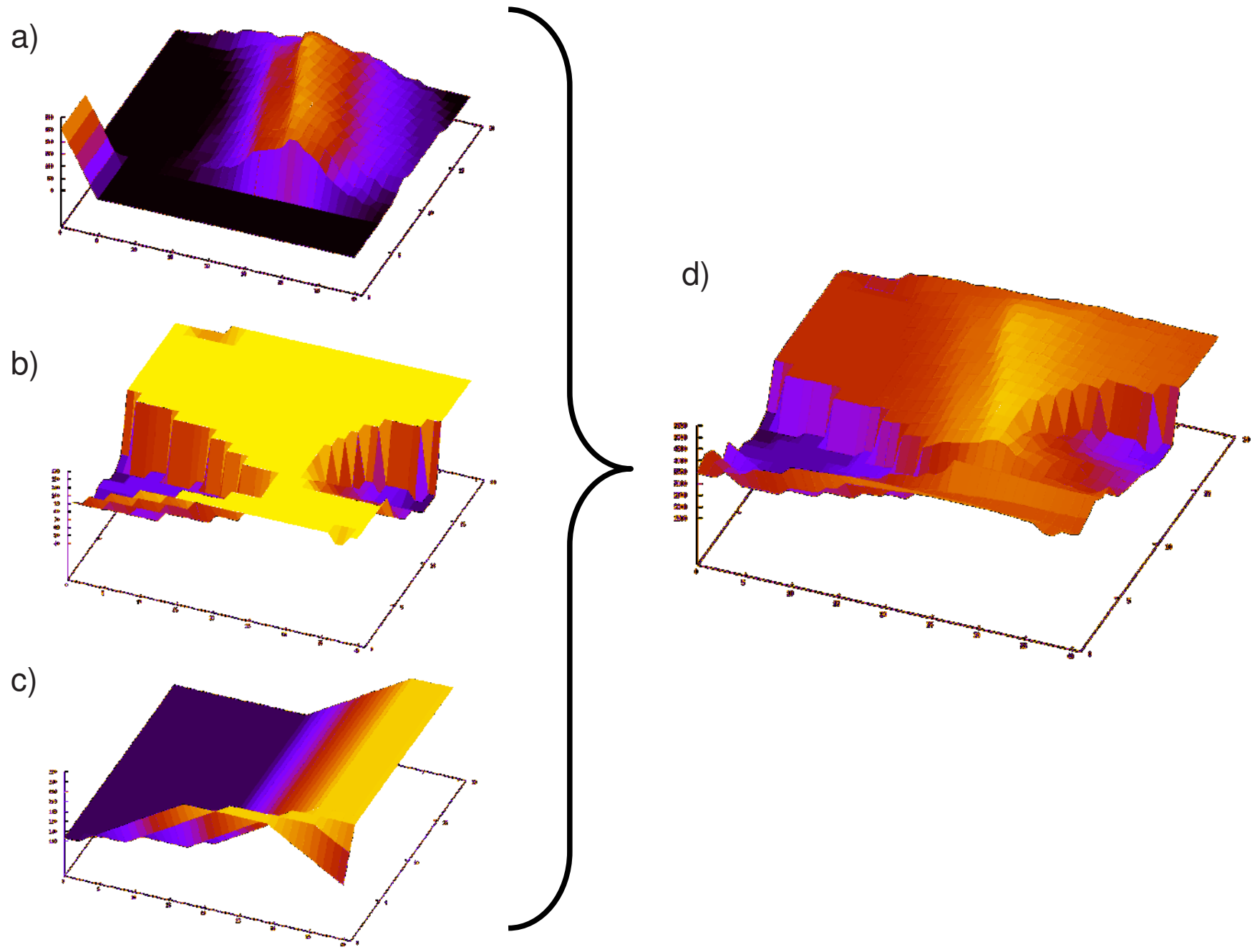

**Fig. 30.** Votes of a) stay in lane, b) avoid obstacles, c) follow waypoints, d) weighted sum.

## 4.4 Vehicle Control

Lateral and longitudinal control are the basics of autonomous vehicle guidance. In the following, both concepts as installed in Caroline for the DARPA Urban Challenge are discussed in detail.

### 4.4.1 Longitudinal Control

While the maximum and minimum speed of the vehicle is chosen by the artificial intelligence, the controller must calculate the braking and accelerator set points in order to maintain a given speed.

For this purpose, the longitudinal controller is separated into an outer and an inner loop controller. Based on the given speed set point, the outer loop controller determines the required acceleration. Finally, the inner loop controller calculates throttle and brake input to track the required acceleration. The acceleration of the vehicle, which is needed for feedback of the lower controller, is provided in high resolution by the GPS/INS system.

Gear shifting is handled via an automatic gear box. However, to switch between forward, backward and parking state, an automatic lever arm is attached at the gearshift. The lever arm position can be commanded with a CAN (Controller Area Network) interface.

**Longitudinal Dynamics.** The driving power must be greater than the sum of all driving resistances, that is the sum of rolling, air and acceleration resistance. Engine torque $M_M$ is a function of throttle $\alpha_A$, engine speed $n_M$ and engine acceleration $\dot{n}_M$.

$$M_M(\alpha_A, n_M, \dot{n}_M) = \frac{r}{\eta_k\, i_k}\,(f_R\, m\, g + c_w\, A\, \frac{\rho}{2}\,(\frac{n_M\, 2\,\pi\, R_0}{i_k})^2 + \lambda\, m \frac{\dot{n}_M\, 2\,\pi\, R_0}{i_k}) \tag{12}$$

The meaning of the parameter is given in table 1.

**Table 1.** Longitudinal model parameters.

| Symbol | Parameter |
|---|---|
| $R_0$ | Wheel Radius, Unloaded |
| $r$ | Wheel Radius, Loaded |
| $\eta_k$ | Degree of Efficiency, Gear Box |
| $i_k$ | Gear Transmission Ratio |
| $f_R$ | Rolling Friction Factor |
| $m$ | Mass |
| $g$ | Gravity |
| $c_w$ | Air Resistance Factor |
| $A$ | Cross Sectional Area |
| $\rho$ | Air Density |
| $\lambda$ | Moulding Bodies Factor |

The model is used for the inner loop controller to simulate different control strategies for the longitudinal control. The plant model for the outer loop controller is the transfer function between desired vehicle acceleration and actual vehicle speed. The inner loop is approximated as a PT1 element. In addition, an integral element is needed to integrate the speed from acceleration:

$$P(s) = \frac{1}{s\,(T\,s + 1)} \tag{13}$$

Introducing measured values of the drive chain into the model, leads to a value of $T = 0.6s$ for system lag.

**P-PD-Control Controller Cascade.** As mentioned above, the longitudinal controller is separated into an outer and inner control loop. The block diagram in Fig. 31 depicts the control structure. $K(s)$ stands for each transfer function of the different controller parts. Different control parameters are used for acceleration and deceleration. While a PD controller is applied for the inner loop, a P controller is introduced for the outer control loop. Control outputs for acceleration and braking are combined via a predefined logic to prevent the system from activating throttle and brake at the same time.

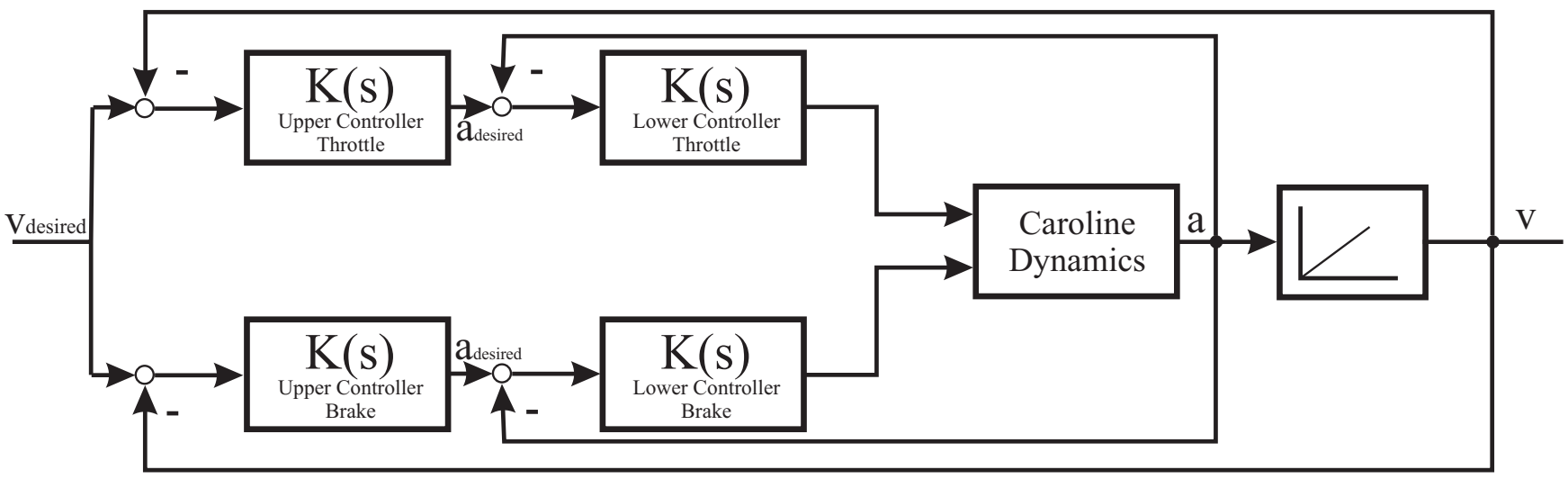

**Fig. 31.** Block diagram of the longitudinal controller.

In addition, an engine map can be used for direct feed forward of the throttle. Fig. 32 shows a typical implementation of an engine map for longitudinal control.

**Performance of the Longitudinal Controller.** Figure 33 illustrates the performance of the longitudinal control strategy. Two different examples are shown with two different speed profiles. While in the first example, the desired speed is changed in long and large steps, in the second example the speed is changed in shorter and smaller steps. The desired as well as the actual speed of Caroline are illustrated.

### 4.4.2 Lateral Control

It is the main goal of the lateral controller to follow a given trajectory with a minimum of track error. Secondly, vehicle driving maneuvers should match certain comfort parameters for smooth driving experience.

**Vehicle Dynamics.** For simulation of the vehicle as well as design of the controllers it is necessary to describe motion behavior with a mathematical model. In the following the bicycle model is used. The bicycle model is based on the following assumptions:

- The center of mass of the car is located at street level.
- Two wheels of each axle are combined as one wheel in the center of the axles.
- The longitudinal acceleration is zero.
- The wheel load of all wheels is constant.
- Lateral forces at the wheel are proportional to skew angle.

A state space representation within following structure is preferred:

$$\dot{\mathbf{x}}(t) = \underline{\underline{A}}\,\mathbf{x}(t) + \underline{\underline{B}}\,\mathbf{u}(t) + \underline{\underline{E}}\mathbf{z}(t), \qquad \mathbf{x}(0) = \mathbf{x}_0 \tag{14}$$

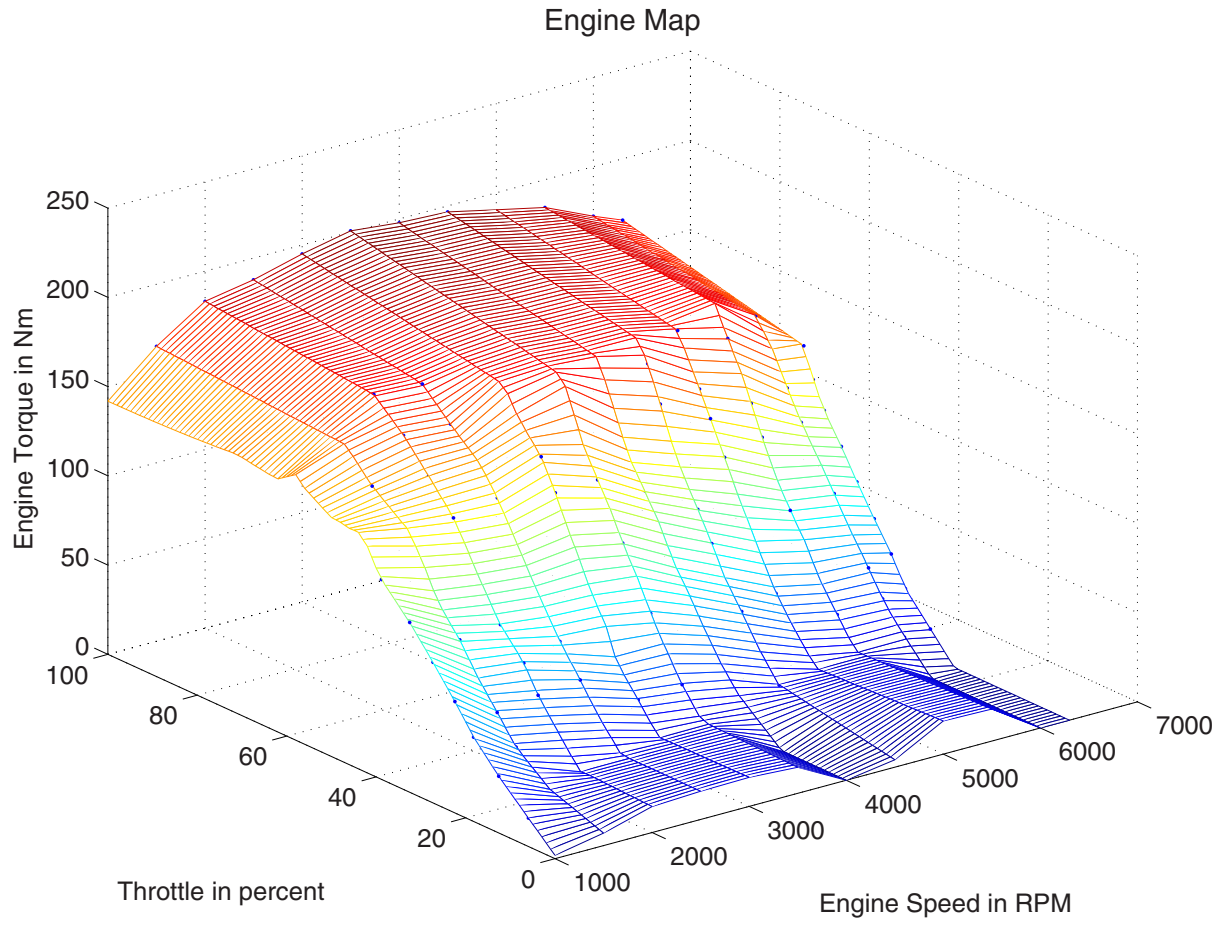

**Fig. 32.** Engine map.

Track error and track angle deviation have to be described mathematically to take them into consideration. Track angle deviation is defined as the difference between desired and actual orientation of the car. It is assumed that the derivation of the track angle $\zeta_{desired}$ can be calculated as the product of the curvature $\kappa$ of the track and the current speed $v$:

$$\zeta_{desired} = \kappa \cdot v \tag{15}$$

Yaw angle $\psi_{rel}$ with respect to the desired track is the difference between absolute yaw angle $\psi$ and desired track angle $\zeta_{desired}$:

$$\psi_{rel} = \psi - \zeta_{desired} \tag{16}$$

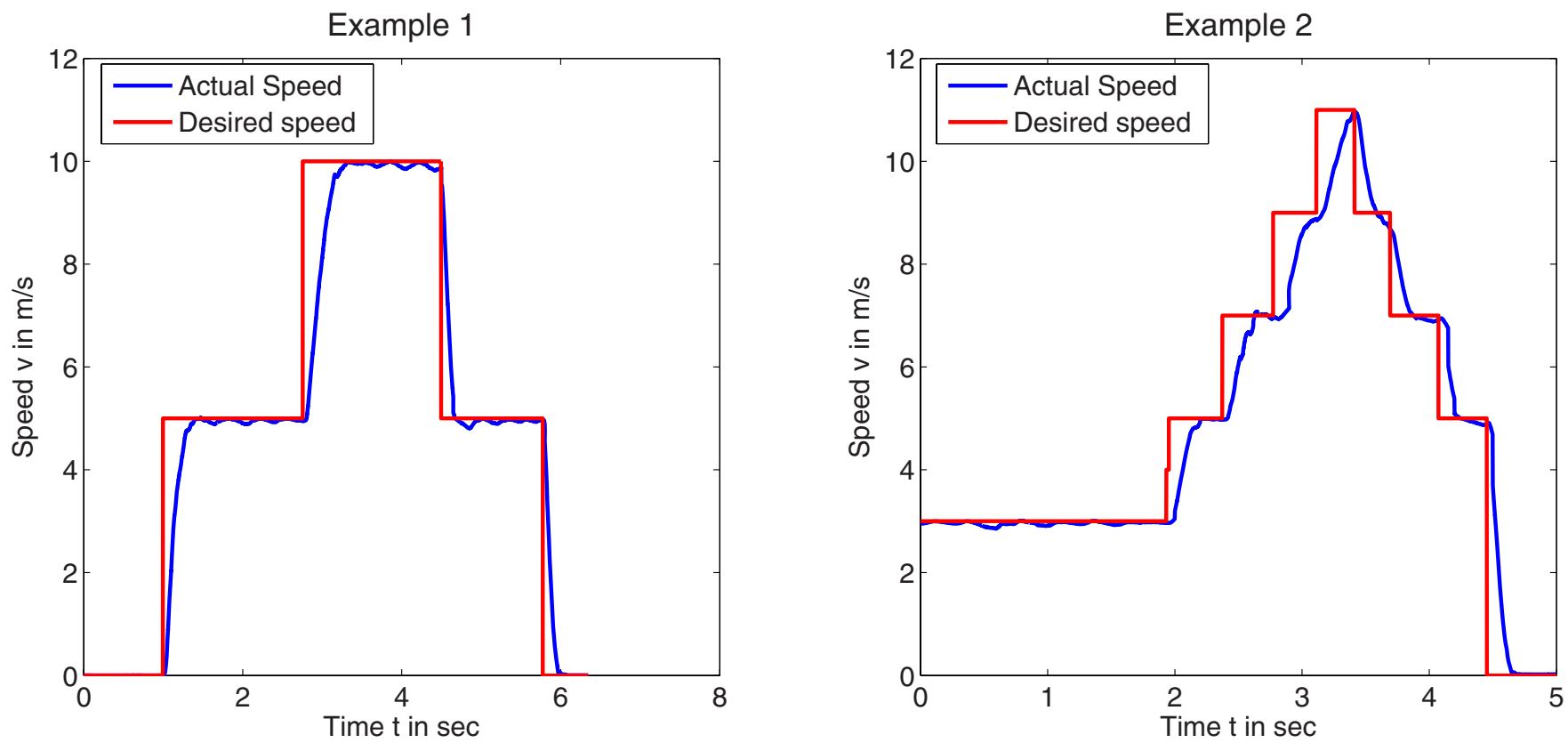

**Fig. 33.** Performance of the longitudinal controller.

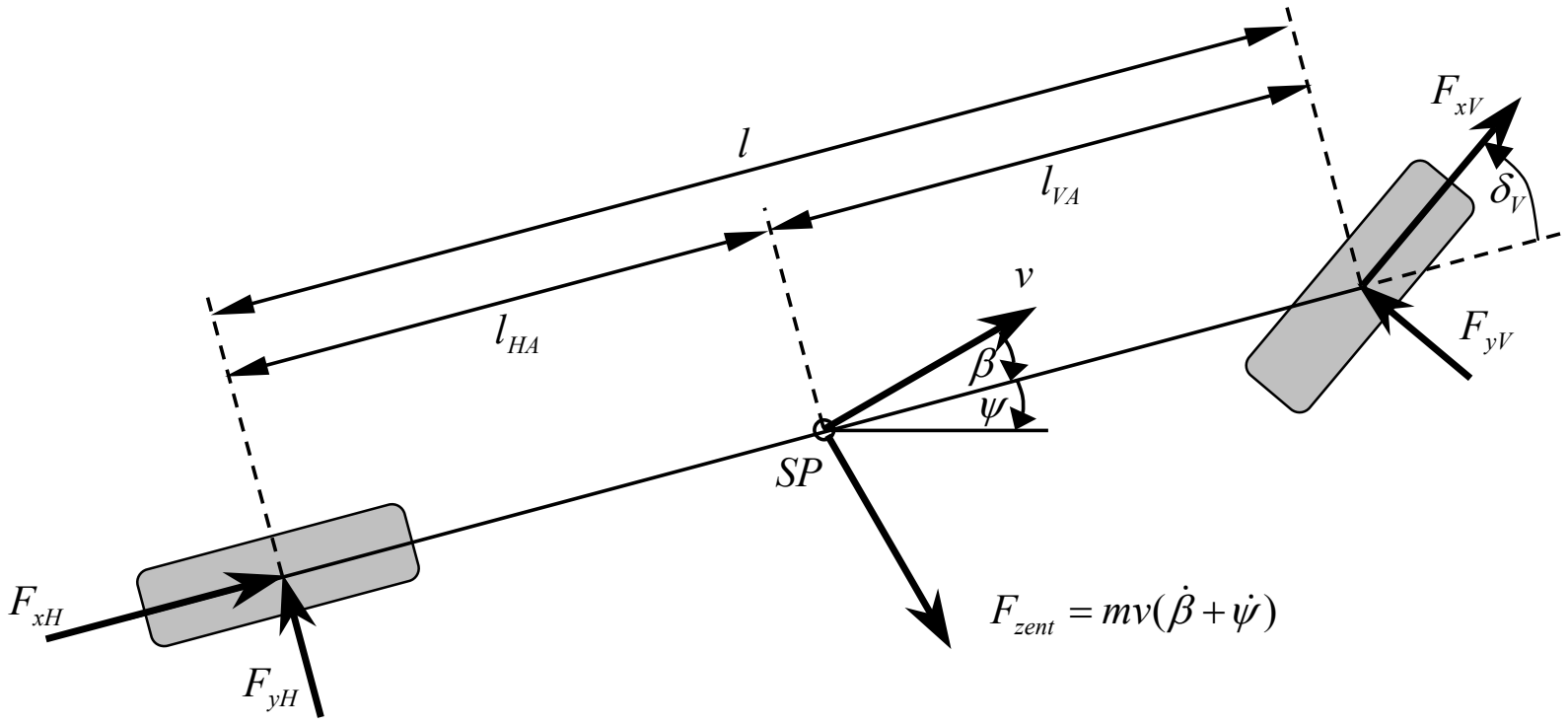

**Fig. 34.** Bicycle model.

As a result, yaw rate $\dot{\psi}_{rel}$ with respect to the desired track can be determined:

$$\dot{\psi}_{rel} = \dot{\psi} - \kappa\, v \tag{17}$$

Moreover, the derivation of the track error $\dot{d}$ can be formulated based on speed $v$, attitude angle $\beta$ and relative yaw angle $\psi_{rel}$:

$$\dot{d} = v\,(\beta + \psi_{rel}) \tag{18}$$

The state space representation of the bicycle model can be combined with the mathematical representation of the track error, track angle deviation and an additional time delay $T_L$ between commanded and actual steering wheel angle. The state vector consists of yaw rate $\dot{\psi}$, attitude angle $\beta$, relative yaw angle $\psi_{rel}$, track error $d$ and actual steering angle $\delta$. The result is the following state space model with the commanded steering angel $\delta_{desired}$ as the input variable and curvature $\kappa$ as outer noise:

$$\begin{pmatrix} \ddot{\psi} \\ \dot{\beta} \\ \dot{\psi}_{rel} \\ \dot{d} \\ \dot{\delta} \end{pmatrix} = \begin{pmatrix} a_{11} & a_{12} & 0 & 0 & a_{15} \\ a_{21} & a_{22} & 0 & 0 & a_{25} \\ 1 & 0 & 0 & 0 & 0 \\ 0 & v & v & 0 & 0 \\ 0 & 0 & 0 & 0 & -\frac{1}{T_L} \end{pmatrix} \cdot \begin{pmatrix} \dot{\psi} \\ \beta \\ \psi_{rel} \\ d \\ \delta \end{pmatrix} + \begin{pmatrix} 0 \\ 0 \\ 0 \\ 0 \\ \frac{i_L}{T_L} \end{pmatrix} \cdot \delta_{desired} + \begin{pmatrix} 0 \\ 0 \\ -v \\ 0 \\ 0 \end{pmatrix} \cdot \kappa \tag{19}$$

with

$$a_{11} = -\frac{c_{\alpha V}\, l_V^2 + c_{\alpha H}\, l_H^2}{\theta\, v}, \quad a_{12} = -\frac{c_{\alpha V}\, l_V + c_{\alpha H}\, l_H}{\theta}, \quad a_{15} = \frac{c_{\alpha V}\, l_V}{\theta} \tag{20}$$

$$a_{21} = -1 - \frac{c_{\alpha V}\, l_V - c_{\alpha H}\, l_H}{m\, v^2}, \quad a_{22} = -\frac{c_{\alpha V} + c_{\alpha H}}{m\, v}, \quad a_{25} = \frac{c_{\alpha V}}{m\, v} \tag{21}$$

The parameters are described in table 2.

The output of the system is the track error $d$.

$$\mathbf{y}(t) = \begin{pmatrix} 0 & 0 & 0 & 1 & 0 \end{pmatrix}^T \mathbf{x}(t) \tag{22}$$

Based on the state space model, the transfer function can easily be determined. The control transfer function is

$$F_c(s) = \frac{i_L}{T_L\, s + 1} \cdot \frac{a_{25} s^2 + (a_{15}\, a_{21} + a_{15} - a_{25}\, a_{11})\, s + (a_{25}\, a_{12} - a_{25}\, a_{12})}{s^2 - (a_{11} + a_{22}) s + (a_{11}\, a_{22} - a_{12}\, a_{21})} \cdot \frac{1}{s} \cdot \frac{v}{s} \tag{23}$$

and the noise transfer function:

$$F_{noise} = -\frac{v}{s} \cdot \frac{v}{s} \tag{24}$$

**Table 2.** Parameters of the bicycle model.

| Symbol | Parameter |
|---|---|
| $c_{\alpha V}$ | Skew Stiffness, Front Wheel |
| $c_{\alpha H}$ | Skew Stiffness, Back Wheel |
| $l_V$ | Wheel Base Front to Center of Mass |
| $l_H$ | Wheel Base Back to Center of Mass |
| $\theta$ | Moment of Inertia |
| $m$ | Mass |

**Parallel Structure Control.** As modeled, the vehicle has three degrees of freedom, which are the $x$ and $y$ position as well as the orientation $\psi$ of the car. Only the steering angle $\delta$ is available for controlling the system. As a result, the three degrees of freedom are handled simultaneously. Track error and track angle deviation are used as feedback signals. The working point is chosen at the speed of 30 km/h.

Figure 35 shows the structure of the control strategy used. Again, $K(s)$ stands for each transfer function of the controller. It consists of two parallel control loops for track error and track angle deviation as well as a pilot control taking the curvature of the desired trajectory into consideration. The map-based pilot control algorithm calculates the steering angle that would be needed to follow the desired track based on parameters of the bicycle model.

**Performance of the Lateral Controller.** Lateral control strategy has to handle different kinds of trajectories. On the one hand, the vehicle has to follow trajectories with a curvature of approximately $\kappa \approx 0$ at higher speeds. On the other hand, the track error in twisting areas is supposed to be as small as possible. Figure 36 shows an example of a trajectory that consists of a long straight part and two sharp curves. On the straight section, the vehicle is accelerated up to a speed of almost $v = 50$ km/h. The curves are driven at a speed of approximately 20 km/h. The speed profile is shown in

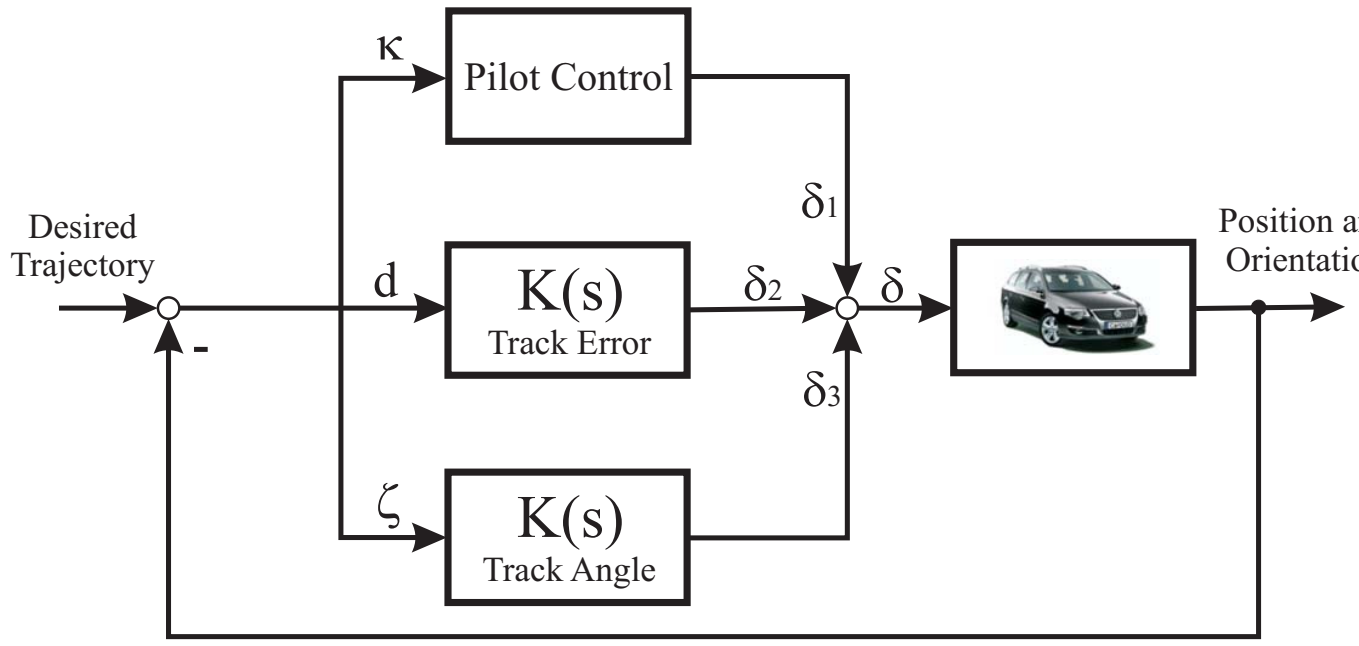

**Fig. 35.** Lateral control strategy.

Fig. 37. The performance of the control strategy in terms of track error can be seen in the same figure.

The control strategy shown worked well during all tests and missions during the DARPA Urban Challenge. It has always been stable with quite a low track error.

### 4.5 Safety

The safety systems of Caroline have to ensure the highest possible safety for the car and the environment in both manned or unmanned operation. It has to monitor the integrity of all viable hardware and software components. In case of an error, it has to bring the car to a safe stop. Furthermore, it must provide an interface for pausing or disabling the car using a remote E-stop controller. We extended these basic features by including the possibility to reset and restart seperate modules independently using hardware and/or software means in order to gain the option of automated failure removal. Figure 38 depicts this watchdog concept.

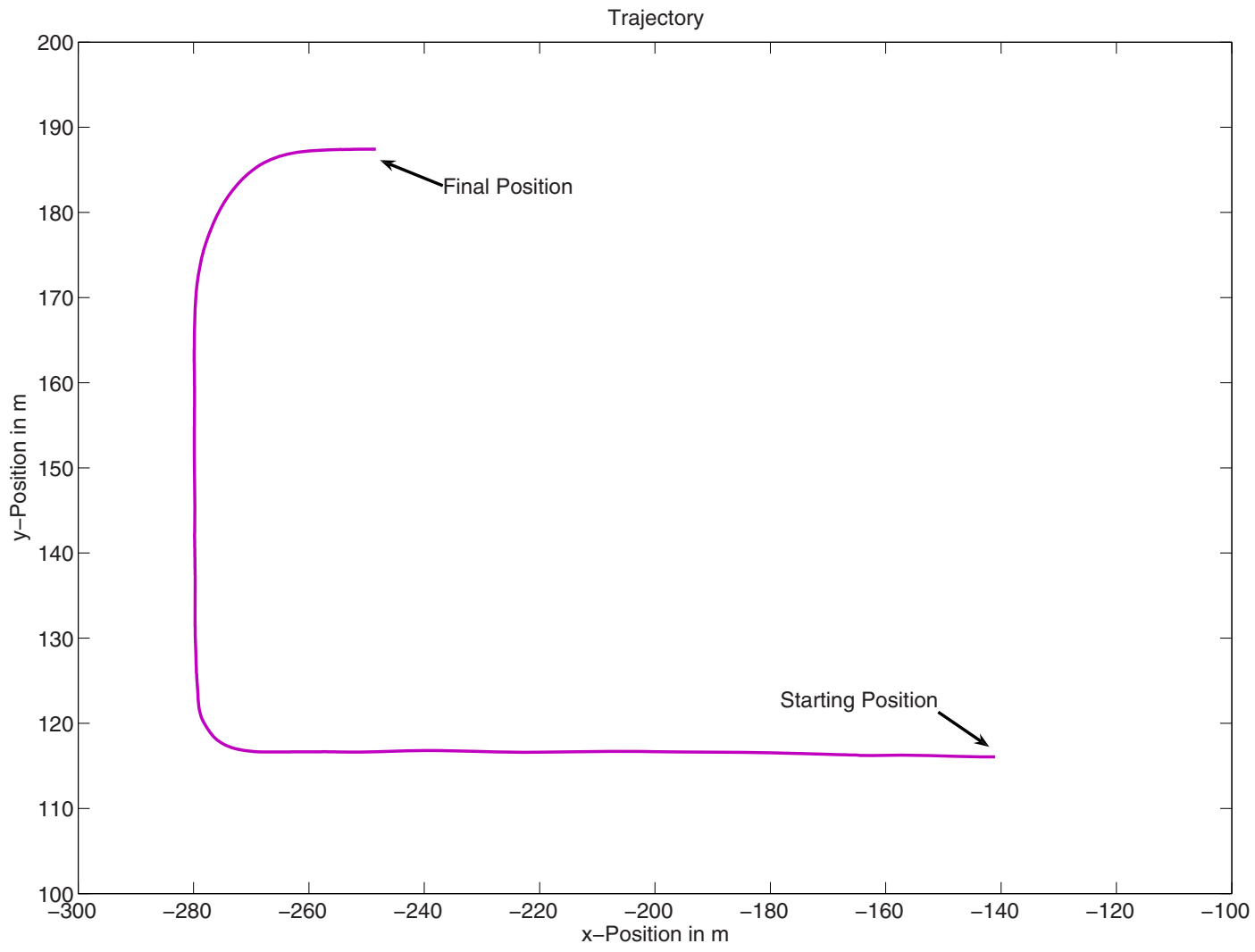

**Fig. 36.** Trajectory.

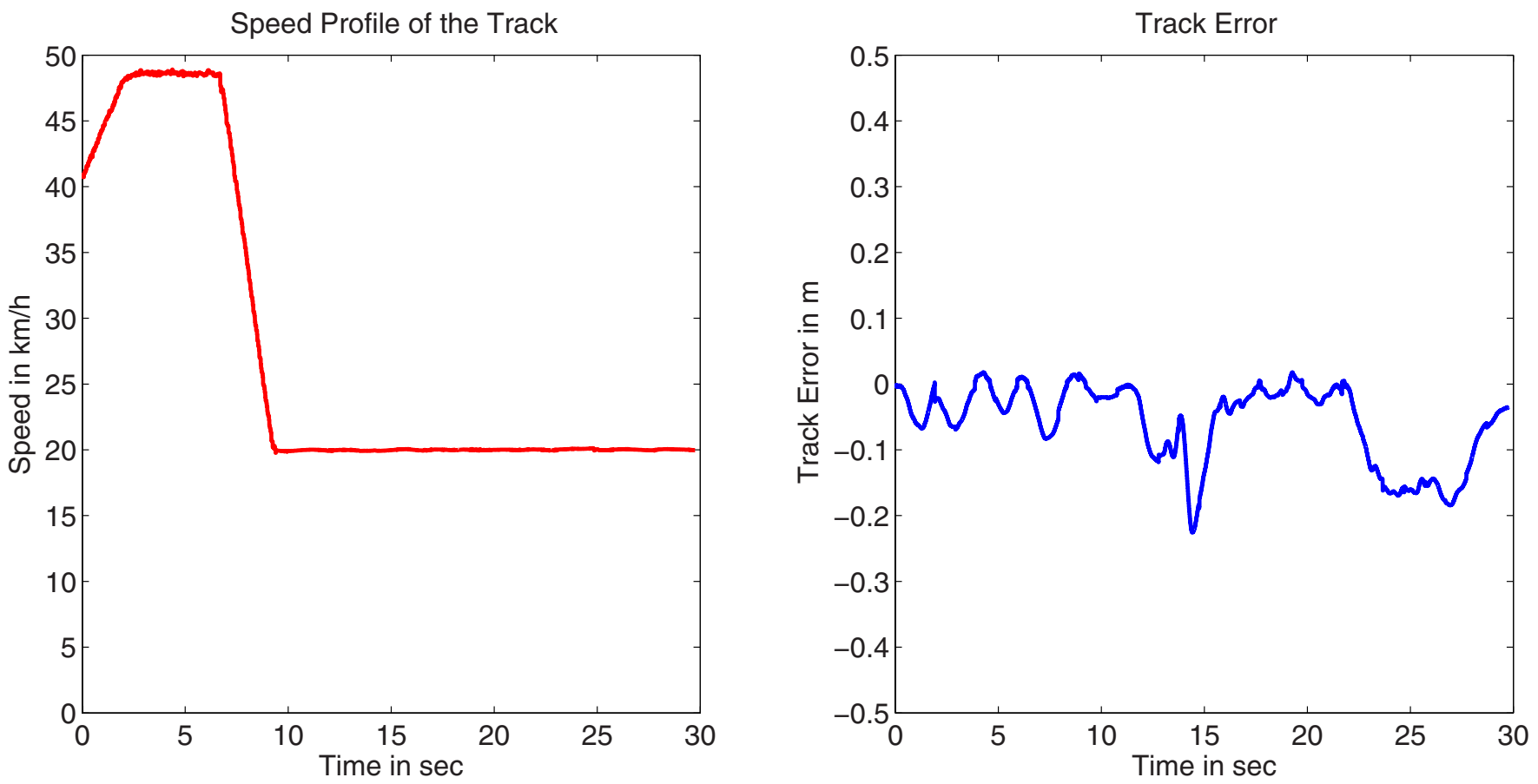

**Fig. 37.** Speed profile and track error of the trajectory.

Caroline is equipped with two separate brake systems. The main hydraulic system and an additional electrical parking brake. The main hydraulic brake is controlled by pressure, usually generated with a foot pedal by the driver. In autonomous mode, this pressure is generated by a small hydraulic brake booster. The parking brake is controlled by a push button in the front console.

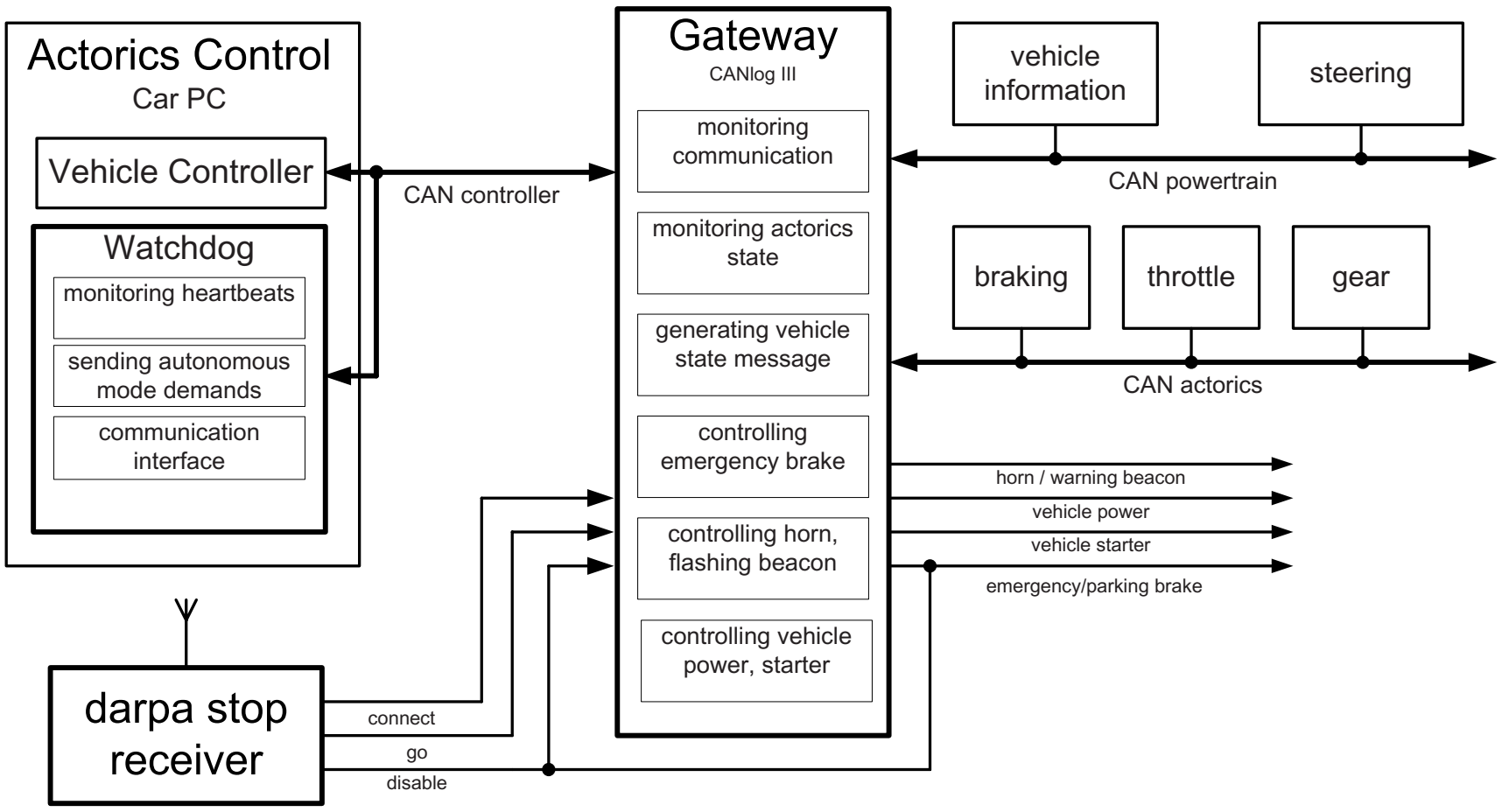

**Fig. 38.** Watchdog architecture.

This brake is a useful additional feature. If the button is pressed while the car is rolling, the main brake system is activated in addition to the parking brake until the car comes to a complete stop. During autonomous mode, the watchdog gateway, the emergency buttons on the top of the car and the receiver for the remote E-stop controller form a safety circuit, which holds a safety relay open. This relay is connected to the push button for the parking brake. If one of the systems fails, or is activated, the safety circuit is opened, the contact of the relay is closed and the push button of the parking brake is activated. During emergency braking, the lateral controller of the car is still able to hold the car on the given course.

Although the watchdog's main purpose is to assure safety it also increases the system's overall reliability. Caroline is a complex system with custom or pre-production hardware and software modules. These components were developed in a very short time and therefore are not as reliable as off-the-shelf commercial products. For this reason we used devices primarily implemented for all safety-relevant subsystems in order to also provide the means to monitor and reset non-safety relevant subsystems.

Each host runs a local watchdog slave daemon, which monitors all local applications as shown in Fig. 39. A process failing to send periodic heartbeats within a given interval indicates a malfunction, such as memory leakage or deadlocks. Therefore the process and all dependent processes are terminated by the local watchdog slave, to be restarted with respect to the order required by process dependencies.

The slave watchdog itself is monitored by a remote central master watchdog. This approach allows the detection of malfunctions that cannot be resolved by the local slave watchdog, e.g. if a computer freezes. If a computer should freeze, an emergency stop is initiated and the failed system is power-cycled to restart in a stable state. The master watchdog is monitored by the CAN gateway, which initiates an emergency stop on failure of the master watchdog.

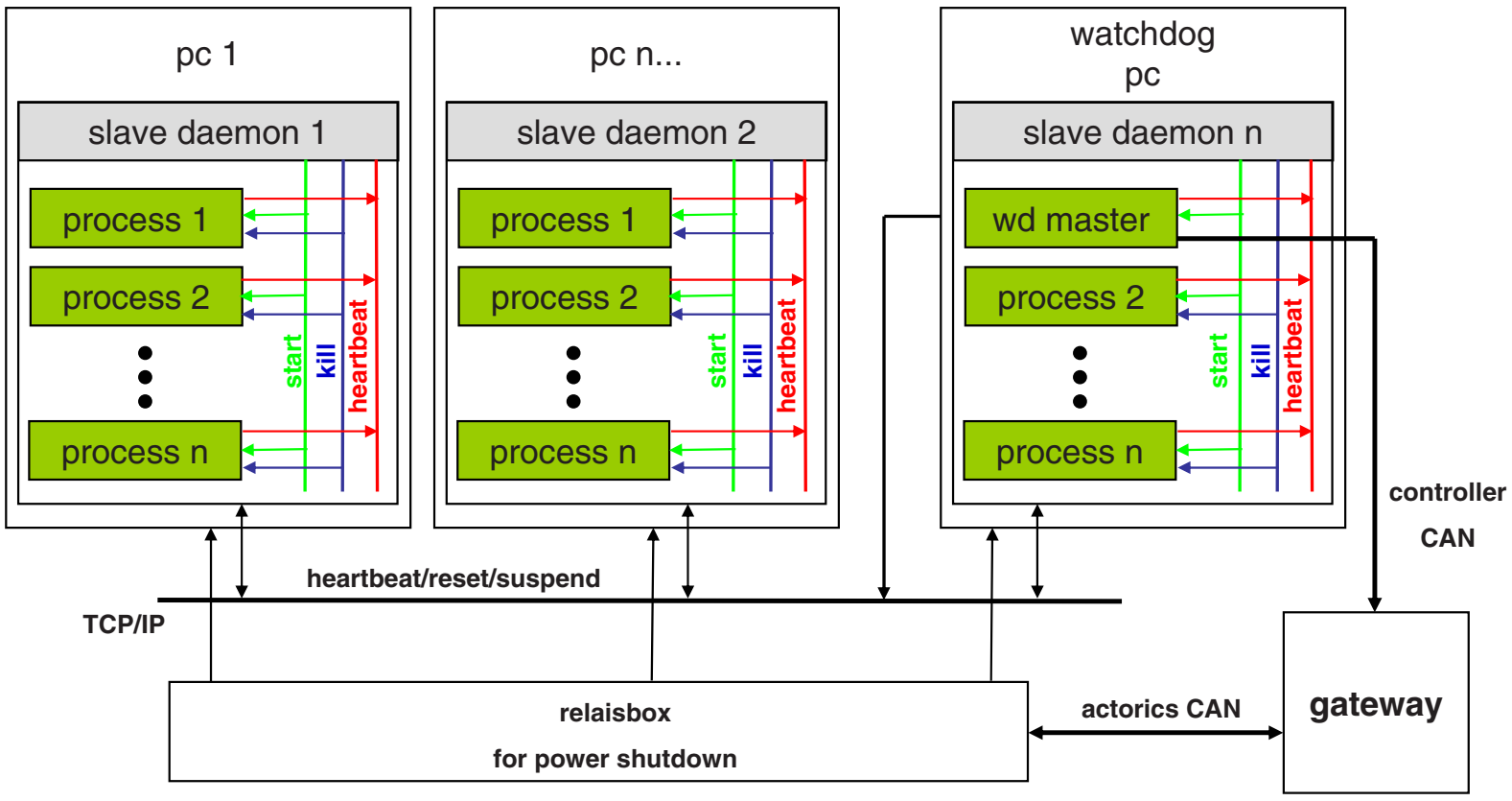

**Fig. 39.** Software watchdog architecture.

## 5 System Development Process

For developing Caroline's software and ensuring its quality, we implemented a multi-level testing process using elements of extreme programming [Beck, 2005] partly realized in an integrated tool chain shown in Fig. 40. The workflow for checking and releasing software formally consists of five consecutive steps. First the source is compiled to check for syntactical errors. While running the test code, the memory leak checker valgrind [Nethercote and Seward, 2003] checks for existing and potential memory leaks in the source code. After the execution of the test code, source code coverage is computed by simply counting the executed statements. The intent is to implement test cases that completely cover the existing source code or to find important parts of the source code that are still lacking test cases. The last step is for optimization purposes only and executes the code in order to find time-consuming parts inside an algorithm.

The tool chain is executed manually by the developer or by using an integrated development environment such as Eclipse. The tool chain itself can be customized by the developer by selecting only necessary stages for the current run, i.e. skipping test suites for earlier development versions of an algorithm. Nevertheless, the complete tool chain is executed every time a new version of the source code is checked in the revision system Subversion [Collins-Sussmann et al., 2004]. Therefore, an independent bugbuster server periodically checks for new revisions on the server. If a new version is found, it is checked out into a clean and safe environment so that the complete tool chain can be run. The results are collected and a report is automatically generated. The report is easily accessible through the project's web portal [Edgewall Software, 2007] for every developer. For measuring the performance or consulting the results of a previous revision, the history of older revisions is kept and accessible via same the web portal.

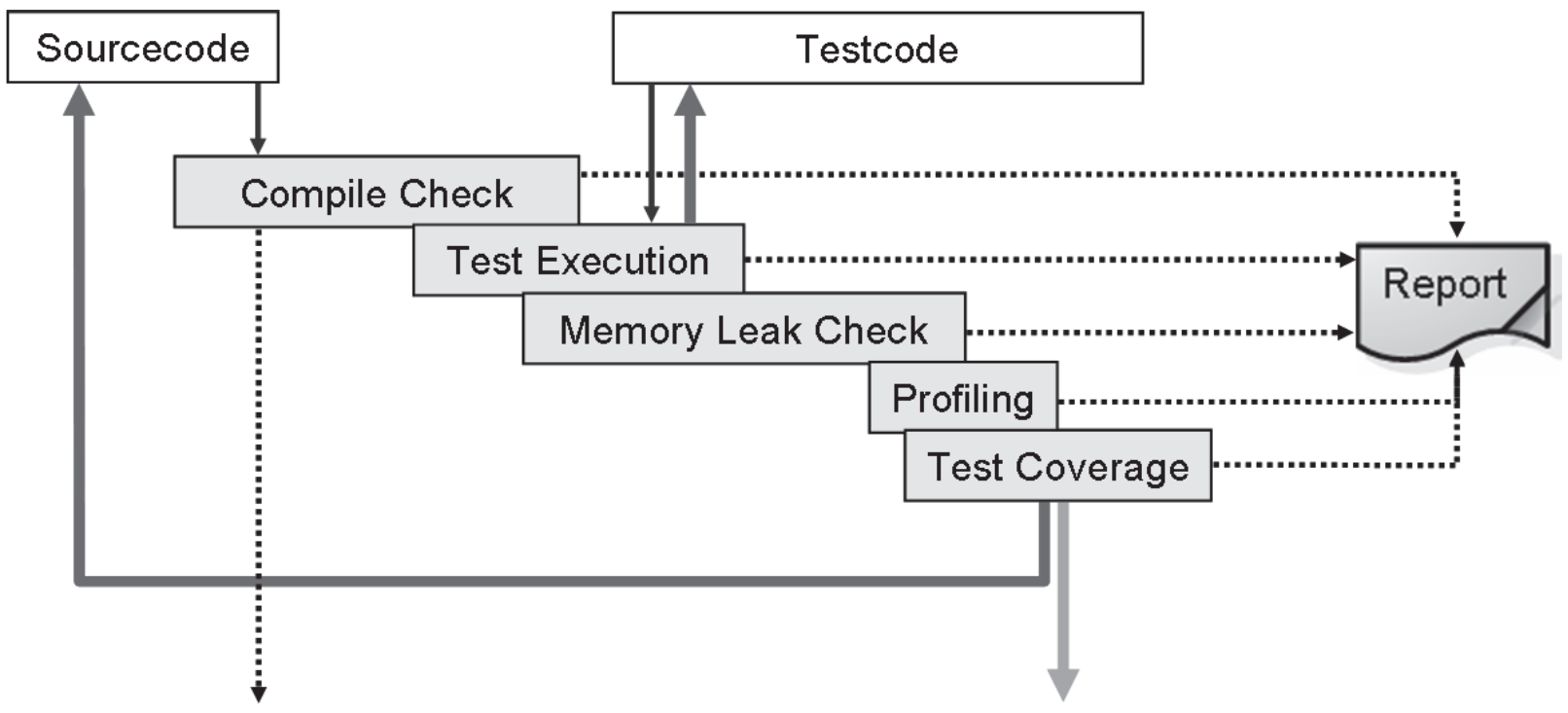

**Fig. 40.** Workflow for testing and releasing software.

The main development process described above mainly covers only unit tests[Liggesmeyer, 2002] for some functions or parts of the complete software system. For the development of Caroline's artificial intelligence, interactive feed back tests using riskless simulations are necessary. Furthermore, the interactive simulations describe different situations for testing the artificial intelligence. After completing the interactive tests, they can be formalized in acceptance tests for automatic execution on another independent server. These test suites are automatically executed after every change to the revision system comparable to the bugbuster server.

The next section describes the simulator development for the CarOLO project. Afterwards, the adoption of the simulator in automatic acceptance tests is explained. This work continues prior work presented in [Basarke et al., 2007a] and [Basarke et al., 2007b].

### 5.1 Simulator

The simulation of various and partly complex traffic situations is the key for developing a high quality artificial intelligence that is able to handle many different situations with different types of preconditions. The simulator provides appropriate feedback to the other parts of the system, by interpreting the steering commands and changing the Ego State and the surroundings.

The simulator can be used for interactively testing newly developed artificial intelligence functions without the need for real vehicle. A developer can simply, safely and quickly test the functions. Therefore, our approach is to provide a simulator that can reliably simulate missing parts of the whole software system. Furthermore, the simulator is also part of an automatic test infrastructure described in the next section.

Figure 41 shows the main classes of the core simulator. The main idea behind this concept is the use of sets of coordinates in a real world model as context and input. These coordinates are stored in the model and used by the simulator. Every coordinate in the model is represented by a simulator object position describing the absolute position and orientation in the world. Every position is linked to a simulator object that represents one single object. These objects can have a variety of behaviors, shapes and other information necessary for the simulation. The model is linked with a simulator control that supervises the complete simulation. The simulator application itself controls the instantiation of every simulator component by using object factories.

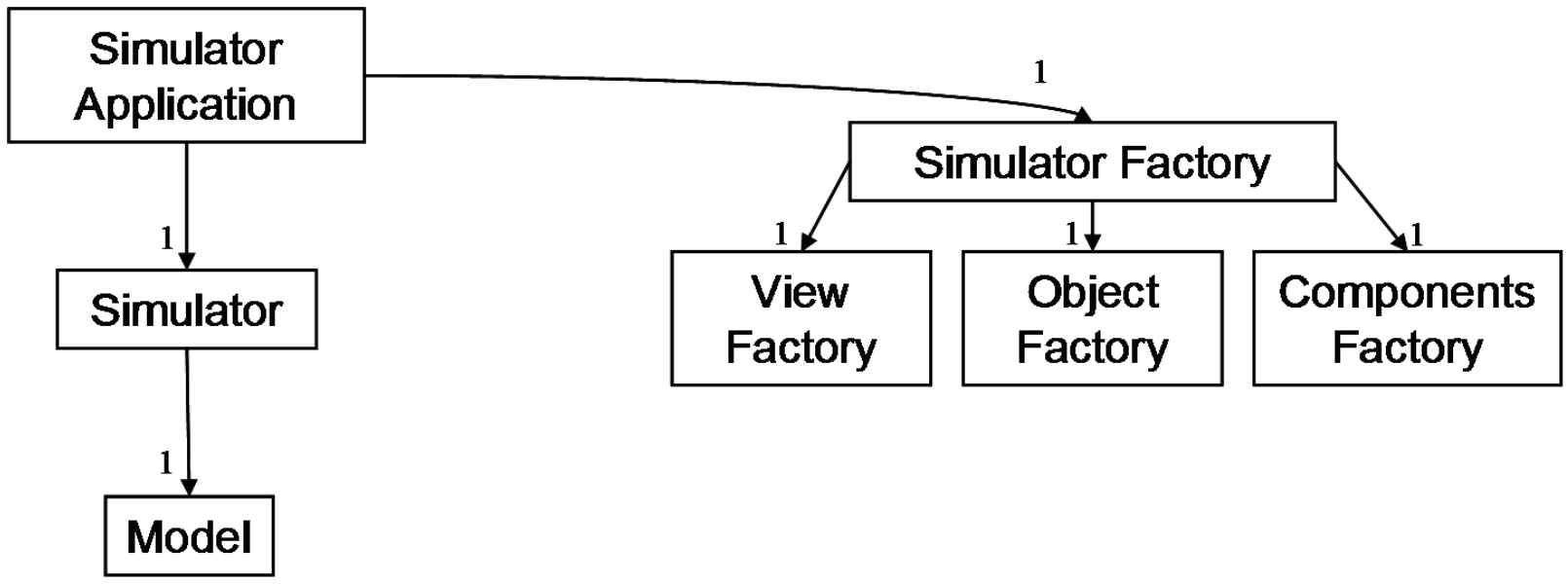

**Fig. 41.** Main classes of the simulator.

Figure 42 shows the factories in detail. The simulator view encapsulates a read-only view of an extract of the world model. Every simulator view is linked with a simulator components group. A component represents missing parts of the whole system like an actorics module for steering and braking or a sensor data fusion module for combining measured values and distributing the fused results. Thus, every component in the components group can access the currently visible data of the core data model by accessing the simulator view. As mentioned above, every simulator object position is linked with a simulator object, each of them equipped with its own configuration. Thus, every component can retrieve the relevant data of the owned simulator object.

The main task of the simulator is to modify the world model over time. For simulating the world it is necessary to proceed a step in the simulation. A simulation step is a function call to the world model with the elapsed time step $\delta t_i > 0$ as a parameter that modifies the world model either sequentially or in parallel.

A simple variant is to modify every simulator object sequentially. In this variant, the list of simulator objects is addressed through an iterator and then modified using original object data. Although this is an efficient approach, it is not appropriate when the objects are connected and rely on behaviors from other objects. Another possibility is to use the algorithms as if a copy of the set of simulator object positions were created. While reading the original data, the modification uses the copy and thus allows a transaction such as a stepwise update of the system, where related objects update their behavior together.

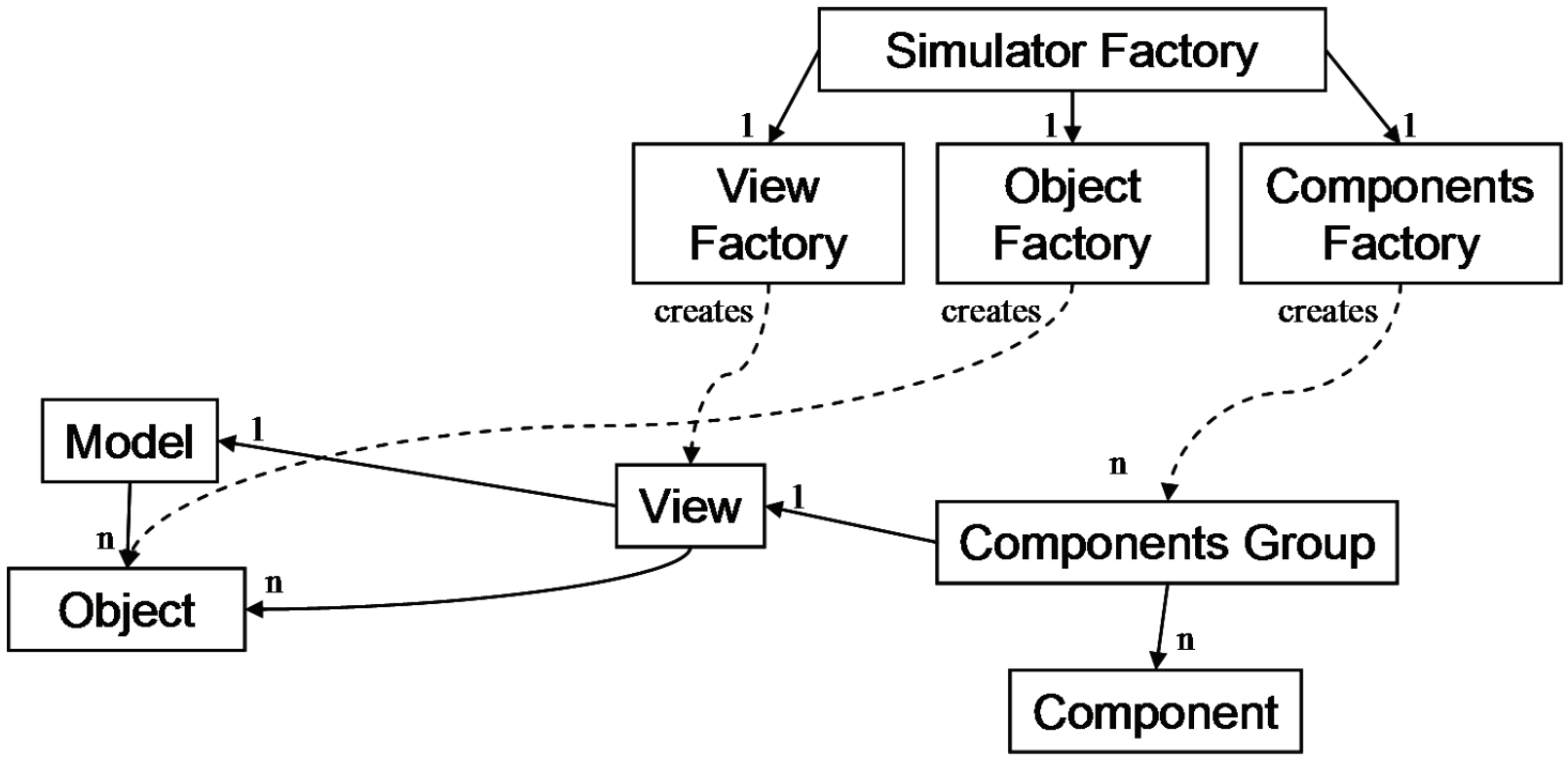

**Fig. 42.** Object factories creating the simulator's surroundings.

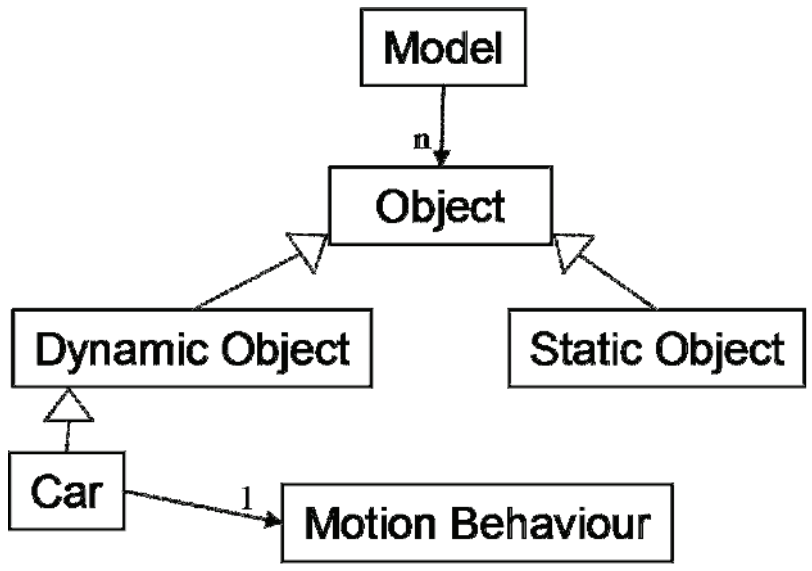

**Fig. 43.** World's model and motion behavior interface.

For modifying an object in the world model, every non-static object in the world model uses an object that implements the interface MotionBehavior as shown in Fig. 43. A motion behavior routine executes a simulation step for an individual object. A simulator component implementing a concrete motion behavior registers itself with the simulator object. For every simulation step the simulator object must call the motion behavior and therefore enables the behavior implementation to modify its own position and orientation according to a simulator component. The decoupling of objects and their motion behavior allows us to change the motion behavior during a running simulation, i.e. because of weather influences. Furthermore, it simplifies the implementation of new motion behaviors at development time. For testing Caroline, we have developed additional motion behaviors like MotionBehaviorByKeyboard for controlling a virtual car in the interactive mode by using keys or a MotionBehaviorByRNDF that controls a car in its surroundings by using a predefined route to follow.

The most interesting motion behavior however, is the MotionBehaviorByTrajectory because it communicates directly with the artificial intelligence. For the best imitation of the behavior of the real car, the simulator uses the same code as the vehicle control module based on trajectories expressed as a string of pearls that form consecutive gates. Furthermore, the motion of the simulated car is computed with 3rd order B-splines such as the vehicle controller module. Using a B-spline yields smoother motion in the simulation and a driving behavior sufficiently close to reality - if it is taken into account that for intelligent driving functions it is not necessary to handle the physical behavior in every detail, but in an abstraction useful for an overall correct behavior.

Using motion behaviors, it is possible to compose different motion behaviors to create a new composed motion behavior. For example, it is possible to build a truck with trailer from two related, but only loosely coupled objects. A composition of the motion behaviors yields a new motion behavior that modifies the position and orientation of the related simulator objects according to inner rules as well as general physical rules.

Getting such a simulator up and running requires quite a number of architectural constraints for the software design. One important issue is that no component of the system being tested tries to call any system functions directly, like threading or communication, but only through an adapter. Depending on whether it is a test or an actual running mode, the adapter decides if the function call is forwarded to the real system or substituted by a result generated by the simulator. Because of the architectural style, it is absolutely necessary that no component retrieves the current time by calling a system function directly. Time is fully controlled by the simulator and therefore knows which time is relevant for a specific software component if different times are used. Otherwise, time-based algorithms will become confused if different time sources are mixed up.

### 5.2 Quality Assurance

As mentioned at the beginning of this section, the simulator is not only used for interactive development of the artificial intelligence. It is also part of a tool chain that is automatically executed on an independent server for assuring the quality of the complete software system consisting of several modules. In the CarOLO project, we analyzed the DARPA Urban Challenge documents to understand the requirements. These documents contained partly functional and non-functional definitions for the necessary vehicle capabilities. In every iteration a set of tasks consisting of new requirements and bugs from previous iterations is chosen by the development team, prioritized and concretely defined using the Scrum process for agile software engineering [Beedle and Schwaber, 2002]. These requirements are the basis for both a virtual test drive and a real test of Caroline.

After designing a virtual test drive the availability of necessary validators is checked. A validator is part of the acceptance tool chain and responsible for checking the compliance of the artificial intelligence's output with the formal restrictions and requirements. Validators implementing intelligent software functions are used to automatically determine differences in the expected values in the form of a constraint that cannot be violated by the test. A validator implements a specific interface that is called up automatically after a simulator step and right before the control flow returns to the rest of the system. A validator checks, for example, distances to other simulator objects, validates whether a car has left its lane or exceeded predefined speed limits. After an unattended virtual test drive, a boolean method is called upon to summarize the results of all test cases. The results are collected and formatted in an email and web page for the project's web portal.

The set of validators covers all basic requirements and restrictions and can be used for automatically checking the functinality of new software revisions.

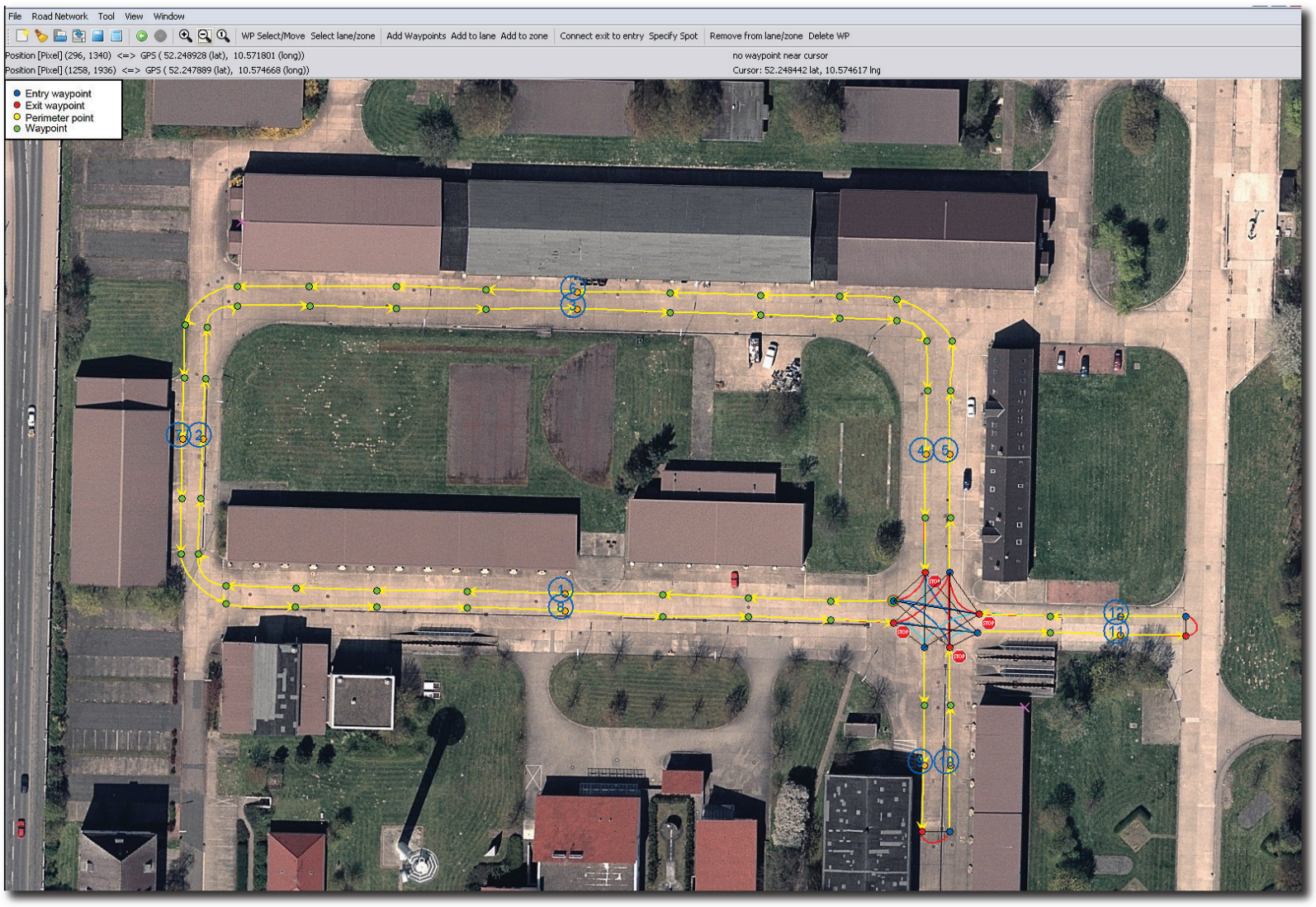

**Fig. 44.** Screenshot of the GUI tool for constructing RNDFs.

The main benefit is that these high level tests are black-box tests and do not rely on the internal structure of the code. Thus, a subgroup of the CarOLO team was able to develop these high level acceptance tests without a deep understanding of the internal structures of the artificial intelligence. Using this approach, more complex traffic situations could be modeled and repeatedly tested without great effort.

To allow for the quick and convenient creation of test scenarios, various concepts and tools have been developed. The following describes how virtual test drives are defined as well as how certain surroundings such as data fusion objects or drivability data is generated and fed into the simulator. To make this clear we briefly illustrate the proceedings on a basis of an example, which deals with the simple passing maneuver as already described in section 4.3.3. Assume we would like to determine wether the artificial intelligence is able to recognize static obstacles in our travel lane and reacts properly by adhering the required minimum distances.

First, an RNDF must be created that contains information about existing lanes, intersections, parking spots and their connections. As an RNDF provides the basis for every test run, many of those route network definitions had to be created. Therefore we developed a GUI tool to simplify the creation of RNDFs as shown in Fig. 44.

Several features including dragging waypoints, connecting lanes and adding stop signs or checkpoints speed up the construction process. Completed RNDFs could be exported to a text file and used as input for the artificial intelligence as well as for the simulator.

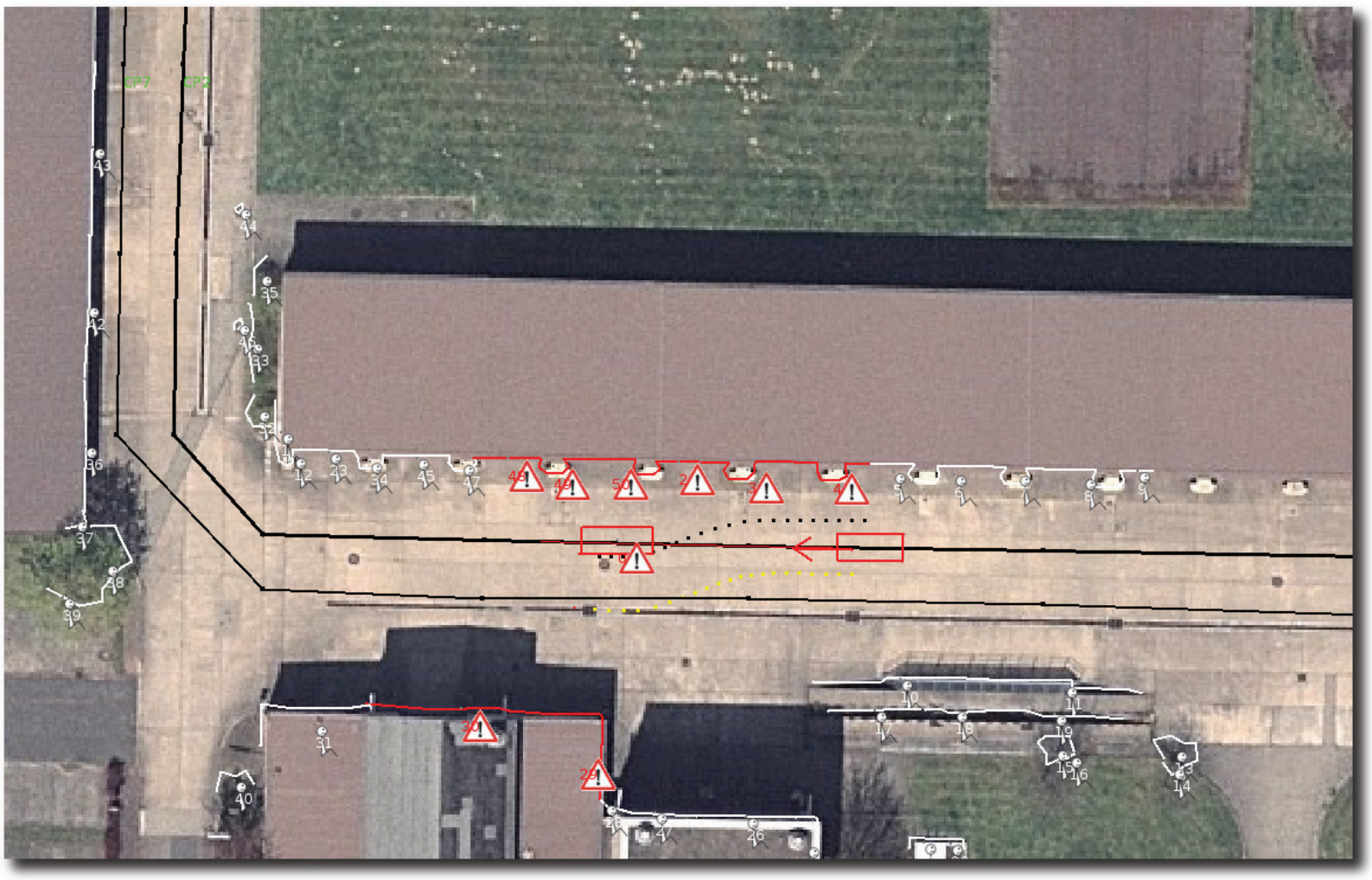

**Fig. 45.** Screenshot with fusion objects.

The purposes of RNDFs within the simulator vary in different ways. One purpose is to check the behavior of the artificial intelligence concerning the RNDF provided and the actual lane. Therefore a second RNDF can be passed to the simulator. The additional and independent RNDF is used to provide lane data, which is normally detected by the computer vision system. This is especially important if there are major differences between the linear distance and the actual route to the next waypoint.

Another use of RNDFs is to define the behavior of dynamic obstacles during the test run, as mentioned earlier. Thus we are able to check relevant software modules for their interaction with dynamic obstacles. This approach is similar to the one used for providing detected lanes. Dynamic obstacles are interacting on a basis of their individual RNDFs by using the MotionBehaviorByRNDF. This concept can be used for simulating scenarios at intersections and even more complex traffic scenarios.

To extend the example of passing a static obstacle we need to create suitable data, which could be translated to sensor fusion objects. Two principal approaches are available to achieve this goal. Generating scenarios with static obstacles can be accomplished by using our visualisation application, which provides the ability to define polygons or by using a drawing tool. Shapes of fusion objects could be exported to a comma-separated file. The simulator parses the textual representation of polygons and translates them to fusion objects to be processed by the artificial intelligence. The use of a drawing tool implies the use of predefined colors. The positions of static obstacles are computed by scanning the created image for special markers with reference to a known coordinate. Fig. 45 depicts a screenshot of our visualisation application where the corresponding fusion objects are displayed.

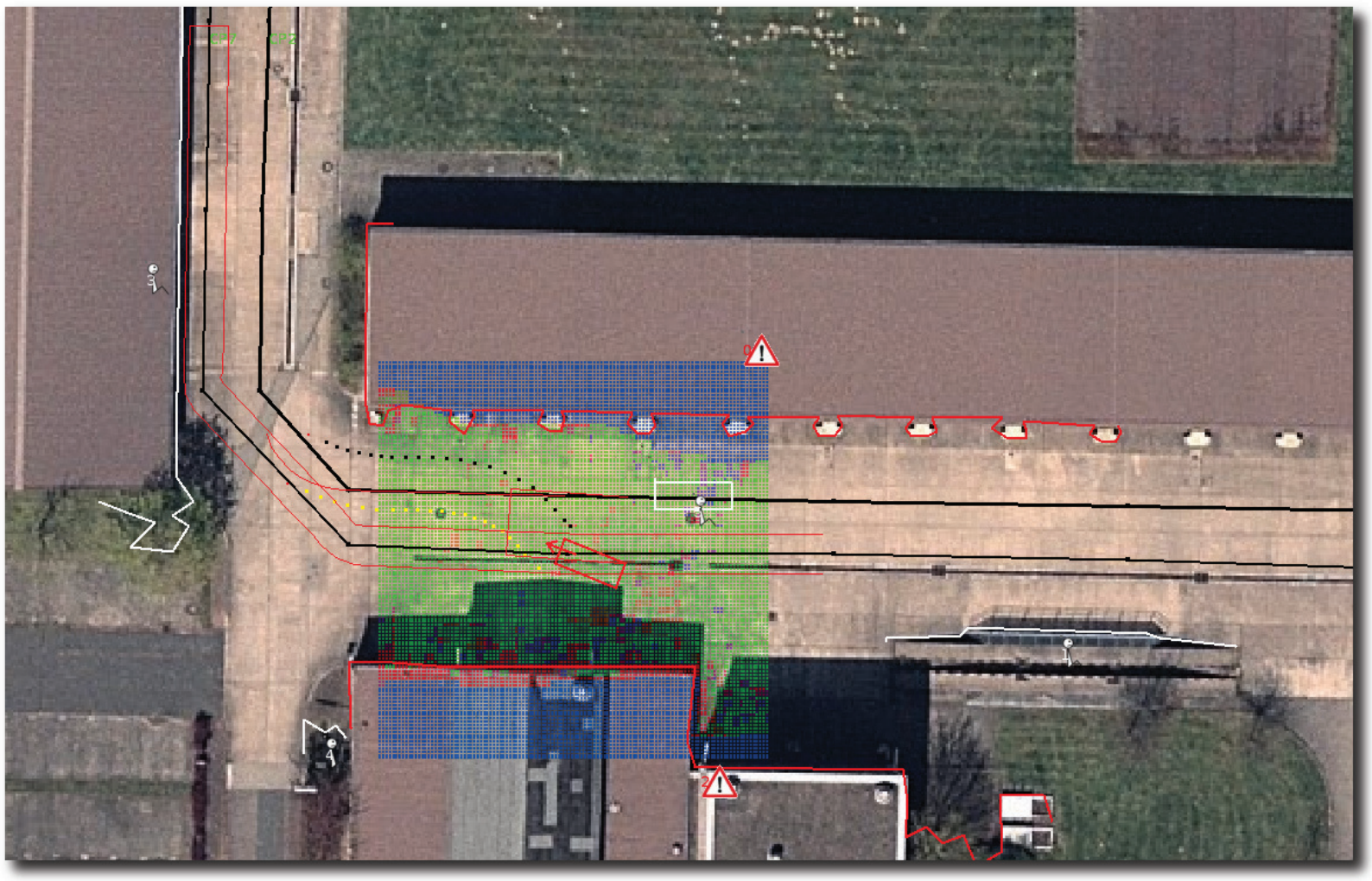

**Fig. 46.** Sreenshot with additional drivability data.

For a more realistic simulation, the data fusion objects generated by the simulator could be created with different quality. This is used to simulate sensor noise and GPS drifts and makes fusion objects suddenly disappear or moves them by a tiny offset away from their original location. The sensor visibility range could be specified to affect the range of fusion objects that will be transmitted to the artificial intelligence.

Adding moderate drivability data completes this test run. This could be accomplished by passing an image file to the simulator, which specifies the required information through different colors. Fig. 46 shows the result. The visualisation of drivability grid displays drivable terrain in green, undrivable terrain in red and unknown terrain with blue cells.

# 6 The Race and Discussion

## 6.1 National Qualification Event

The National Qualification Event took place from October 26 to October 31 on the former George Airforce Base in Victorville, California as depicted in Fig. 2. The entire area was divided into three major parts named "Area A", "Area B" and "Area C" as shown in Fig. 47. First of all, Caroline had to demonstrate the proper function of her safety system to participate in the National Qualification Event. As expected Caroline stopped within the necessary range using the E-stop remote controller as well as the emergency stop buttons mounted on her roof.

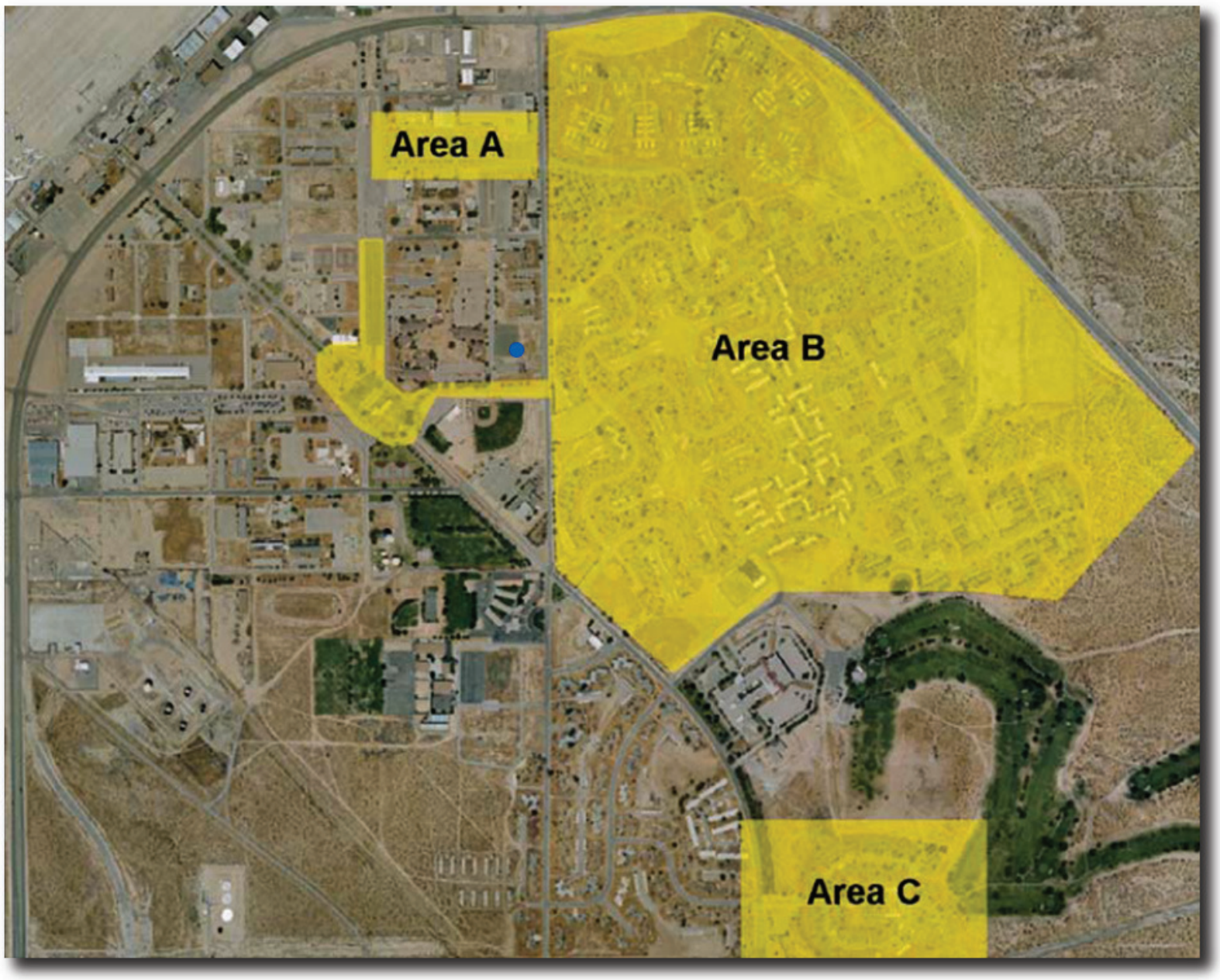

**Fig. 47.** Layout of the former George Airforce Base for the National Qualification Event. The blue dot indicates the pit area for our team.

### 6.1.1 Area A

For our team, the National Qualification Event started in "Area A". The main task for Caroline in that part was to merge into and through moving traffic. Therefore, several other vehicles controlled by human drivers drove within predefined speed limits to ensure the 10 seconds time slots as demanded by the DARPA's requirements. Fig. 48 shows the layout of the track. Caroline was placed at checkpoint 2. She had to drive downward to the T-junction, wait for an appropriate time slot and then turn left through the moving traffic. Afterwards, she had to pass checkpoint 1 by following other vehicles and drive to the upper junction. After waiting for an appropriate time slot, she had to turn into the street to pass checkpoint 2 again. The goal was to drive as many rounds as possible in this area.

Compared to other competitors, Caroline had to pass this task several times. The first run in this part let Caroline drive into the opposite lane. Analyzing this obviously strange behavior afterwards using our simulator as depicted in Fig. 49, we figured out that the barriers shown by white lines around the course narrowed the proper lane. Therefore, Caroline, shown as a red rectangle driving downwards to the lower T-junction, interpreted them as stationary obstacles in her way which she tried to overtake which can be seen in the computed trajectory shown by yellow and black pearls that leads into the opposite lane.

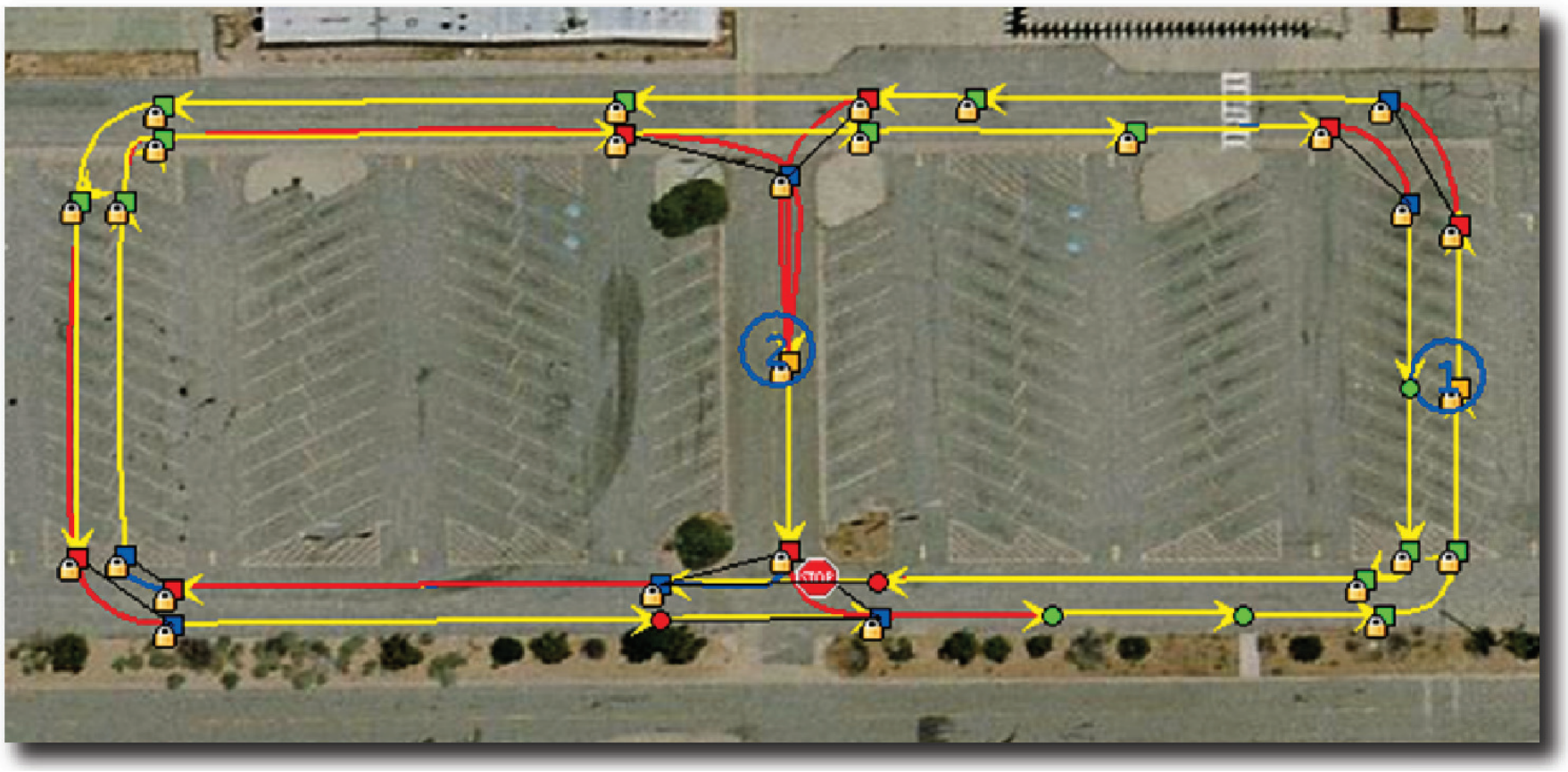

**Fig. 48.** Layout of "Area A".

After modifying several parameters, we had our second try in "Area A". She drove five rounds, merged into moving traffic correctly, waited at stop lines and followed other vehicles very well. Unfortunately, some problems occurred on the above right corner, when Caroline decided to turn right instead of following the road to the junction. We found out, that Caroline got in trouble with the street surface in that corner. There was a mixture of concrete and tar each with different colors. Thus, Caroline educated that color difference and tried to drive towards areas with a similar surface.

After modifying that behavior, we got another try in that course. Caroline started a perfect first run but waited too long for the second one. Therefore, the judges paused our vehicle and demanded a more progressive behavior of Caroline. Tuning again some parameters, we tried the course a fourth time short time later. This time, Caroline drove very swiftly but she did not give way to oncoming traffic. So, we changed the parameters again to get a safer behavior again and convinced the judges in our last try in that area of Caroline's abilities to merge correctly into moving traffic after demonstrating approximately eigth perfect rounds.

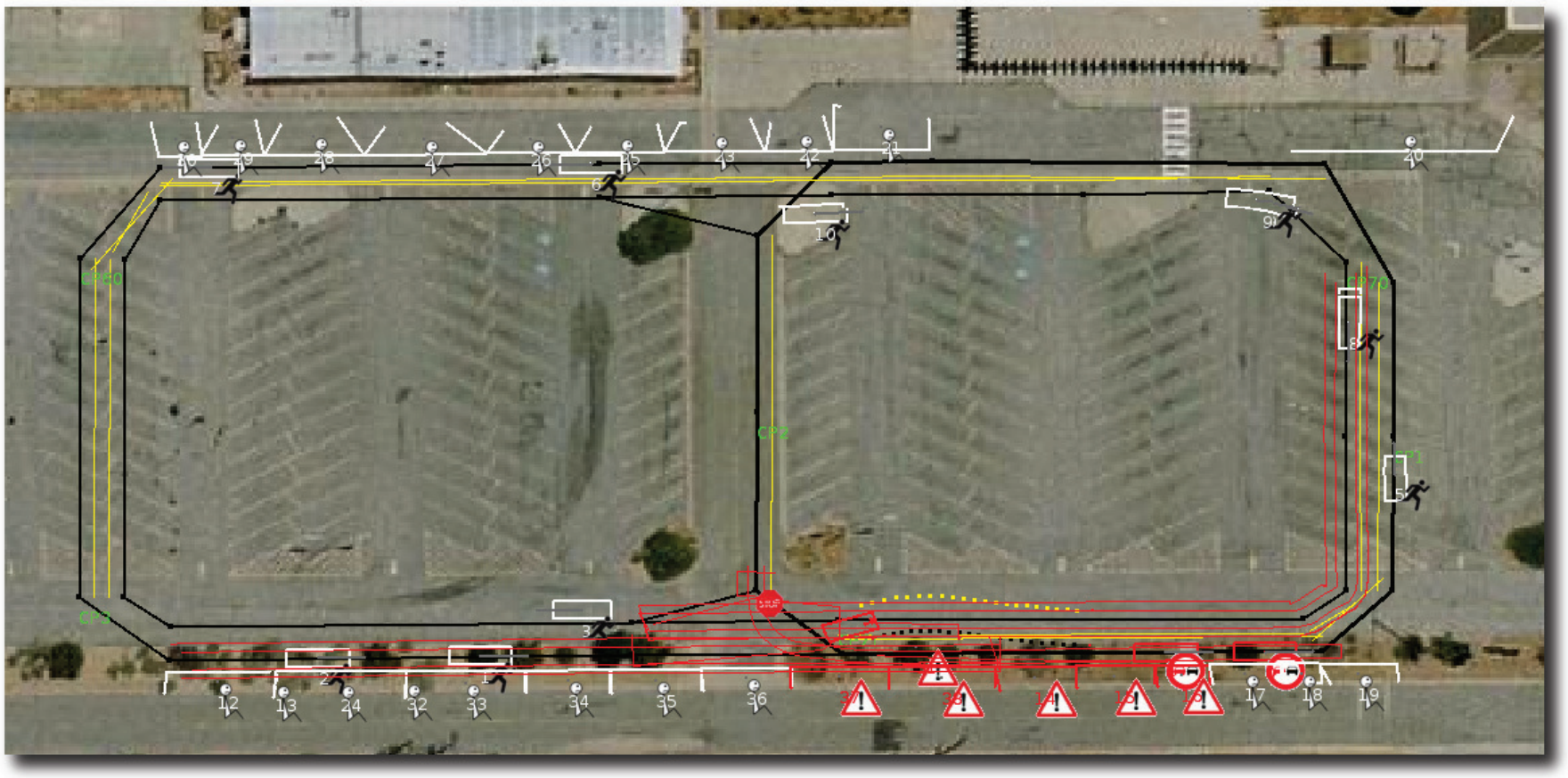

**Fig. 49.** Analysis of Caroline's behavior in "Area A".

### 6.1.2 Area B

After encountering difficulties in the first task, we were unsure how Caroline would perform in "Area B" since several teams already failed to complete this part. The entire course is shown in Fig. 50. The main task was to overtake stationary obstacles, handle free navigation zones without any lane markings and to park safely inside those zones between other vehicles. The course itself could not be seen completely, so Caroline had to drive for herself without any observation by our team. We only could hear her progress by the team radio and by her siren.

Caroline started within a concrete start chute laid inside a free navigation zone. Many other teams already failed to leave this zone into the traffic circle correctly. She entered smoothly the traffic circle, left the circle and turned into the part on the right hand side of Fig. 50. In the center of the lower circle she had to park between other vehicles. The entry to that zone was very rough and several other teams already damaged the tires of their vehicle. We analyzed the video right after the task and remarked heavy vibration of the camera's picture but she entered the zone smoothly. After finishing the parking she left the zone to proceed the course.

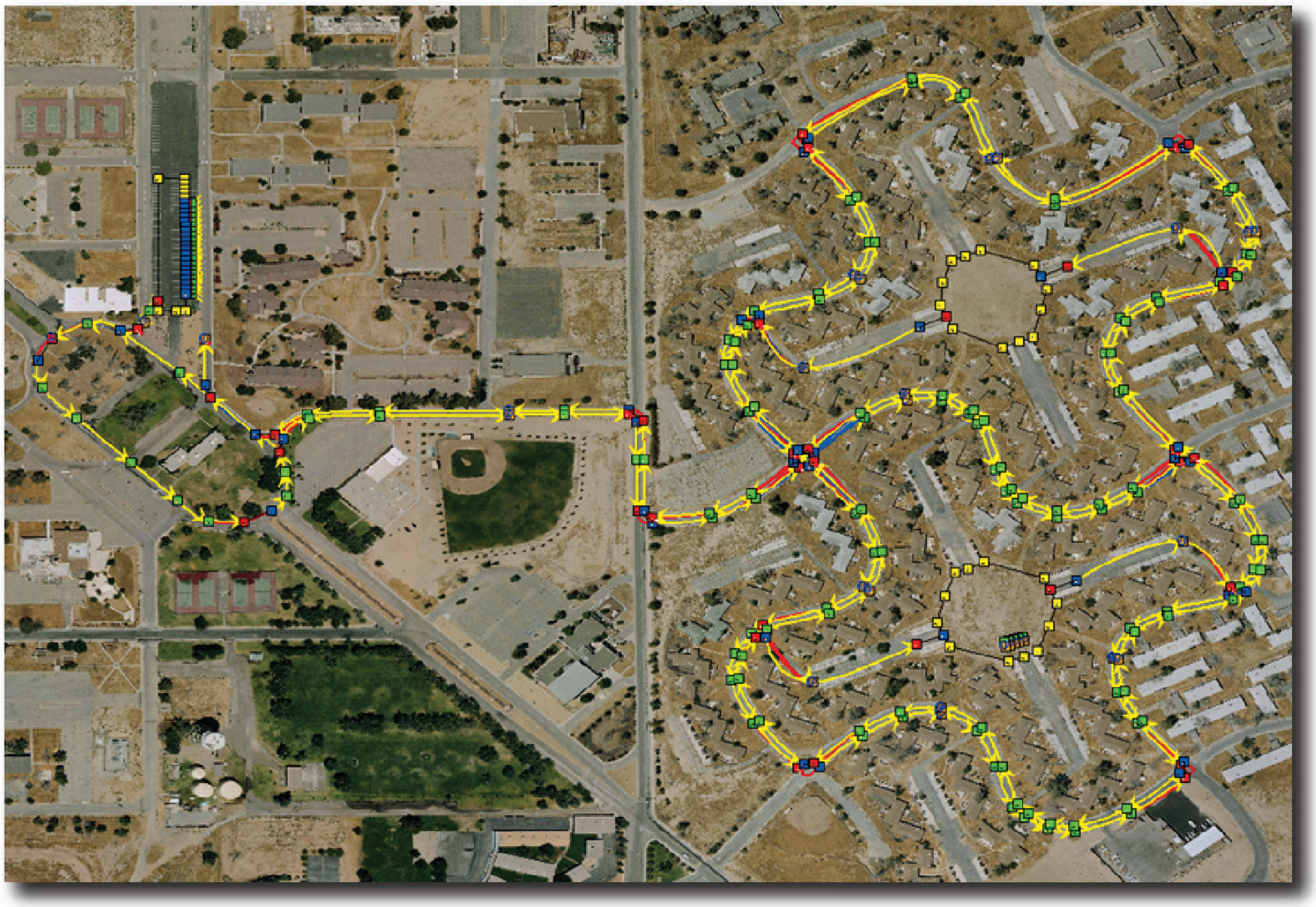

**Fig. 50.** Layout of "Area B".

Furthermore, Caroline had to deal with a gate located right at the exit of the upper circle. Due to our sensor layout she had to attempt several times to find the right way for leaving that circle. Returning to the start chutes again, she honked twice to indicate the completion of her mission after passing the last checkpoint. With this successful run, Caroline was one of only three vehicles to accomplish this course completely and in time.

### 6.1.3 Area C

On the same day, Caroline was faced with "Area C". This area is shown in Fig. 51. The main task was to handle intersections correctly and deal with blocked roads.

Caroline started near checkpoint 30 in the upper left corner on the outer lane. She handled both intersections on the left hand side and the right hand side several times correctly with every combination of other vehicles she was faced. Right in front of checkpoint 30 in the center part of this course, Caroline encountered a road blockage as shown in Fig. 52. We were unsure wether Caroline would detect the barrier since it had no contact to the ground and our sensors could look right through that barrier.

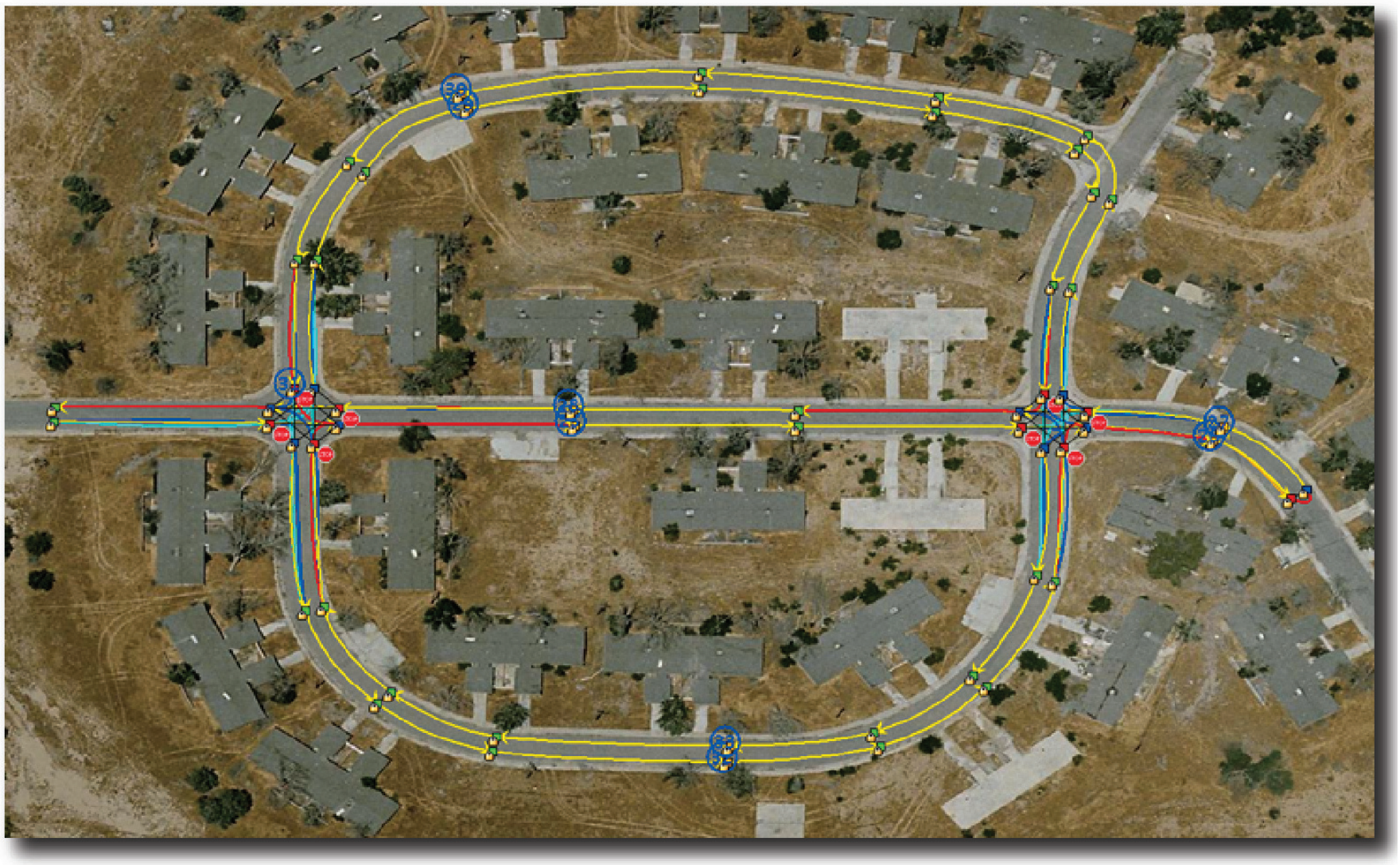

**Fig. 51.** Layout of "Area C".

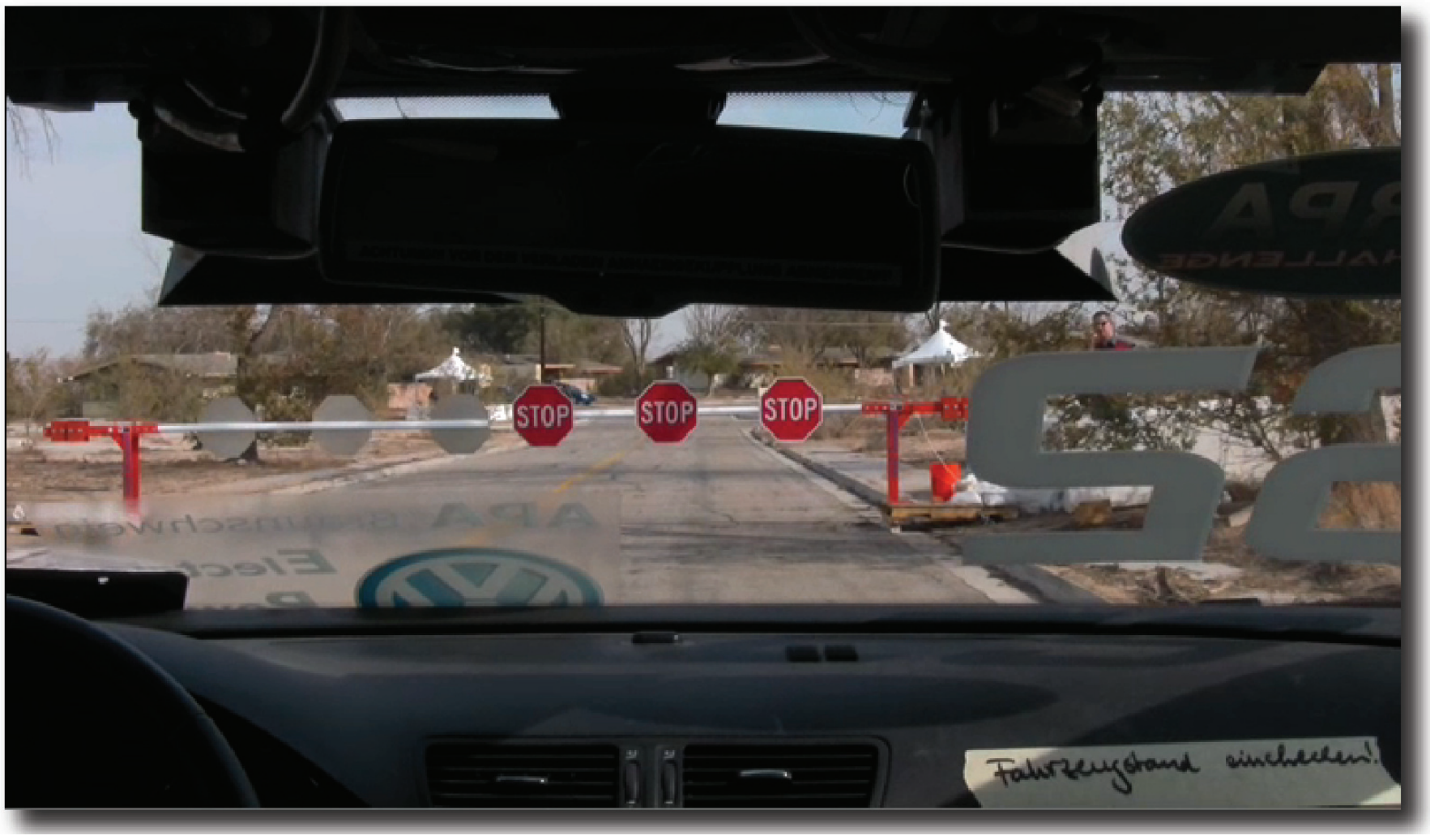

**Fig. 52.** Blocked round in "Area C" by a barrier.

But Caroline detected that barrier properly and initiated the U-turn to choose another route the checkpoint. Afterwards, she passed all further traffic and intersection situations correctly and finished "Area C" finally. With all results achieved in the three areas, Caroline qualified early as a newcomer for the final event besides the well-established team with their experience of the Grand Challenges.

## 6.2 Mandatory Practice for DARPA Urban Challenge Final Event

The day before the DARPA Urban Challenge Final Event was scheduled, everyone of the eleven finalists had to participate in a practice session. By using this session, DARPA would ensure that every vehicle was able to leave the start chute and turn into the traffic circle. Assuming that this would be an easy task, we put Caroline into autonomous mode and waited for her to begin her run. But she did not leave her start chute and our team failed that practice session. We figured out a problem by parsing the RNDF provided by the DARPA. This issue did not let Caroline understand the road network for the final. After fixing this problem, we got another try. But Caroline still did not leave her start chute. Thus, DARPA placed us in the last of the eleven start chutes and cancelled the practice for our team.

Later analyzing the data we figured out the jitter in the GPS signal while significantly waiting for the "RUN" mode that yielded leaving the calculated trajectory. After fixing this issue we finally prepared Caroline for the DARPA Urban Callenge Final Event on the following day.

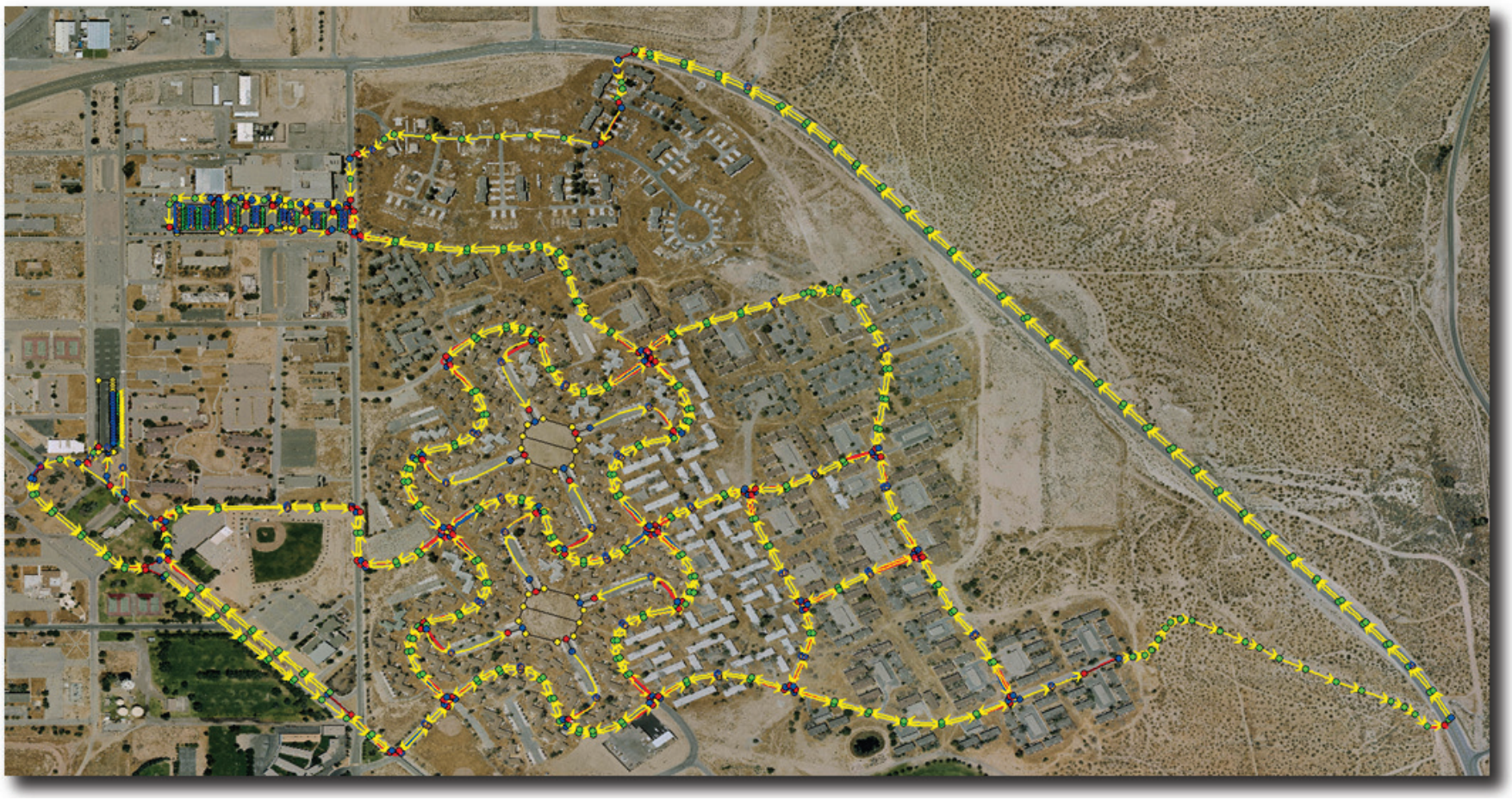

**Fig. 53.** Layout for the DARPA Urban Challenge Final Event.

### 6.3 DARPA Urban Challenge Final Event

Figure 53 shows the enlarged "Area B" track for the DARPA Urban Challenge Final Event, including the former "Area A" as a parking lot. The start chutes were the same as for the run in "Area B". Additionally, in the lower-right corner of the map, there was a sandy off-road track located yielding a two-lane road return the inner part of the DARPA Urban Challenge Final Event area.

On November 3, 2007 at 6:53 am PST we loaded the first of three mission files into Caroline and set her into "PAUSE" mode. She calculated the route for the first checkpoint and started her run at 7:27 am PST. Fig. 54 shows the first part of her way during the first mission.

The asterisk in Fig. 54 indicates the location where two members of our team had to accompany the DARPA judges. Caroline had passed approximately 2.5 kilometers until she was paused by the DARPA control vehicle right behind her. Fig. 55 shows the reason for "PAUSE" mode.

Caroline got stuck after she turned into the berms. Fig. 55 (a) and (b) shows Caroline approaching a traffic jam right in front of her. Obviously, she tried to pass the stopped vehicle by interpreting it as a stationary obstacle using the clearance next the last car. The result of this attempt is shown in Fig. 55 (c): Caroline got stuck and could not get free without human intervention.

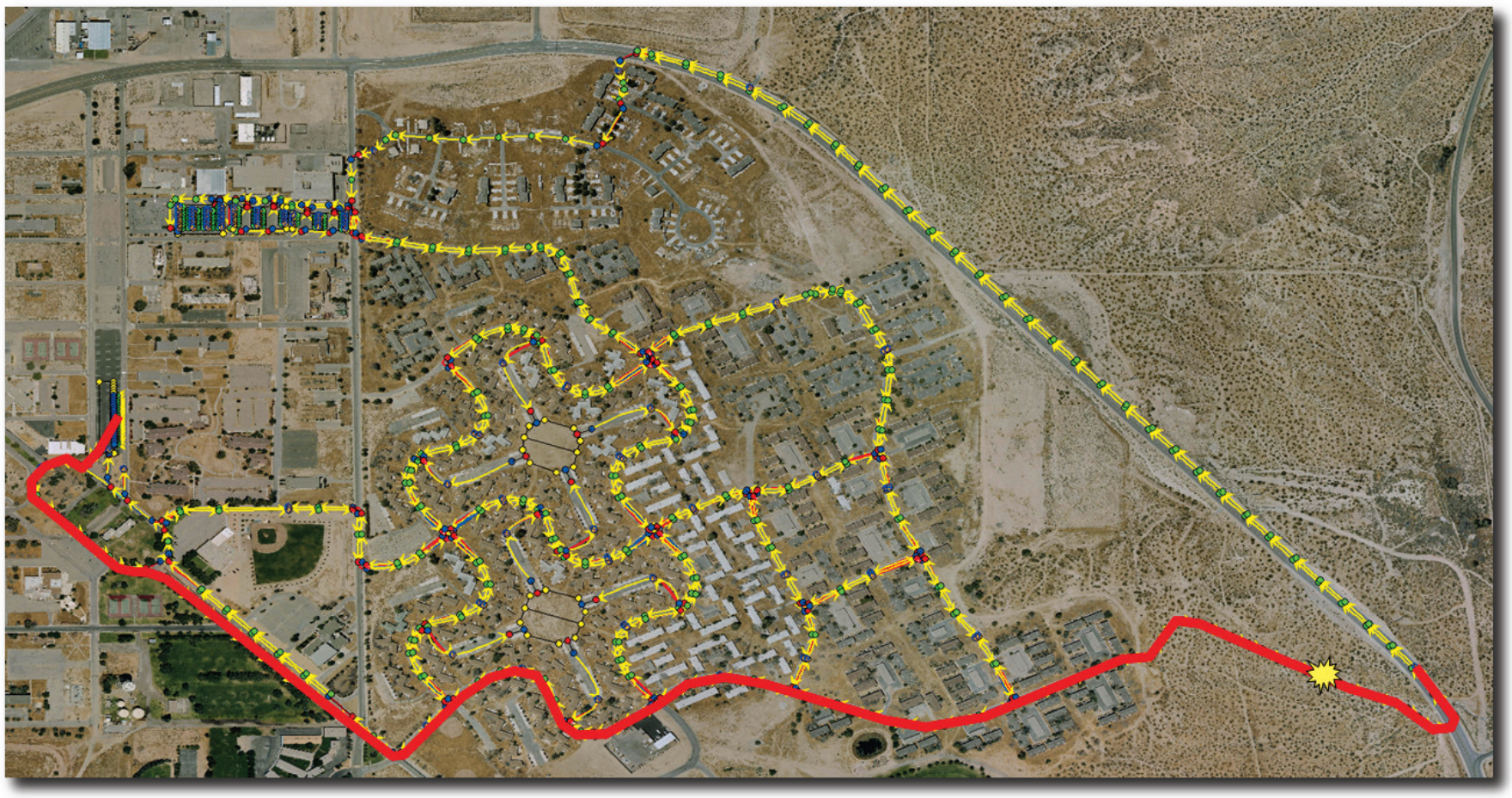

**Fig. 54.** Passed way before the first problem.

After she got freed and set in "RUN" mode again right at the beginning of the two-lane road, she continued her route and passed several checkpoints. The next incident was after 11.4 kilometers shown as the asterisk in Fig. 57.

At that location Caroline did not yield right of way to Talos, the autonomous vehicle from team MIT. Therefore, the DARPA paused both vehicles and let team members from MIT come to that location. After replacing Talos, both vehicles were sequentially set to "RUN" mode and passed safely each other. Unfortunately, the reason for not yielding right of way to Talos could not be figured out analyzing our log files. Since the situation was a left turn through oncoming traffic, it could be a problem detecting and tracking Talos due to problems either with our front sensors or with the interpretation in the artificial intelligence.

As shown in Fig. 58, Caroline continued her route. Additionally, she parked in the parking lot shown in the upper left picture of Fig. 58. After the parking

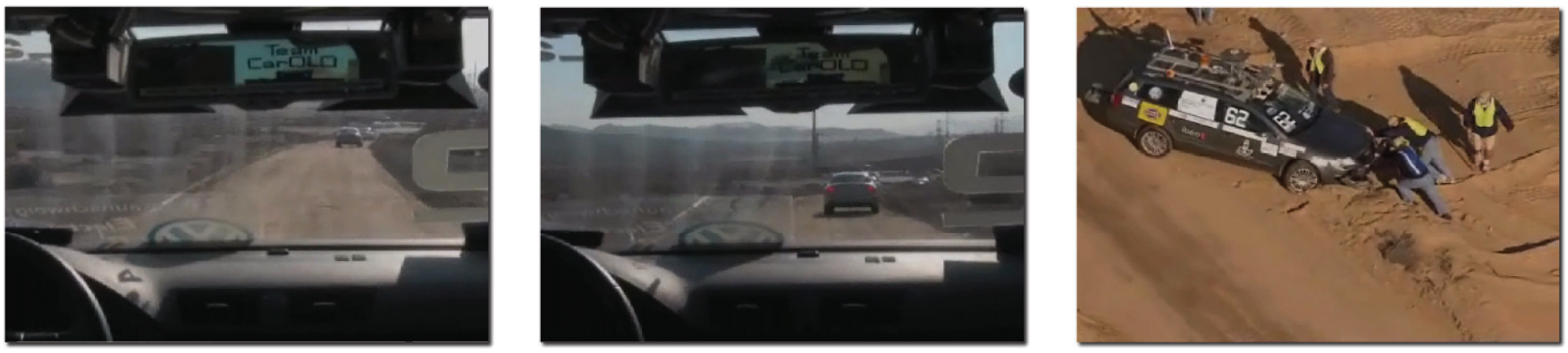

(a) Blocked road in front of Caroline.

(b) Caroline is approaching the traffic jam.

(c) Caroline got stuck after turning left into the berms.

**Fig. 55.** Caroline got stuck after 2.5 kilometers.

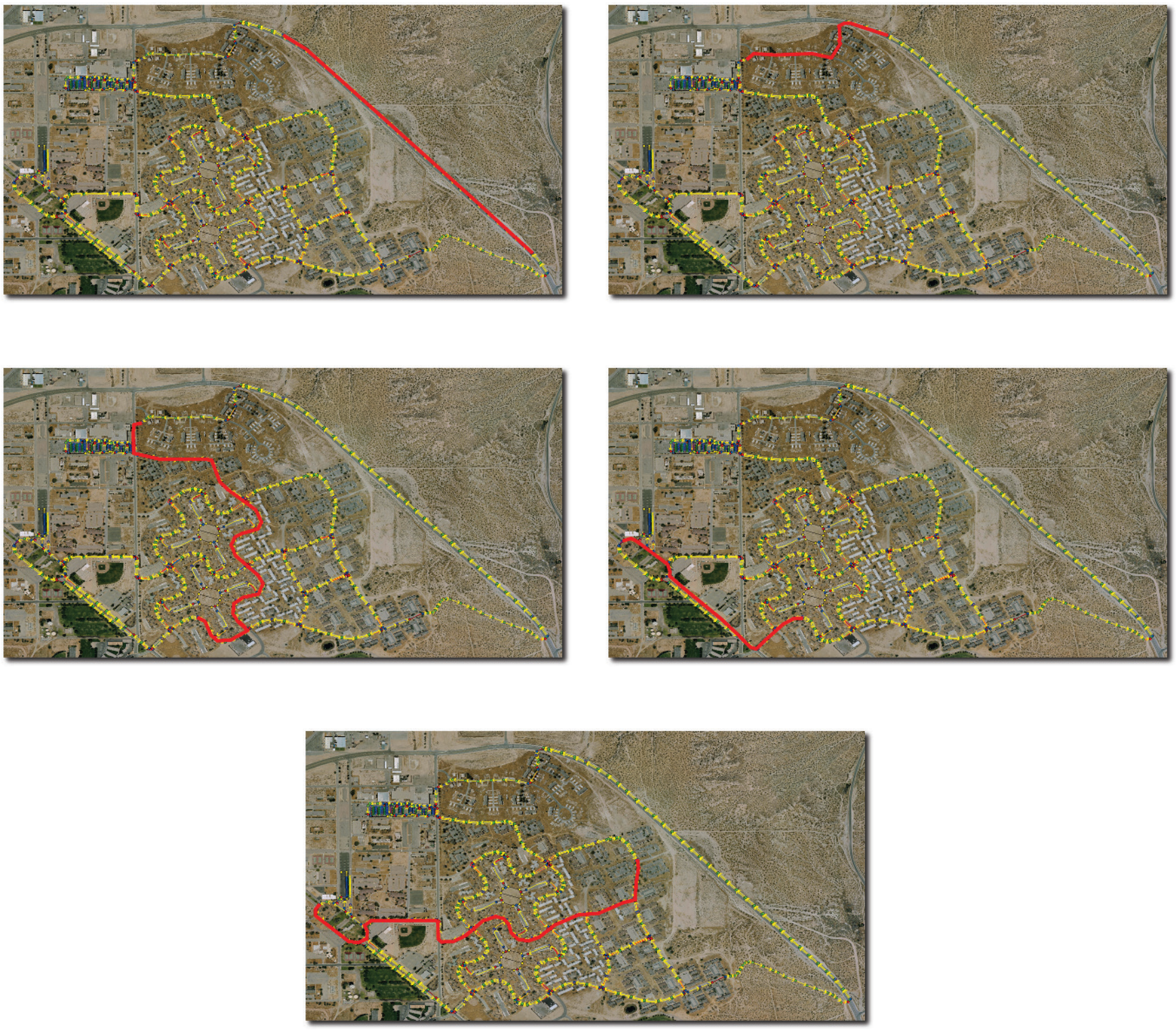

**Fig. 56.** Caroline went on after she got stuck.

maneuver, she returned the second time to the traffic circle and continued her mission 1.

At approximately 9:55 am PST, again two team members from team CarOLO were driven to Caroline, who met Talos from team MIT for the second time in a free navigation zone. This incident is shown as an asterisk in Fig. 59.

Our team members were faced with a twisted carrier rod of the Ibeo laser scanners due to a collision with Talos from team MIT as shown in Fig. 60. Until today it is still unresolved which car was in charge of the accident. Caroline interpreted the situation as described in the technical evaluation criteria [DARPA, 2006] by the section "Obstacle field". Therefore, Caroline tried to pass the oncoming Talos by pulling to the right side. Unfortunately, further interpretation is impossible due to missing detailed log files of that situation. Finally, DARPA retired Caroline as the fourth and last vehicle from the DARPA Urban Challenge Final Event.

Altogether, Caroline drove 16.4 kilometers in total and was retired from the race at 10:05 am PST. At 8:03 am PST, the watchdog module reset the SICK laser scanners mounted on the roof due to communication problems. At approximately 9:00 am PST, the watchdog missed heartbeats from the IMU, and therefore triggered a reset. Right after the collision with Talos from team MIT, the watchdog observed communication problems with the laser scanners mounted in the front of Caroline. After a reset, the communication was re-established. During the race, computer "Daq1" as shown in Fig. 4 froze two times and had to be reset.

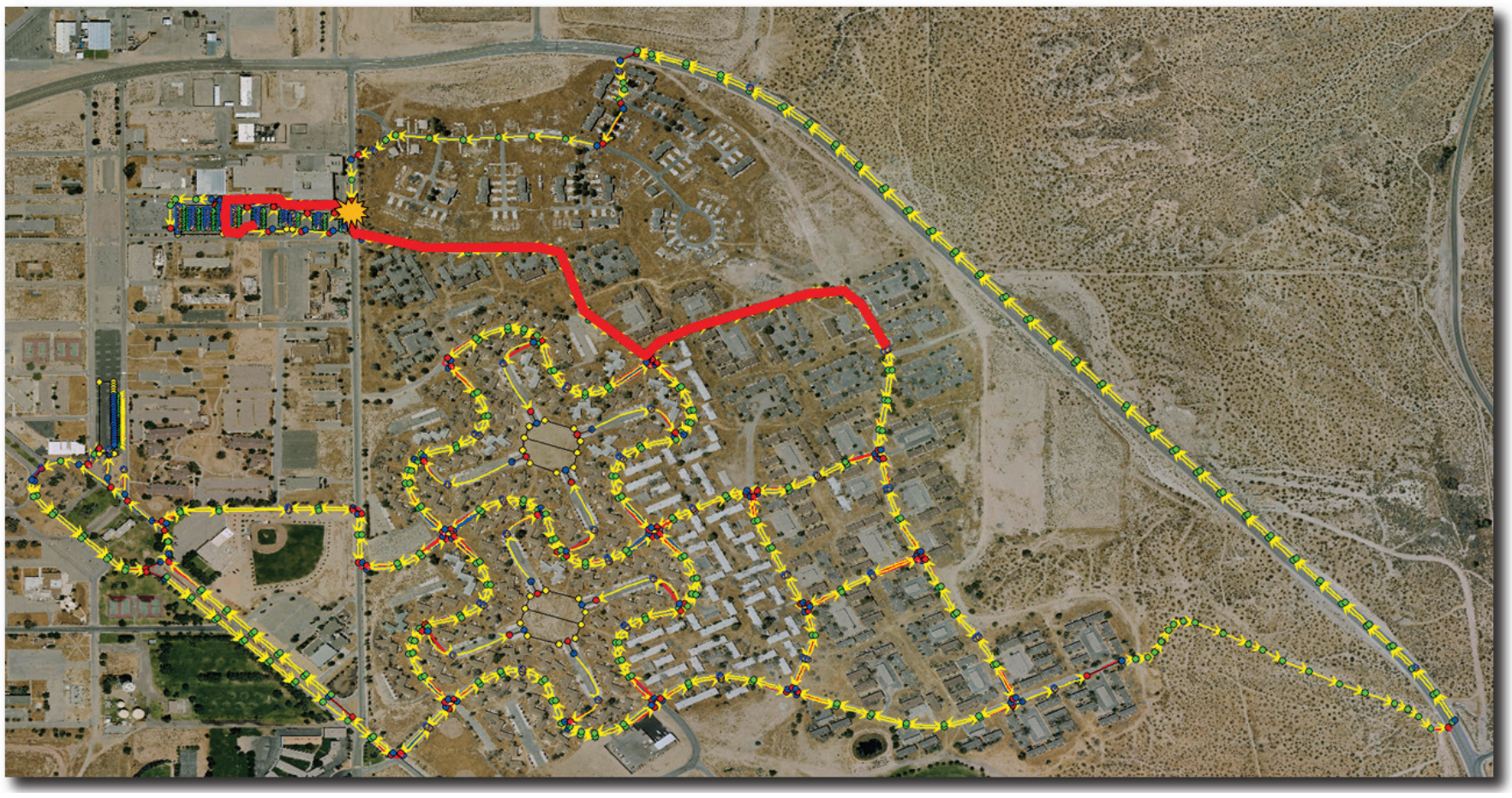

**Fig. 57.** Next incident including Caroline and Talos from team MIT.

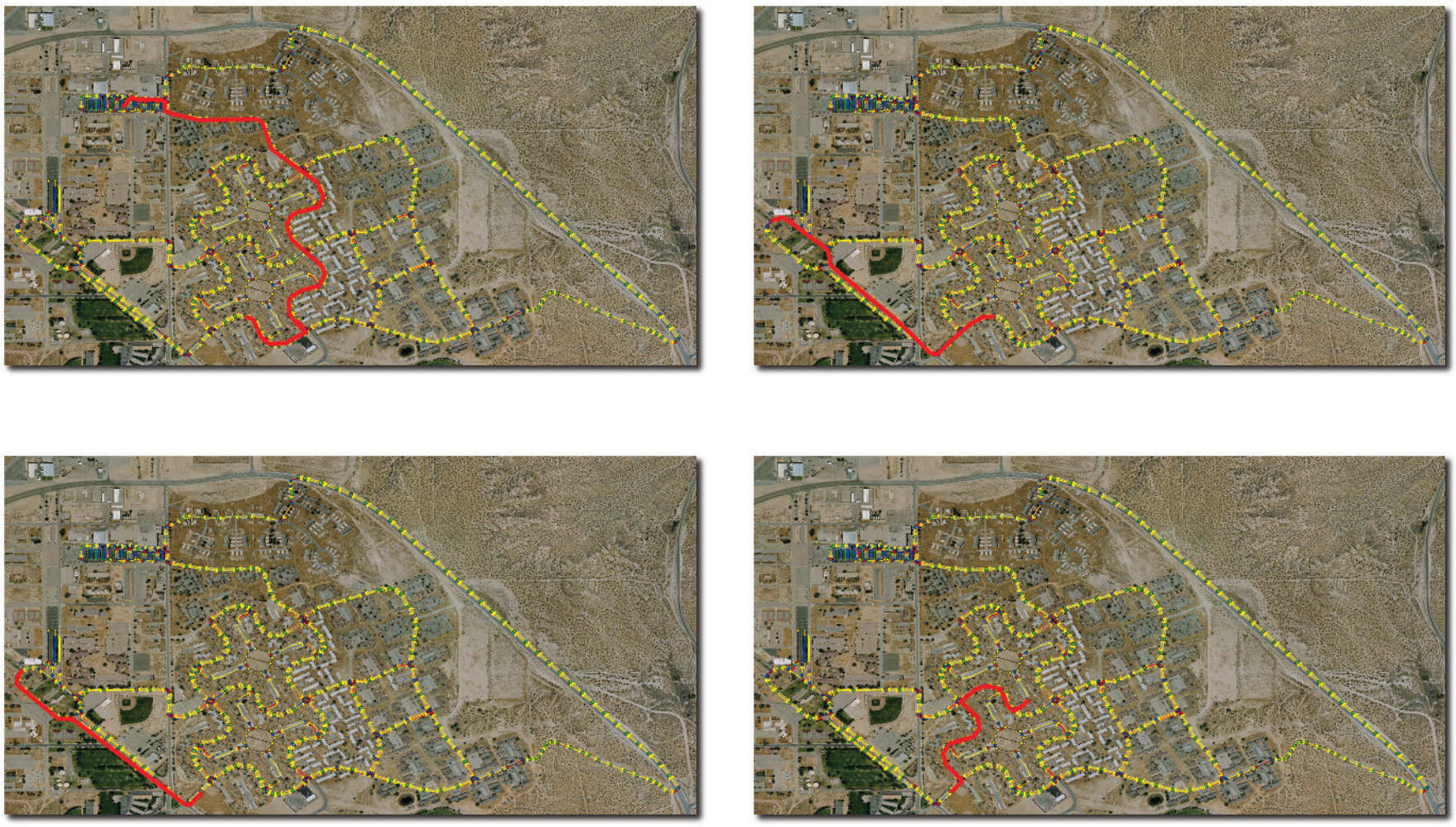

**Fig. 58.** Caroline went on after not yielding right of way to Talos.

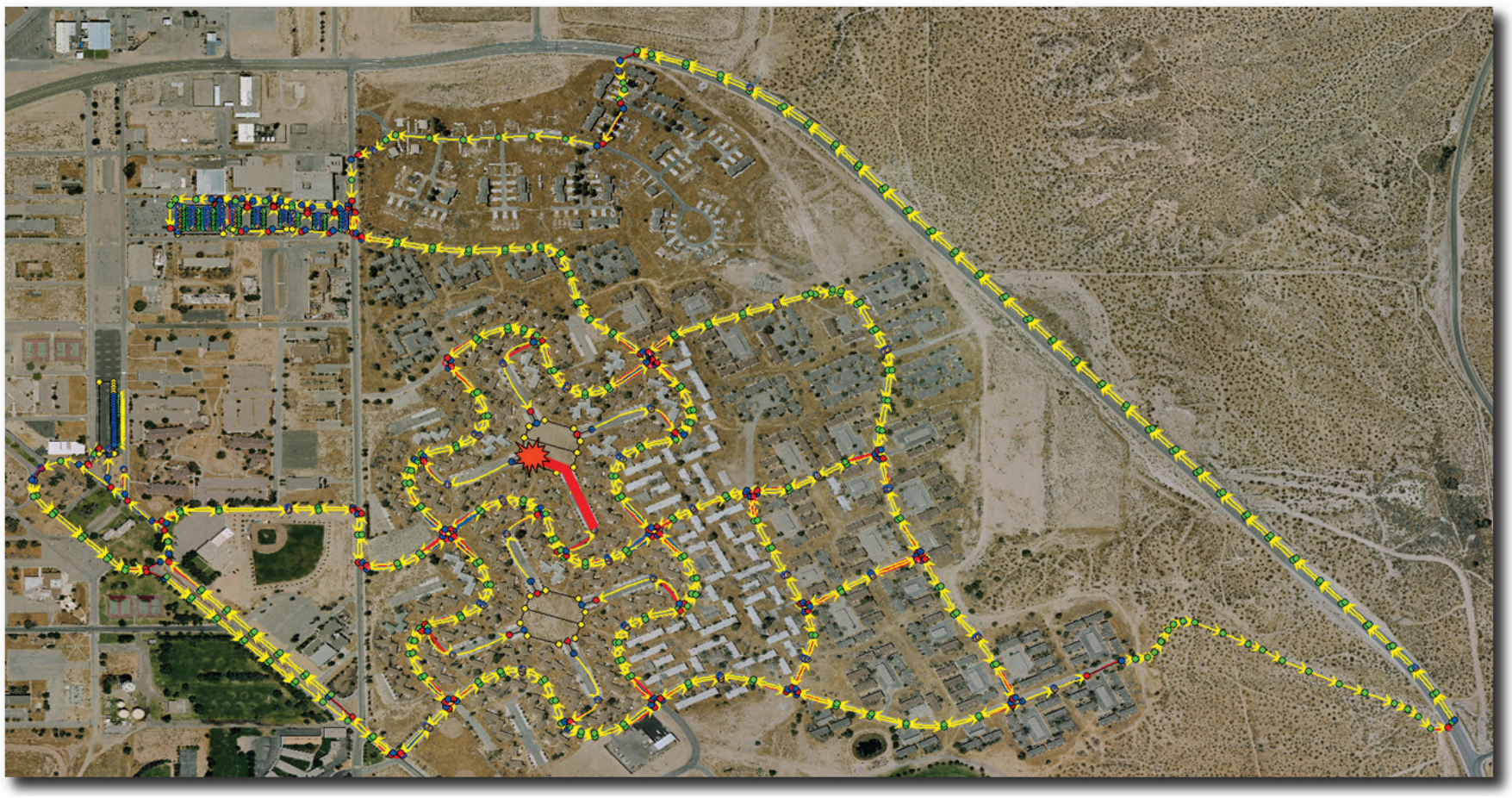

**Fig. 59.** Passed way before the first problem.

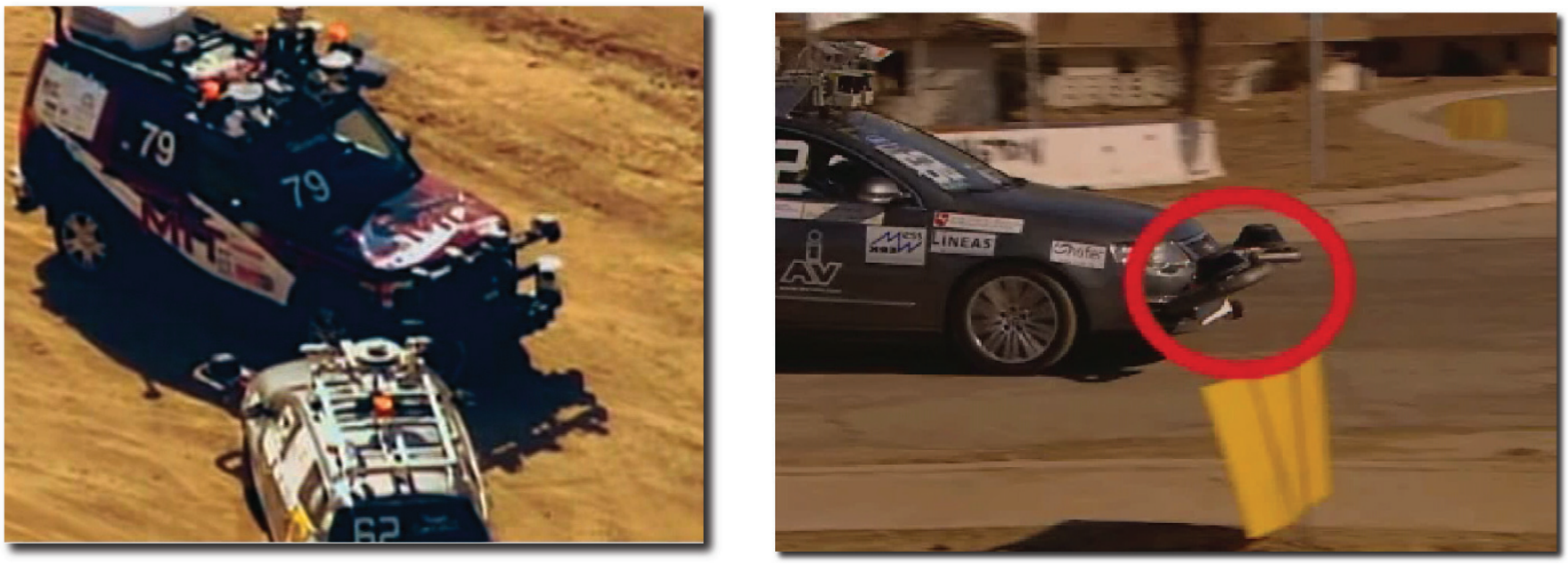

(a) Collision between Caroline and Talos from MIT.

(b) The red circle shows the twisted carrier rod.

**Fig. 60.** Caroline was retired after the collision with MIT.

## 7 Conclusion

Team CarOLO is an interdisciplinary team made up of members from the faculties of computer science and mechanical and electrical engineering which is significantly supported by industrial sponsors. Our vehicle Caroline is a standard 2006 Volkswagen Passat station wagon built to European specifications that is able to detect and track stationary and dynamic obstacles at a distance of up to 200 meters. The system's architecture comprises eight main modules: Sensor Data Acquisition, Sensor Data Fusion, Image Processing, Digital Map, Artificial Intelligence, Vehicle Path Planning and Low Level Control, Supervisory Watchdog and Online-Diagnosis, Telemetry and Data Storage for Offline Analysis. The signal flow through these modules is generally linear in order to decouple the development process. Our design approach uses multi-sensor fusion of lidar, radar and laser scanners, extending the classical point shape based approach to handle extensive dynamic targets expected in urban environments. Image processing detects lane markings along with drivable areas. Artificial intelligence is modeled according to DAMN architecture, redesigned and enhanced to meet requirements of special behavior in urban environments. Our approach is able to handle complex situations and ensure Caroline's proper behavior, e.g. obeying traffic regulations at intersections or performing U-turns when roads are blocked. Decisions of the artificial intelligence are sent to the path planner, which calculates optimal vehicle trajectories with respect to its dynamics in real time. Safety and robustness is ensured by supervisory watchdog monitoring of all vehicle's hardware and software modules. Failures or malfunctions immediately result in a safe and complete stop by Caroline. Since we are a large heterogeneous team with a very tight project schedule, we recognized very early the need for efficient quality assurance during the development process. Thus, we implemented an automatic multi-level test process. Each new feature or modification runs through a series of unit tests or comprehensive simulations before being deployed on the vehicle.

As a competitor in the DARPA Urban Challenge Final Event, Caroline is able to autonomously perform missions in urban environments. She drove approximately 17 kilometers in about three hours in the final.

## Acknowledgments

The authors thank their colleagues, students and professors from five institutes of the Technische Universität Braunschweig, who have developed Caroline. As a large amount of effort and resources were necessary to attempt this project, it would not have been successful if not for the many people from the university and local industry that had sponsored material, manpower and financial assistance. Particular thanks go to Volkswagen AG, IAV GmbH and the Ministry of Science and Culture of Lower Saxony. The authors' team also greatly thanks Dr. Bartels, Dr. Hoffmann, Professor Hesselbach, Mr. Horch, Mr. Lange, Professor Leohold, Dr. Lienkamp, Mr. Kuser, Professor Seiffert, Mr. Spichalsky, Professor Varchmin, Professor Wand and Mr. Wehner for their help on various occasions.

## References

Basarke et al., 2007a. Basarke, C., Berger, C., Homeier, K., Rumpe, B.: Design and quality assurance of intelligent vehicle functions in the "virtual vehicle". Virtual Vehicle Creation (2007a)

Basarke et al., 2007b. Basarke, C., Berger, C., Rumpe, B.: Software & systems engineering process and tools for the development of autonomous driving intelligence. Journal of Aerospace Computing, Information, and Communication 4 (2007b)

Beck, 2005. Beck, K.: Extreme Programming Explained: Embrace Change. Addison-Wesley, Reading (2005)

Beedle and Schwaber, 2002. Beedle, M., Schwaber, K.: Agile Software Development with Scrum. Prentice-Hall, Englewood Cliffs (2002)

Bilmes, 1997. Bilmes, J.: A gentle tutorial on the em algorithm and its application to parameter estimation for gaussian mixture and hidden markov models. Technical report (1997)

Collins-Sussmann et al., 2004. Collins-Sussmann, B., Fitzpatrick, B.W., Pilato, C.M.: Version Control with Subversion. O'Reilly, Sebastopol (2004)

Cormen et al., 2002. Cormen, T.H., Leiserson, C.E., Rivest, R.L., Stein, C.: Introduction to Algorithms, 2nd edn. (2002)

DARPA, 2006. DARPA, Technical evaluation criteria (2006)

Duda and Hart, 1973. Duda, R.O., Hart, P.E.: Pattern Classification and Scene Analysis. John Wiley & Sons Inc., Chichester (1973)

Edgewall Software, 2007. Edgewall Software, Trac. Edgewall Software (2007)

Gary Bradski, 2005. Bradski, G., Adrian Kaehler, V.P.: Learning-based computer vision with intels open source computer vision library, pp. 126–139 (2005)

Heikkil and Silvn, 1996. Heikkil, J., Silvn, O.: Calibration procedure for short focal length off-the-shelf ccd cameras. In: 13th International Conference on Pattern Recognition, Vienna, Austria, pp. 166–170 (1996)

Kalman, 1960. Kalman, R.E.: A new approach to linear filtering and prediction problems. In: Transactions of the ASME-Journal of Basic Engineering, pp. 35–45 (1960)

Liggesmeyer, 2002. Liggesmeyer, P.: Software-Qualitaet: Testen, Analysieren und Verifizieren von Software. Spektrum, Akad. Verl. (2002)

Nethercote and Seward, 2003. Nethercote, N., Seward, J.: Valgrind: A program supervising framework. Theoretical Computer Science 89 (2003)

OpenCV Website, 2007. OpenCV Website, The open cv library (2007)

Pitteway and M.L.V., 1967. Pitteway, M.L.V.: Algorithmn for drawing ellipses or hyperbolae with a digital plotter. Computer Journal 10(3), 282–289 (1967)

Rosenblatt, 1997. Rosenblatt, J.: DAMN: A Distributed Architecture for Mobile Navigation. PhD thesis, Robotics Institute, Carnegie Mellon University, Pittsburgh, PA (1997)

Shafer, 1976. Shafer, G.: A Mathematical Theory of Evidence. Princeton University Press, Princeton (1976)

Shafer, 1990. Shafer, G.: Perspectives on the theory and practice of belief functions. International Journal of Approximate Reasoning (3), 1–40 (1990)

Thrun et al., 2006. Thrun, S., Montemerlo, M., Dahlkamp, H., Stavens, D., Aron, A., Diebel, J., Fong, P., Gale, J., Halpenny, M., Hoffmann, G., Lau, K., Oakley, C., Palatucci, M., Pratt, V., Stang, P., Strohband, S., Dupont, C., Jendrossek, L.-E., Koelen, C., Markey, C., Rummel, C., van Niekerk, J., Jensen, E., Alessandrini, P., Bradski, G., Davies, B., Ettinger, S., Kaehler, A., Nefian, A., Mahoney, P.: Winning the darpa grand challenge. Journal of Field Robotics (2006)

Ulrich and Nourbakhsh, 2000. Ulrich, I., Nourbakhsh, I.: Appearance-based obstacle detection with monocular color vision. In: Proceedings of the AAAI National Conference on Artificial Intelligence, Austin, TX, pp. 866–871 (2000)